\documentclass{article} %
\usepackage{iclr2025_conference,times}

\usepackage{amsmath,amsfonts,bm}

\def\eqref#1{equation~\ref{#1}}

\def\1{\bm{1}}

\DeclareMathAlphabet{\mathsfit}{\encodingdefault}{\sfdefault}{m}{sl}
\SetMathAlphabet{\mathsfit}{bold}{\encodingdefault}{\sfdefault}{bx}{n}

\DeclareMathOperator*{\argmin}{arg\,min}

\usepackage{hyperref}
\usepackage{url}
\usepackage{tabularx,booktabs} %
\usepackage{graphicx}
\usepackage{amsmath}
\usepackage{bm}
\usepackage{mathtools}
\usepackage{xcolor, colortbl, soul}
\usepackage{enumitem}
\usepackage{relsize}
\usepackage{subcaption}
\usepackage{multirow}
\usepackage{wrapfig}
\usepackage{float}
\usepackage[export]{adjustbox}

\title{Perm: A Parametric Representation for Multi-Style 3D Hair Modeling}
\author{Chengan He$^{1}$\thanks{The work was mainly conducted at Adobe Research.}, Xin Sun$^{2}$, Zhixin Shu$^{2}$, Fujun Luan$^{2}$, S\"{o}ren Pirk$^{3}$, \\
\textbf{Jorge Alejandro Amador Herrera}$^{4}$, \textbf{Dominik L. Michels}$^{4}$, \textbf{Tuanfeng Y. Wang}$^{2}$, \\
\textbf{Meng Zhang}$^{5}$, \textbf{Holly Rushmeier}$^{1}$, \textbf{Yi Zhou}$^{2}$ \\
$^{1}$Yale University \quad $^{2}$Adobe Research \quad $^{3}$Kiel University \quad $^{4}$KAUST \quad $^{5}$NJUST \\
\texttt{\{chengan.he,holly.rushmeier\}@yale.edu} \\
\texttt{\{xinsun,zshu,fluan,yangtwan,yizho\}@adobe.com} \\
\texttt{sp@informatik.uni-kiel.de} \\
\texttt{\{jorge.amadorherrera,dominik.michels\}@kaust.edu.sa} \\
\texttt{mengzephyr@njust.edu.cn} \\
}

\newcommand{\new}[1]{\textcolor{black}{#1}}

\definecolor{vermilion}{rgb}{0.89, 0.26, 0.2}

\definecolor{burgundy_red}{rgb}{0.5, 0, 0.125}

\iclrfinalcopy %
\begin{document}

\maketitle

\begin{abstract}
We present \textsc{Perm}, a learned parametric representation of 3D human hair designed to facilitate various hair-related applications. Unlike previous work that jointly models the global hair structure and local curl patterns, we propose to disentangle them using a PCA-based strand representation in the frequency domain, thereby allowing more precise editing and output control. Specifically, we leverage our strand representation to fit and decompose hair geometry textures into low- to high-frequency hair structures, termed guide textures and residual textures, respectively. 
These decomposed textures are later parameterized with different generative models, emulating common stages in the hair grooming process. We conduct extensive experiments to validate the architecture design of \textsc{Perm}, and finally deploy the trained model as a generic prior to solve task-agnostic problems, further showcasing its flexibility and superiority in tasks such as single-view hair reconstruction, hairstyle editing, and hair-conditioned image generation.
More details can be found on our project page: \url{https://cs.yale.edu/homes/che/projects/perm/}.
\end{abstract}

\section{Introduction}
\label{sec:intro}
3D hair modeling, as a crucial and expensive component in the realm of digital humans for industries like gaming, animation, VFX, and virtual reality, combines the complex and artistic processes to model the geometry of individual strands to create specific hairstyles in the 3D environment.
With the recent availability of high-quality 3D hair data~\citep{hu2015single,shen2023CT2Hair,zhou2018hairnet}, machine learning methods have emerged for automatic hairstyle synthesis~\citep{zhou2023groomgen, sklyarova2023haar}, and 3D hair reconstruction from images~\citep{wu2022neuralhdhair, zheng2023hair, takimoto2024dr, kuang2022deepmvs, zhou2018hairnet} and monocular videos~\citep{wu2024monohair, luo2024gaussianhair, sklyarova2023neural_haircut}.

Despite their achievements, these methods often overlook the inherent \emph{biscale} nature of hair, where the global structure defines the hair flow, volume and length and the local structure defines the strand's curl patterns. 
Neglecting these biscale variations not only limits the quality of hair generation and reconstruction (illustrated in the comparisons in Fig.~\ref{fig:groomgen-comp} and Fig.~\ref{fig:single-view-supp}), but also restricts the editing capability on different scales.
Moreover, most of the models proposed in existing methods are heavy and \emph{task-specific}, lacking the generalization capability to different down-stream tasks. In contrast, the field of human body modeling has well-accepted lightweight parametric models like SMPL~\citep{SMPL:2015} and MANO~\citep{MANO:SIGGRAPHASIA:2017}, which designed disentangled coefficients for pose and shape and are widely applied as a generic prior in deep learning applications such as human reconstruction and animation.
To address this gap, we propose \textsc{Perm}, a parametric representation of 3D human hair that is both lightweight and generic for various hair-related applications.
\begin{figure}[ht]
    \centering
    \def\svgwidth{\textwidth}
    %% Creator: Inkscape 1.3.2 (091e20e, 2023-11-25), www.inkscape.org
%% PDF/EPS/PS + LaTeX output extension by Johan Engelen, 2010
%% Accompanies image file 'teaser.pdf' (pdf, eps, ps)
%%
%% To include the image in your LaTeX document, write
%%   \input{<filename>.pdf_tex}
%%  instead of
%%   \includegraphics{<filename>.pdf}
%% To scale the image, write
%%   \def\svgwidth{<desired width>}
%%   \input{<filename>.pdf_tex}
%%  instead of
%%   \includegraphics[width=<desired width>]{<filename>.pdf}
%%
%% Images with a different path to the parent latex file can
%% be accessed with the `import' package (which may need to be
%% installed) using
%%   \usepackage{import}
%% in the preamble, and then including the image with
%%   \import{<path to file>}{<filename>.pdf_tex}
%% Alternatively, one can specify
%%   \graphicspath{{<path to file>/}}
%% 
%% For more information, please see info/svg-inkscape on CTAN:
%%   http://tug.ctan.org/tex-archive/info/svg-inkscape
%%
\begingroup%
  \makeatletter%
  \providecommand\color[2][]{%
    \errmessage{(Inkscape) Color is used for the text in Inkscape, but the package 'color.sty' is not loaded}%
    \renewcommand\color[2][]{}%
  }%
  \providecommand\transparent[1]{%
    \errmessage{(Inkscape) Transparency is used (non-zero) for the text in Inkscape, but the package 'transparent.sty' is not loaded}%
    \renewcommand\transparent[1]{}%
  }%
  \providecommand\rotatebox[2]{#2}%
  \newcommand*\fsize{\dimexpr\f@size pt\relax}%
  \newcommand*\lineheight[1]{\fontsize{\fsize}{#1\fsize}\selectfont}%
  \ifx\svgwidth\undefined%
    \setlength{\unitlength}{900bp}%
    \ifx\svgscale\undefined%
      \relax%
    \else%
      \setlength{\unitlength}{\unitlength * \real{\svgscale}}%
    \fi%
  \else%
    \setlength{\unitlength}{\svgwidth}%
  \fi%
  \global\let\svgwidth\undefined%
  \global\let\svgscale\undefined%
  \makeatother%
  \begin{picture}(1,0.3625)%
    \lineheight{1}%
    \setlength\tabcolsep{0pt}%
    \put(0,0){\includegraphics[width=\unitlength,page=1]{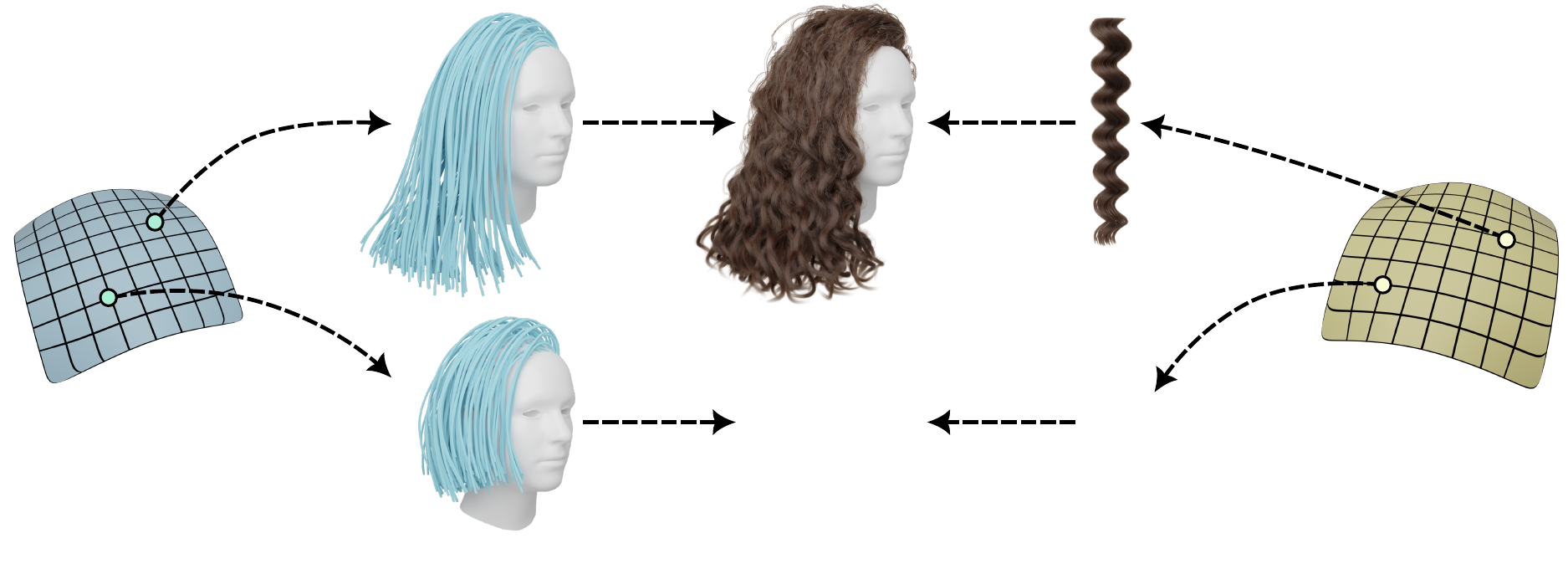}}%
    \put(0.09,0.085){\color[rgb]{0,0,0}\makebox(0,0)[t]{\lineheight{4}\smash{\begin{tabular}[t]{c}\fontsize{7pt}{1em} \textbf{Parameter Space of} $\vec{\theta}$\end{tabular}}}}%
    \put(0.9,0.085){\color[rgb]{0,0,0}\makebox(0,0)[t]{\lineheight{4}\smash{\begin{tabular}[t]{c}\fontsize{7pt}{1em} \textbf{Parameter Space of} $\vec{\beta}$\end{tabular}}}}%
    \put(0.31,0.0){\color[rgb]{0,0,0}\makebox(0,0)[t]{\lineheight{4}\smash{\begin{tabular}[t]{c}\fontsize{7pt}{1em} \textbf{Guide Strands}\end{tabular}}}}%
    \put(0.53,0.0){\color[rgb]{0,0,0}\makebox(0,0)[t]{\lineheight{4}\smash{\begin{tabular}[t]{c}\fontsize{7pt}{1em} \textbf{Full Hair Models}\end{tabular}}}}%
    \put(0.707,0.0){\color[rgb]{0,0,0}\makebox(0,0)[t]{\lineheight{4}\smash{\begin{tabular}[t]{c}\fontsize{7pt}{1em} \textbf{Wisps}\end{tabular}}}}%
    \put(0,0){\includegraphics[width=\unitlength,page=2]{teaser.pdf}}%
  \end{picture}%
\endgroup%

    \caption{\textsc{Perm} is a learned parametric representation of 3D human hair that is designed with disentangled parameters $\vec{\theta}$ and $\vec{\beta}$ to respectively control the global hair structure (represented as guide strands) and local curl patterns (represented as wisps).}
    \label{fig:teaser}
    \vspace{-2mm}
\end{figure}

\textsc{Perm} is novelly formulated as $\mathcal{M}(\vec{\theta}, \vec{\beta}, \mathcal{R})$, where the output hair is conditioned on the guide strand parameter $\vec{\theta}$ for its global structure, the hair styling parameter $\vec{\beta}$ for its curl patterns, and a pre-defined root set $\mathcal{R}$ for the hair density (as illustrated in Fig.~\ref{fig:teaser}).
To independently control the global structure and curl pattern, we propose a strand representation based on principal component analysis (PCA) in the frequency domain, which naturally aligns with the cyclical nature of hair growth~\citep{hoover2023physiology}. 
We use around three millions of different strands to train the PCA model and compare it with several deep neural network-based models.
Despite the PCA model's simplicity, it demonstrates high effectiveness in forming a compact subspace for strands that preserves geometry details much better than the deep neural network alternatives while demanding much less computation and memory consumption.

\begin{wrapfigure}[12]{r}{0.52\textwidth}
    \centering
    \vspace{-15pt}
    \addtolength{\tabcolsep}{-2pt}
    \begin{tabular}{cccc}
        \includegraphics[height=0.21\textwidth]{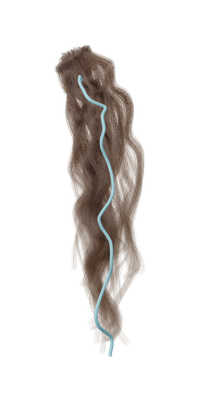} &
        \includegraphics[height=0.21\textwidth]{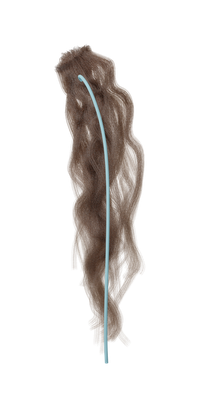} &
        \includegraphics[height=0.21\textwidth]{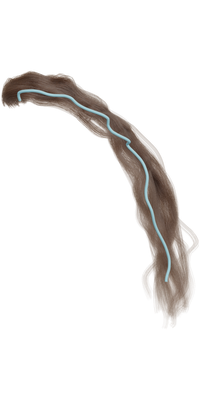} &
        \includegraphics[height=0.21\textwidth]{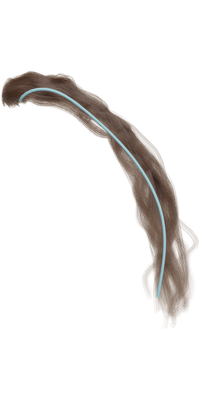} \\
        \tiny Center Strand & \tiny Our Guide Strand & \tiny Center Strand & \tiny Our Guide Strand \\
    \end{tabular}
    \caption{Guide strand of a wisp from PCA. 
    }
    \label{fig:wisp}
    \vspace{-2mm}
\end{wrapfigure}

The learned principal components form an \emph{interpretable} space for hair decomposition, where the initial components capture low-frequency information, such as the rough direction and length, while the subsequent components encode high-frequency details like curliness. As visualized in Fig.~\ref{fig:wisp}, this representation allows for the intuitive extraction of smooth guide strands from the given wisp, closer to what artists really use when grooming 3D hair in industrial software like Maya XGen~\citep{maya}.

For parameterizing the full hair, we follow previous literature to compute the $uv$ of the scalp, and map the geometry features of each strand onto a 2D texture based on its root position on the scalp. With the strand PCA representation, we can fit and store the dominant coefficients into \emph{guide textures} for guide strands and higher-order coefficients into \emph{residual textures} for detailed strand patterns.
These decomposed textures are then used to train a framework composed of multiple generative models as illustrated in Fig.~\ref{fig:overview}.

\begin{figure}[ht]
    \centering
    \addtolength{\tabcolsep}{-6.3pt}
    \begin{tabular}{cccccc}
      & & & & \multicolumn{2}{c}{\small \textit{``wavy and short hair, white sweater''}} \\
     \includegraphics[width=0.163\textwidth]{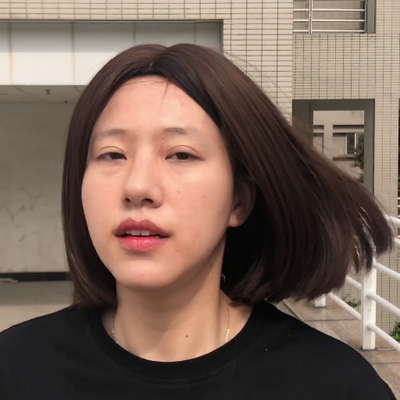} &
     \includegraphics[width=0.163\textwidth]{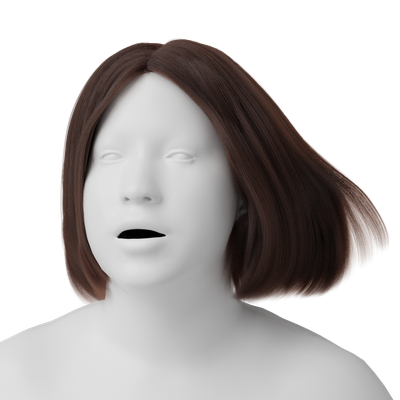} &
     \includegraphics[width=0.163\textwidth]{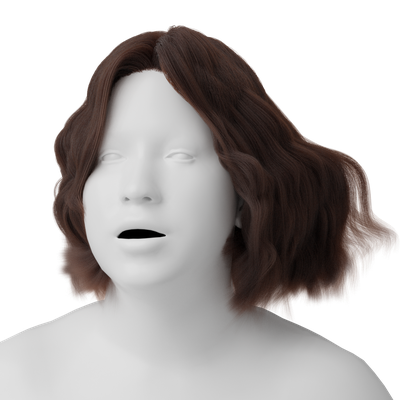} &
     \includegraphics[width=0.163\textwidth]{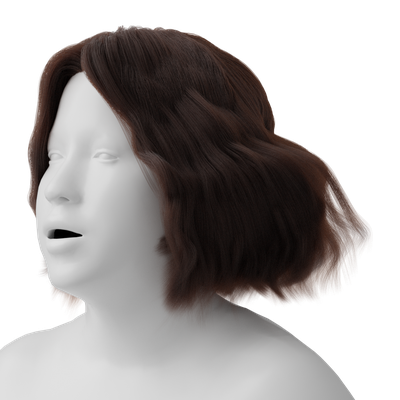} &
     \includegraphics[width=0.163\textwidth]{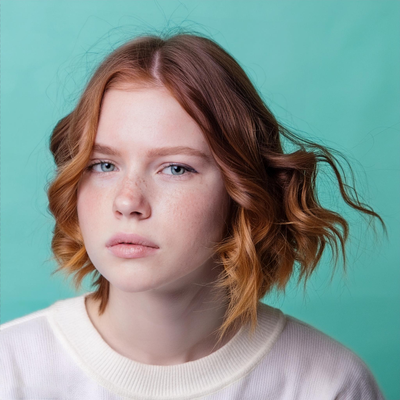} &
     \includegraphics[width=0.163\textwidth]{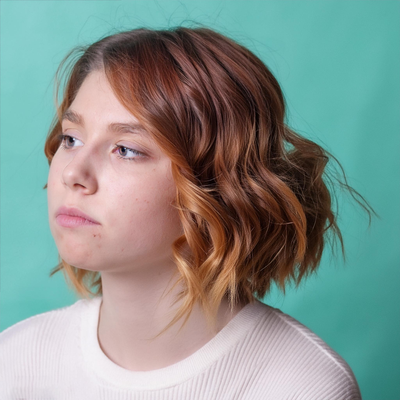} \\
     \small Input Image & \small Single-view Recon. & \multicolumn{2}{c}{\small \textsc{Perm} Parameter Editing} & \multicolumn{2}{c}{\small Hair-conditioned Image Generation} \\
    \end{tabular}
    \caption{An application demo of \textsc{Perm}, where we first fit \textsc{Perm} parameters to reconstruct 3D hair from the input image, then edit the parameters to change its style from straight to wavy, and finally use the edited hairstyle and certain viewing direction as a condition for image generation with Adobe Firefly~\citep{firefly}. }
    \label{fig:demo}
    \vspace{-2mm}
\end{figure}

After training \textsc{Perm} with $20k$ different hair styles, we obtain a compact, robust and editable generative hair model that can function as a generic hair prior for solving task-agnostic problems.
We demonstrate its capability across multiple applications, including single-view hair reconstruction and hairstyle editing, such as changing a hairstyle from smooth to bouncy while maintaining a similar haircut. Despite not being trained specifically for any of these tasks, \textsc{Perm} achieves performance equivalent or superior to state-of-the-art task-specific alternatives in our experiments. 
Moreover, we introduce a novel application of using \textsc{Perm}-generated hair for conditional image generation and editing hairstyles in the 3D latent space. A demo of these applications is provided in Fig.~\ref{fig:demo}.

\section{Related Work}
\label{sec:related}

Due to the limited availability of 3D hair data, early hair modeling works~\citep{wang2009hair} primarily focused on generating variations from a single hairstyle using methods like PCA. With the recent release of high-quality 3D hair datasets~\citep{hu2015single,shen2023CT2Hair,zhou2018hairnet}, machine learning methods have emerged to tackle various hair-related tasks. Hair representations used in these methods are generally categorized into two types: volumetric hair representation and strand-based hair representation.

\paragraph{Volumetric Hair Representation} Considering the volumetric nature of hair, \citet{saito2018_hairvae} first proposed a volumetric hair representation that converts a 3D hair into a combination of a 3D occupancy field and an orientation field, where strands could later be generated from discretized voxels. As an efficient spatially-aware intermediate representation for network training, this representation has been widely adopted in subsequent works for a range of hair-related applications, spanning from sketch-based hair modeling~\citep{shen2020deepsketchhair}, video-based dynamic hair modeling~\citep{yang2019dynamic,wang2023neuwigs}, to 3D hair reconstruction from images~\citep{zhang2019hair,takimoto2024dr, kuang2022deepmvs} or monocular videos~\citep{wu2024monohair, sklyarova2023neural_haircut}. However, since discrete volumetric fields are not well-suited for representing curly and kinky hair, another line of research has sought to improve this volumetric representation. These advancements include implicit volumetric representations~\citep{wu2022neuralhdhair,zheng2023hair} and hybrid representations that attach multiple neural volumetric primitives to pre-obtained sparse guide strands~\citep{wang2022hvh,wang2023local}.

\paragraph{Strand-based Hair Representation} Since real human hair consists of individual strands, the most intuitive way to represent it is as a collection of strands, where each strand can be represented as a 3D polyline with a fixed number of points. \citet{zhou2018hairnet} adopted this strand-based representation and proposed to organize strand features as a 2D hair texture according to the scalp $uv$ parameterization, which facilitates the training of 2D convolutional networks. This idea has been expanded by subsequent works, incorporating more advanced networks such as U-Net~\citep{ronneberger2015u} and diffusion models~\citep{ho2020denoising} to solve tasks such as neural hair rendering~\citep{rosu2022neuralstrands}, text-conditioned hair generation~\citep{sklyarova2023haar} and 3D hair capture from monocular videos~\citep{sklyarova2023neural_haircut}. While these diffusion-based methods produce impressive results, they typically involve a heavy denoising process and lack a compact latent space for hair reconstruction and editing. Recent work on 3D Gaussian Splatting~\citep{kerbl3Dgaussians} offers a potential enhancement to this representation, where Gaussian splats can be attached to each segment of the strand. This hybrid representation enables the joint learning of both hair geometry and appearance, showing promise in 3D hair capture~\citep{zakharov2024gh, luo2024gaussianhair, zhou2024groomcap}. However, all these strand-based works are task-specific and neglect the biscale nature of hair, thereby limiting the quality and versatility of the generated or reconstructed hairstyles.

\paragraph{Generative Hair Model} GroomGen~\citep{zhou2023groomgen}, most related to our work, learns a hair prior using hierarchical latent spaces. However, their hierarchy is only for the low-resolution hair texture, which refers to a sparse set of strands, and the high-resolution hair texture, which refers to the final full hair, is upsampled in a deterministic way. Although they denote the sparse strands as guide strands, those are just like the center strands as illustrated in Fig.~\ref{fig:wisp} without the filtered structure. They do not have the disentangled parameters for controlling the overall hair structure features and the detailed hair curl patterns, which leads to a different architecture design from our method.

While GroomGen reported promising results in their paper, we found it is both theoretically and experimentally unreliable through our experiments. Its VAE-based strand representation struggles to preserve the strand structure and the entire architecture is very unstable to train. We further found that some curly hairstyles showcased in the paper are unachievable with the described strand resolution. We include our experiments of GroomGen in Appendix~\ref{supp:groomgen}.

\paragraph{PCA for Hair} 
The biology community has widely adopted PCA to analyze and classify different hair types in the real world~\citep{hoover2023physiology,de2007shape,loussouarn2007worldwide}. We transfer these biological insights into the machine learning domain, introducing the first neural network architecture to incorporate a PCA-based strand representation for human hair. While \citet{wang2009hair} previously explored PCA-based representations for variations of a single hair model, we aim at generating high variety of hairstyles trained on large-scale dataset. This approach has proven to be both efficient and highly accurate in preserving strand details, which can be deployed as a generic hair prior to solve task-agnostic problems.

\section{Model Formulation}
\label{sec:model}

\begin{figure*}[ht]
    \centering
    \includegraphics[width=\linewidth]{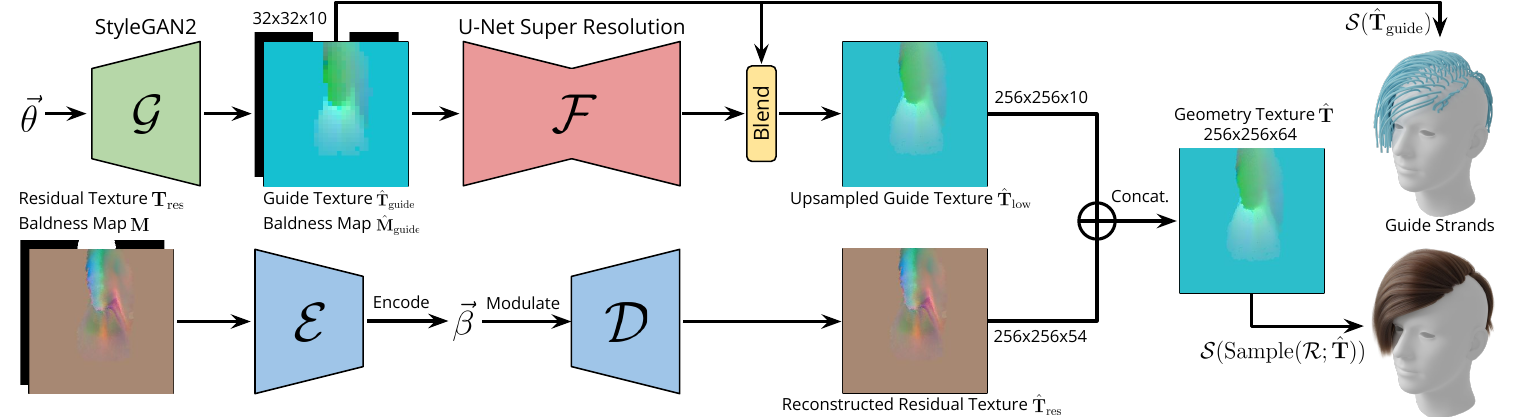}
    \vspace{-10pt}
    \caption{Architecture overview. 
    }
    \label{fig:overview}
    \vspace{-2mm}
\end{figure*}

In \textsc{Perm}, hair is represented as geometry textures $\mathbf{T}$ storing the strand PCA coefficients $\vec{\gamma}$ (Sec.~\ref{sec:strand-repr}). These textures are decomposed into a lower-order component, the guide texture, which captures the global hair structure, and a higher-order component, the residual texture, which encodes local curl pattern information.
To parameterize these textures, a parameter $\vec{\theta}$ is designed to generate the guide textures $\mathbf{T}_{\text{guide}}$ and $\mathbf{M}_{\text{guide}}$ through the function $\mathcal{G}(\vec{\theta})$, covering the sparse set of guide strands and the hair growing area on the scalp respectively (Sec.~\ref{sec:guide-tex-synthesis}). These guide textures are upsampled with a deterministic function $\mathcal{F}(\cdot)$, whose output contains the complete hair structure but lacks high-frequency details (Sec.~\ref{sec:guide-tex-upsample}). To complement that, another parameter $\vec{\beta}$ is introduced to infer the residual texture $\mathbf{T}_{\text{res}}$ through the function $\mathcal{D}(\vec{\beta})$, storing high-frequency information such as different curl patterns (Sec.~\ref{sec:res-tex-synthesis}). Concatenating the upsampled guide texture and residual texture yields the geometry texture, which can then be sampled and decoded with the strand function $\mathcal{S}(\vec{\gamma})$ formulated in Sec.~\ref{sec:strand-repr} to produce the final strand geometry. An overview of this architecture is illustrated in Fig.~\ref{fig:overview}, and the resulting \textsc{Perm} model, denoted as $\mathcal{M}(\vec{\theta}, \vec{\beta}, \mathcal{R})$, can then be expressed as:
\begin{equation}
    \mathcal{M}(\vec{\theta}, \vec{\beta}, \mathcal{R}; \Phi) = \mathcal{S}\Big(\text{Sample}\big(\mathcal{R}; \mathcal{F}(\mathcal{G}(\vec{\theta}))\oplus\mathcal{D}(\vec{\beta})\big)\Big),
\end{equation}
where $\text{Sample}(\cdot)$ denotes nearest neighbor interpolation used to sample the synthesized texture with 2D root coordinates in the pre-defined root set $\mathcal{R}$, $\oplus$ represents the concatenation operation along the feature channels, and $\Phi$ specifies the full set of trainable parameters in \textsc{Perm} used to approximate the functions above. Once trained, parameters in $\Phi$ are held fixed, and novel 3D hairstyles are synthesized by varying $\vec{\theta}$ and $\vec{\beta}$ respectively, while the hair density can be adjusted by manipulating the number of roots in $\mathcal{R}$.
The dimension of parameters $\vec{\theta}$ and $\vec{\beta}$ are both set to $512$ in our model ($|\vec{\theta}| = |\vec{\beta}| = 512$). In the following, we delve into the details of each term in our model.

\subsection{Strand Representation}
\label{sec:strand-repr}

For human hair, its cyclical growth behavior has been extensively documented in the biology community, where PCA has also been employed to analyze different hair types~\citep{hoover2023physiology,de2007shape,loussouarn2007worldwide}.
Inspired by these biological findings, we define a low-dimensional parameter space based on PCA in the frequency domain.
Formally, for 3D hair strands with a fixed number of points $\mathbf{S} = \{\mathbf{p}_1, \mathbf{p}_2, \dots, \mathbf{p}_L\} \in \mathbb{R}^{L\times 3}$ ($L=100$ in our experiments), they are represented as the output of a linear function $\mathcal{S}(\vec{\gamma})$ in the frequency domain:
\begin{equation}
\label{eq:strand-pca}
    \mathcal{S}(\vec{\gamma}; \mathbf{X}) = \text{iDFT}(\bar{\mathbf{S}} + \vec{\gamma}^\top\mathbf{X}) = \text{iDFT}(\bar{\mathbf{S}} + \sum_{n=1}^{|\vec{\gamma}|}\vec{\gamma}_n\mathbf{X}_n),
\end{equation}
where $\vec{\gamma} = [\gamma_1, \dots, \gamma_{|\vec{\gamma}|}]^\top$ is a vector of strand coefficients, and $\mathbf{X} = [\mathbf{X}_1, \mathbf{X}_2, \dots, \mathbf{X}_{|\vec{\gamma}|}]^\top \in \mathbb{R}^{|\vec{\gamma}| \times 6k}$ forms a matrix of orthonormal principal components that capture phase variations in different strands, with $k=\lfloor L / 2 \rfloor + 1$ referring the number of frequency bands. The term $\bar{\mathbf{S}}=[\bar{\mathbf{S}}_{\text{real}}, \bar{\mathbf{S}}_{\text{imag}}] \in \mathbb{R}^{k \times 3 \times 2}$ represents the mean phase vector of strands, and $\text{iDFT}(\cdot)$ denotes the inverse Discrete Fourier Transform that maps strands back from the frequency domain to the spatial domain. In other sections, we often omit $\mathbf{X}$ in Eq.~\ref{eq:strand-pca} for notational convenience.

To obtain quantities in Eq.~\ref{eq:strand-pca}, we first apply the Discrete Fourier Transform (DFT) along the $x$, $y$, $z$ axes to all strands in our collected hair models, and then compute the mean phase vector $\bar{\mathbf{S}}$ and solve the principal components $\mathbf{X}$ by performing PCA on the computed Fourier bases, which is similar to the prevalent PCA-based modeling methods in digital humans~\citep{blanz19993dmm, SMPL:2015, FLAME:SiggraphAsia2017}. We set the number of PCA coefficients $|\vec{\gamma}|=64$, which explained almost $100\%$ of the variance in the training set (see Fig.~\ref{fig:strand-pca}).

Our PCA-based representation also facilitates a meaningful decomposition of strand geometry by forming an \emph{interpretable} parameter space. Our analysis in Appendix~\ref{supp:strand-repr-exp-advanced} reveals that the first $10$ PCA coefficients are enough to effectively capture the global structure of each strand, and the remaining $54$ coefficients encode high-frequency details, such as curl patterns (see Fig.~\ref{fig:pca-analysis}).
Leveraging this intuitive decomposition, we can effortlessly edit a given hairstyle or transfer the curl patterns from one hairstyle to another. Examples are provided in Fig.~\ref{fig:strand-repr}.

\subsection{Full Hair Representation}
\label{sec:tex-repr}

We define a 2D parameterization of hairstyles on the scalp surface as a regular $uv$ texture map, where each texel stores the strand PCA coefficients $\vec{\gamma}$ formulated in Eq.~\ref{eq:strand-pca}. These textures are referred to as hair geometry textures, denoted as $\mathbf{T} \in \mathbb{R}^{256 \times 256 \times 64}$. Similar to GroomGen~\citep{zhou2023groomgen}, we separately model the baldness area as a baldness map $\mathbf{M} \in \mathbb{R}^{256 \times 256}$. Both $\mathbf{T}$ and $\mathbf{M}$ are downsampled to $32 \times 32$ to represent guide strands (denoted as the guide texture $\mathbf{T}_{\text{guide}}$ and mask $\mathbf{M}_{\text{guide}}$), where only the low-rank PCA coefficients ($|\vec{\gamma}_{\text{guide}}|=10$) are kept in $\mathbf{T}_{\text{guide}}$.
As not all texels will be decoded to meaningful strands, $\mathbf{T}_{\text{guide}}$ typically accommodates around $400$ strands, as visualized in Fig.~\ref{fig:teaser}.
Note that the concept of guide strands here is different from previous works~\citep{zhou2023groomgen,shen2023CT2Hair}, where the guide strands are just quantitatively down-sampled strands as visualized in Fig.~\ref{fig:wisp} without the filtered structure.

Since guide strands only serve as a sparse representation of the full hair, an upsampling process is necessary to obtain the final hair strands. However, as illustrated in GroomGen, this process entails a \emph{one-to-many} mapping~\citep{zhou2023groomgen}, wherein the same guide strands can be upsampled to yield diverse hairstyles with varying strand randomness and curliness. Achieving this property proves challenging, necessitating an orthogonal decomposition of the global hair structure and local curl patterns. Previous methods thus either downgraded to a deterministic one-to-one mapping~\citep{sklyarova2023haar} or resorted to a manual post-processing step~\citep{zhou2023groomgen}.
To address this challenge, we further decompose our hair geometry texture $\mathbf{T}$ into $\mathbf{T}_{\text{low}}\in\mathbb{R}^{256 \times 256 \times 10}$ and $\mathbf{T}_{\text{res}}\in\mathbb{R}^{256 \times 256 \times 54}$ by splitting along the feature channels. Given the explainability of our strand PCA coefficients, we discern that $\mathbf{T}_{\text{low}}$ encapsulates the global hair structure, while $\mathbf{T}_{\text{res}}$ embodies the local curl patterns, denoted as the residual texture. 

Leveraging this decomposition, we employ different neural networks to parameterize them, simulating the automated process of hair modeling pipeline.

\subsubsection{Guide Texture Synthesis}
\label{sec:guide-tex-synthesis}
Formally, for guide textures, we train a neural network to approximate the function, $\mathcal{G}(\vec{\theta}; \phi_1)$: $\mathbb{R}^{|\vec{\theta}|} \mapsto \mathbb{R}^{32 \times 32 \times (10+1)}$, which synthesizes $\mathbf{T}_{\text{guide}}$ and $\mathbf{M}_{\text{guide}}$ from the guide strand parameter $\vec{\theta}$, where $\phi_1$ denotes the trainable network parameters.
In our formulation, we choose StyleGAN2~\citep{Karras2019stylegan2} as the backbone of our model, which has been proved to be a powerful generator in both 2D images and 3D feature tri-planes~\citep{Chan2022eg3d}. It also ensures our guide strand parameter will follow the normal distribution $\vec{\theta} \sim \mathcal{N}(\mathbf{0}, \mathbf{I})$, making it easy to sample. Moreover, the intermediate $\mathcal{W}$ and $\mathcal{W}+$ spaces introduced in StyleGAN allow for more faithful latent embedding~\citep{Abdal_2019_ICCV}, which in our case facilitates hair reconstruction either from 3D strands or 2D orientation images.
Since our generator synthesizes both the guide texture $\mathbf{T}_{\text{guide}}$ and mask $\mathbf{M}_{\text{guide}}$, we use a similar dual discrimination method as EG3D~\citep{Chan2022eg3d}, where we concatenate these two images and feed them into the discriminator. This operation ensures consistency between the generated texture and mask, which helps the generator to place zero-length strands in bald areas.
Our training objective is identical to StyleGAN2, which consists of a non-saturating GAN loss~\citep{goodfellow2014generative} with $R_1$ regularization~\citep{mescheder2018training} on both the texture and mask, where the regularization strengths are set to $5$ and $1$, respectively.

\subsubsection{Guide Texture Upsampling}
\label{sec:guide-tex-upsample}
To upsample the synthesized guide textures, we formulate it as an image super resolution problem, which can be defined as the function, $\mathcal{F}(\mathbf{T}_{\text{guide}}, \mathbf{M}_{\text{guide}}; \phi_2): \mathbb{R}^{32 \times 32 \times 11} \mapsto \mathbb{R}^{256 \times 256 \times 10}$, with trainable network parameters $\phi_2$. Note that this function operates as a deterministic mapping without parameter control, which aligns with both previous deep learning-based methods~\citep{zhou2023groomgen, sklyarova2023haar} and the adaptive interpolation algorithm used in current hair modeling software.
To approximate $\mathcal{F}(\cdot)$, we train a U-Net~\citep{ronneberger2015u} on the bilinearly upsampled textures, which translates them to $14$-channel weight maps, where the first $4$ channels represent the weight vector $\vec{\omega}$ for the $4$ neighboring guide strands, and the last $10$ channels represent a residual vector $\vec{\delta}$ to correct the blended coefficients. Therefore, the final coefficient $\vec{\gamma}_{\text{low}}$ can be calculated as:
\begin{equation}
    \vec{\gamma}_{\text{low}} = \vec{\omega}^\top\bm{\gamma}_{\text{guide}} + \vec{\delta} = \sum_{n=1}^4\omega_n\vec{\gamma}_{\text{guide}}^n + \vec{\delta}.
\end{equation}
Our experiments demonstrate that this weight-based blending output allows the network to converge to a sharper result compared to predicting coefficients directly.

We train the U-Net in a supervised manner, where we employ both an $L_1$ loss for the blended texture $\hat{\mathbf{T}}_{\text{low}}$ and a geometric loss $\mathcal{L}_{\text{geo}}$ for the decoded strand geometry. Specifically, the geometric loss $\mathcal{L}_{\text{geo}}$ encompasses an $L_1$ loss for the point position $\hat{\mathbf{p}}_n$, a cosine distance for the orientation $\hat{\mathbf{d}}_n=\hat{\mathbf{p}}_{n+1} - \hat{\mathbf{p}}_n$~\citep{rosu2022neuralstrands}, and an $L_1$ loss for the curvature, represented as the $L_2$ norm of binormal vector $\hat{\mathbf{b}}_n = \|\hat{\mathbf{d}}_n \times \hat{\mathbf{d}}_{n+1} \|_2$~\citep{sklyarova2023neural_haircut}:
\begin{equation}
    \mathcal{L}_{\text{geo}} = \sum_{n=1}^L \|\hat{\mathbf{p}}_n - \mathbf{p}_n \|_1 + (1 - \hat{\mathbf{d}}_n \cdot \mathbf{d}_n) + \|\hat{\mathbf{b}}_n - \mathbf{b}_n\|_1.
\end{equation}
A regularization term is included as well to constrain the residual vector $\vec{\delta}$ towards $0$. The overall loss function, denoted as $\mathcal{L}_{\text{superres}}$, then can be expressed as:
\begin{equation}
    \label{eq:superres}
    \mathcal{L}_{\text{superres}} = \lambda_{\text{tex}}\|\hat{\mathbf{T}}_{\text{low}} - \mathbf{T}_{\text{low}}\|_1 + \lambda_{\text{geo}}\mathcal{L}_{\text{geo}} + \lambda_{\text{reg}}\|\vec{\delta}\|_2^2,
\end{equation}
where the weighting factors $\lambda_{\text{tex}}$, $\lambda_{\text{geo}}$ and $\lambda_{\text{reg}}$ are set to $1$, $1$ and $0.1$, respectively.

\subsubsection{Residual Texture Synthesis}
\label{sec:res-tex-synthesis}
To simulate the artistic process of creating different curl patterns, we train a neural network to approximate the function, $\mathcal{D}(\vec{\beta}; \phi_3): \mathbb{R}^{|\vec{\beta}|} \mapsto \mathbb{R}^{256 \times 256 \times 54}$, where we learn to synthesize the residual texture $\mathbf{T}_{\text{res}}$ from the hair styling parameter $\vec{\beta}$, and $\phi_3$ represents the trainable network parameters.

Although StyleGAN2 performs well in generating textures for guide strands, we found it struggles with residual textures, as they contain more data than high-resolution RGB images. Therefore, we opted for VAE~\citep{kingma2013auto} as the backbone, where the encoder adopts an architecture similar to pSp~\citep{richardson2021encoding}, and the decoder takes the same architecture as StyleGAN2 generator. Essentially, the encoder $\mathcal{E}$ projects the residual texture $\mathbf{T}_{\text{res}}$ and baldness map $\mathbf{M}$ into the latent space, where the latent vectors will then be used to modulate the decoder to reconstruct the input. A similar loss function as Eq.~\ref{eq:superres} is defined as the training objective for the VAE, which can be expressed as:
\begin{equation}
    \mathcal{L}_{\text{res}} = \lambda_{\text{tex}}(\|\hat{\mathbf{T}}_{\text{res}} - \mathbf{T}_{\text{res}}\|_1 + \|\hat{\mathbf{M}} - \mathbf{M} \|_1) + \lambda_{\text{geo}}\mathcal{L}_{\text{geo}} + \lambda_{\text{KL}}\mathcal{L}_{\text{KL}},
\end{equation}
where the reconstruction terms are computed on the residual texture, baldness map, and decoded strand geometry. The weighting factors $\lambda_{\text{tex}}$, $\lambda_{\text{geo}}$, and $\lambda_{\text{KL}}$ are set to $10$, $1$, and $1e^{-4}$, respectively.

\section{Experiments}
\label{sec:exp}

We train \textsc{Perm} on an augmented version of USC-HairSalon~\citep{hu2015single}, which contains a total of $21,054$ data samples. For evaluation, we compiled a separate dataset of $17$ publicly available hair models, comprising $151,829$ strands in total. Detailed data augmentation and source of our testing data are provided in Appendix~\ref{supp:dataset}.
Since USC-HairSalon lacks sufficient curly hair data, we curated a private dataset consisting of $80$ manually groomed hairstyles, comprising a total of $4,368,679$ strands with a greater diversity of style. We conducted experiments on this private dataset to further validate the robustness of our representation, whose results are detailed in Appendix~\ref{supp:strand-repr-exp-advanced}. 

\subsection{Strand Representation}
\label{sec:strand-repr-exp}

We first conducted extensive experiments to evaluate our PCA-based strand representation against various alternative deep learning-based representations and a simpler PCA-based formulation without DFT. Contrary to the common belief that neural network methods outperform linear models like PCA, our experiments show that the PCA method performs better in the context of hair modeling.

To quantitatively assess their reconstruction capabilities, we present the mean position error (pos. err.), computed as the average Euclidean distance between corresponding points on the reconstructed strands and the ground truth. Additionally, we report the mean curvature error (cur. err.), defined as the $L_1$ norm between the curvatures of reconstructed and ground truth strands. Please refer to Appendix~\ref{supp:metrics} for details of the metrics.

\begin{table}[ht]
    \centering
    \caption{Reconstruction errors reported on different strand representations. Here \textbf{boldface} corresponds to the best result and \underline{underline} means the second best.}
    \label{tab:strand-repr}
    \addtolength{\tabcolsep}{-0.2em}
    \begin{tabularx}{\columnwidth}{*{5}{Xcccccc}}
    \toprule
                  & Freq. VAE & CNN VAE      & Transformer VAE & MLP VAE  & PCA          & Freq. PCA (Ours)   \\ \midrule
    \# params.    & $15.67$M  & $759.24$K    & $2.29$M         & $15.67$M & $19.31$K     & $19.50$K     \\ \midrule          
    pos. err.     & $1.211$   & $0.446$      & $0.288$         & $0.158$  & $\bm{0.019}$ & $\underline{0.020}$ \\ \midrule
    cur. err.     & $\bm{0.910}$   & $6.250$      & $8.539$         & $\underline{1.594}$  & $2.361$      & $2.150$      \\ \bottomrule
  \end{tabularx}
  \vspace{-2mm}
\end{table}

\begin{figure*}[ht]
    \centering
    \addtolength{\tabcolsep}{-7pt}
    \begin{tabular}{lccccccc}
        \includegraphics[height=0.16\textwidth]{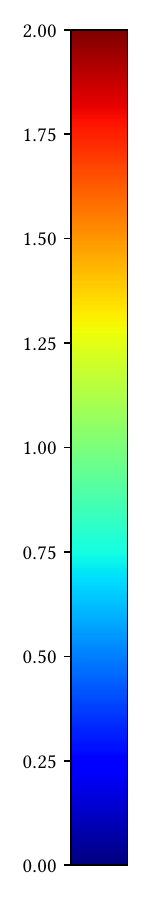}  &
        \includegraphics[width=0.144\textwidth, trim=40 0 40 0, clip]{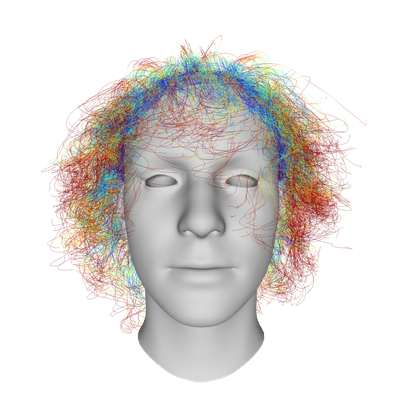} &
        \includegraphics[width=0.144\textwidth, trim=40 0 40 0, clip]{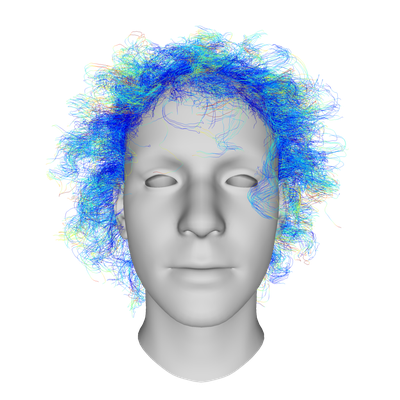} &
        \includegraphics[width=0.144\textwidth, trim=40 0 40 0, clip]{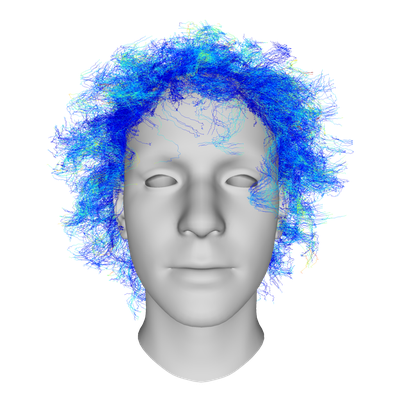} &
        \includegraphics[width=0.144\textwidth, trim=40 0 40 0, clip]{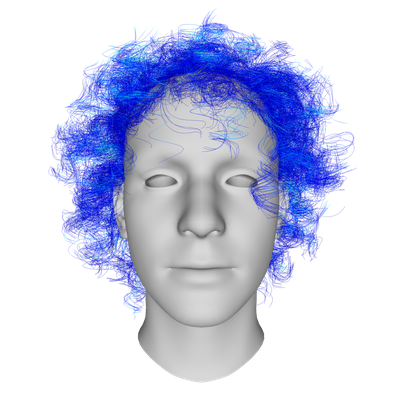} &
        \includegraphics[width=0.144\textwidth, trim=40 0 40 0, clip]{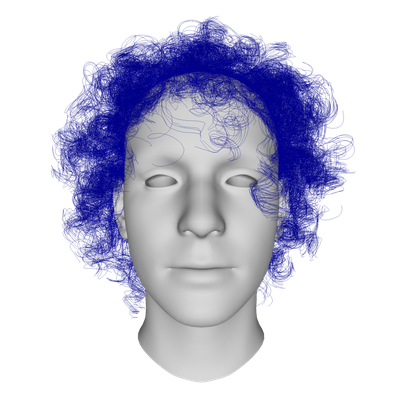} &
        \includegraphics[width=0.144\textwidth, trim=40 0 40 0, clip]{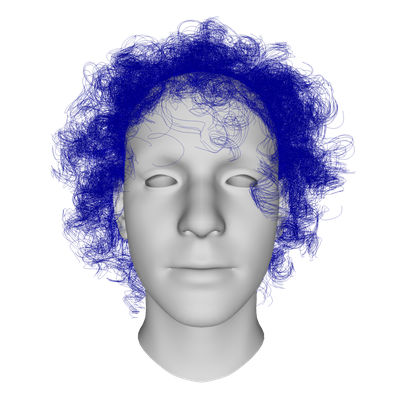} &
        \includegraphics[width=0.144\textwidth, trim=40 0 40 0, clip]{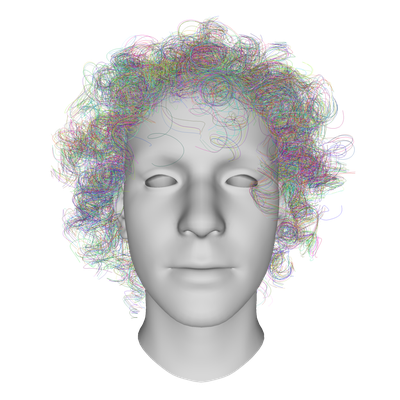} \\
         & \tiny Freq. VAE & \tiny CNN VAE & \tiny Transformer VAE & \tiny MLP VAE & \tiny PCA & \tiny Freq. PCA (Ours) & \tiny Ground Truth
    \end{tabular}
    \caption{
    Comparison of our PCA-based strand representation (Freq. PCA) with other VAE and PCA-based representations.
    Reconstructed strands are color-coded by their position error.
    }
    \label{fig:strand-repr-comp}
    \vspace{-2mm}
\end{figure*}

Quantitative results are reported in Table~\ref{tab:strand-repr}, where the deep learning-based strand representations include a VAE model trained with frequency features (Freq. VAE)~\citep{zhou2023groomgen}, and other VAE variants with CNN, Transformer~\citep{vaswani2017attention}, and ModSIREN MLP~\citep{mehta2021modulated} decoders. All configurations compress strands to $64$-dimensional vectors. While simple, our PCA-based representations demonstrate a remarkably small position error, achieving this with a significantly reduced number of parameters compared to all different VAE variants. It also significantly reduces training time and GPU memory consumption due to its analytical computation. 
Note that position error is more dominant in determining the quality of reconstruction. Although some VAEs may achieve a lower curvature error, their reconstructed strands often appear noticeably different from the ground truth because they fail to preserve the overall shape (see Fig.~\ref{fig:strand-repr-comp}).
To further demonstrate the robustness of our strand representation, we trained all models on the curlier private dataset and evaluated them using the same $17$ publicly available hairstyles. Detailed experimental results can be found in Appendix~\ref{supp:strand-repr-exp-advanced}, which reveal a trend similar to those in Table~\ref{tab:strand-repr} (our PCA-based representation achieves the lowest position error and a comparatively low curvature error).

\begin{figure}[htbp]
	\centering

 \includegraphics[width=\textwidth, trim=0 0 0 165, clip]{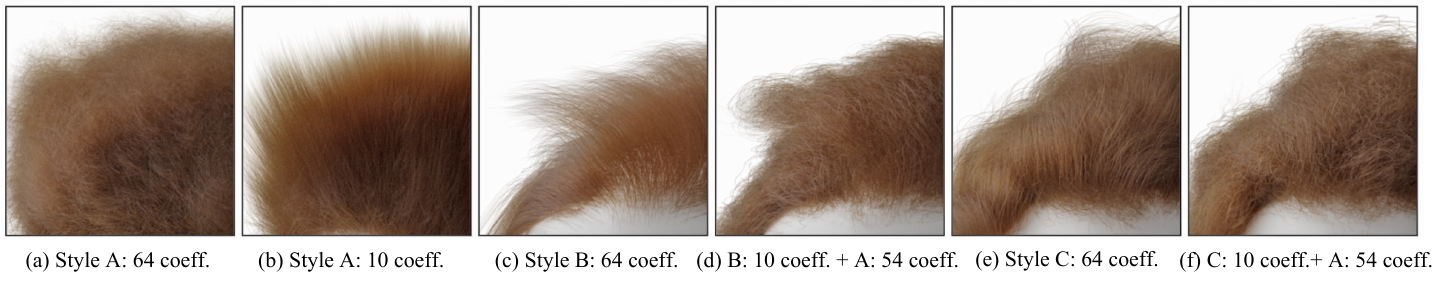}
 \vspace{-5mm}
\caption{
Examples of hairstyle editing with our PCA-based strand representation. (a) and (b) demonstrate hair smoothing by truncating the strand PCA coefficient from $64$ to $10$. (c) -- (f) show the detail transfer between different hairstyles.
}
\label{fig:strand-repr}
\vspace{-2mm}
\end{figure}

Additionally, with our proposed PCA-based strand representation, we can smooth a given hairstyle by truncating its strand PCA coefficients (Fig.~\ref{fig:strand-repr}a, b), or transfer hairstyle details by blending the low-rank and high-rank portions of PCA coefficients from different hairstyles (Fig.~\ref{fig:strand-repr}c -- f).

\subsection{Full Hair Representation}

\paragraph{3D Hair Parameterization}
\label{sec:perm-fitting}

To evaluate our proposed hairstyle representation, we first fit \textsc{Perm} parameters to target 3D hair models through differentiable optimization, which we term as \emph{3D hair parameterization}. Technical details are provided in Appendix~\ref{sup:perm-fitting}.
In Fig.~\ref{fig:perm-fitting} we present a subset of fitted results that illustrate our model's capability to accurately recover the given 3D hair models.
Even if the target hair has no guide strands, our model can generate reasonable guide strands to depict their overall shapes, obtaining directly from $\mathcal{S}\big(\mathcal{G}(\vec{\theta}^\ast)\big)$.

\paragraph{Comparison with GroomGen}
\begin{wrapfigure}[14]{r}{0.7\textwidth}
    \centering
    \vspace{-10pt}
    \addtolength{\tabcolsep}{-13pt}
    \begin{tabular}{cccc}
        \includegraphics[height=0.21\textwidth]{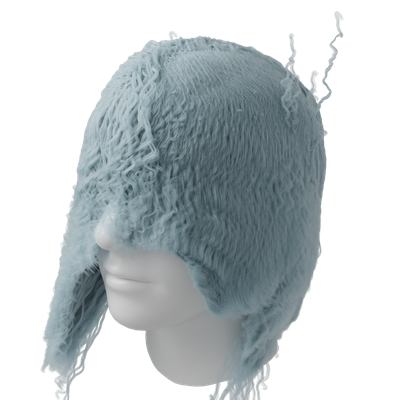} &
        \includegraphics[height=0.21\textwidth]{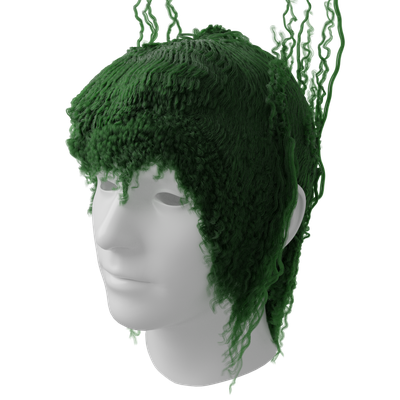} &
        \includegraphics[height=0.21\textwidth]{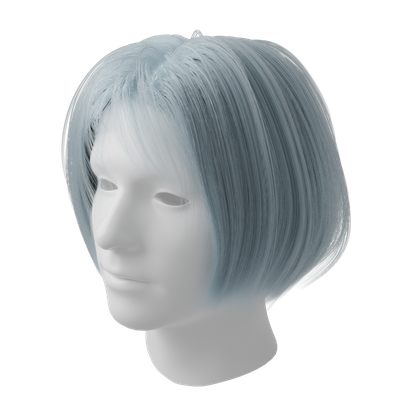} &
        \includegraphics[height=0.21\textwidth]{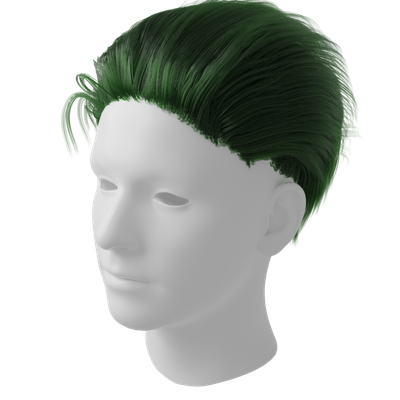} \\
        \multicolumn{2}{c}{\small GroomGen~\citep{zhou2023groomgen}} & \multicolumn{2}{c}{\small Ours} \\
    \end{tabular}
    \caption{Comparison with our implementation of GroomGen~\citep{zhou2023groomgen} on random hairstyle synthesis. Full results are available in Fig.~\ref{fig:perm-sampling}.
    }
    \label{fig:groomgen-comp}
    \vspace{-2mm}
\end{wrapfigure}

The most relevant previous work is GroomGen~\citep{zhou2023groomgen} that learns hierarchical representations of 3D hair.
As there is no publicly available GroomGen code, we implemented and trained it on the same augmented USC-HairSalon dataset. We also contacted the authors to obtain part of their official checkpoints to verify the correctness of our implementation. Detailed verification can be found in Appendix~\ref{supp:groomgen}, where we compared each network module with the available official checkpoints to demonstrate the correctness of our implementation. Through experiments with our implementation, we found the neural upsampler module in GroomGen, which is a GAN architecture, is very unstable to train, frequently leading to collapsed results.
In Fig.~\ref{fig:groomgen-comp} we visualize randomly synthesized hairstyles by sampling our parameter space and GroomGen's latent space with the same Gaussian noise, demonstrating their collapsed output with weird shapes and curls. We further conducted a quantitative evaluation of reconstruction errors on the $17$ testing hairstyles, where our model achieves a position error of $1.658$ and curvature error of $0.769$, both of which are lower than GroomGen's corresponding values (position error $2.570$, curvature error $6.199$).
During quantitative evaluation, we also found that $100$ points per strand are insufficient to faithfully represent long kinky hairstyles. Examples are provided in Fig.~\ref{fig:groomgen-verify}.
\begin{figure}[ht]
    \centering
    \includegraphics[width=\textwidth]{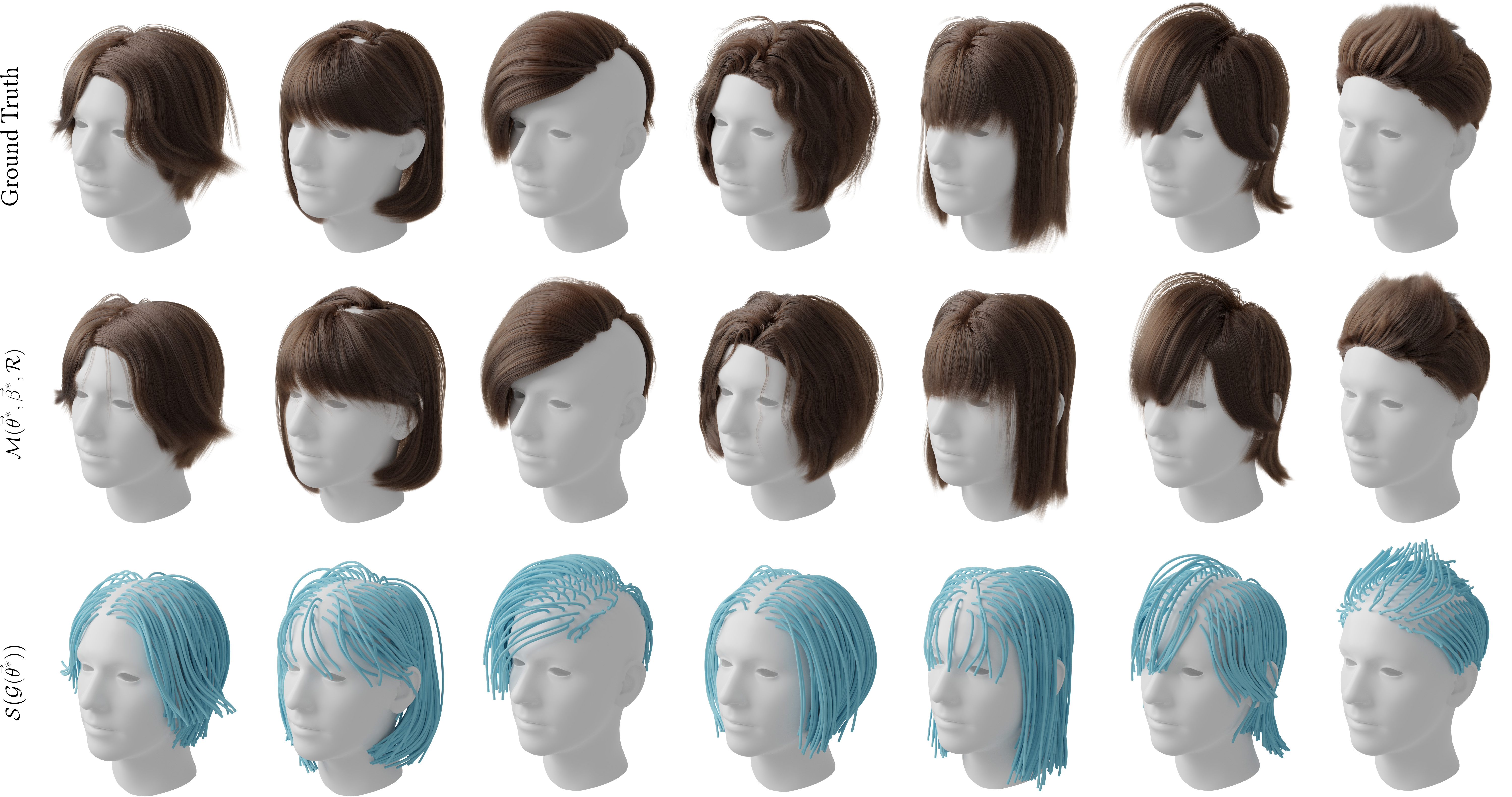}
    \caption{A subset of 3D hair models fitted by \textsc{Perm}.}
    \label{fig:perm-fitting}
    \vspace{-2mm}
\end{figure}

\section{Applications}

\paragraph{Single-view Hair Reconstruction}

\begin{wrapfigure}[18]{r}{0.45\textwidth}
    \centering
    \vspace{-10pt}
    \addtolength{\tabcolsep}{-5pt}
    \begin{tabular}{ccc}
     \includegraphics[width=0.32\linewidth]{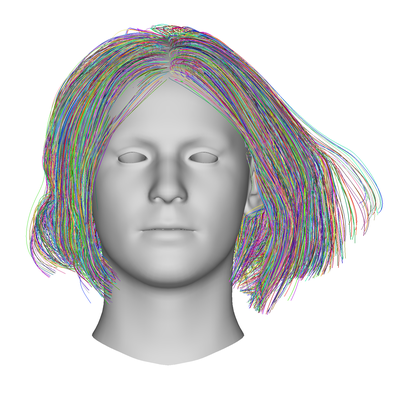} &
     \includegraphics[width=0.32\linewidth]{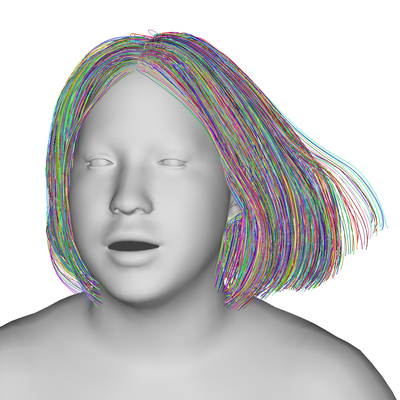} &
     \includegraphics[width=0.32\linewidth]{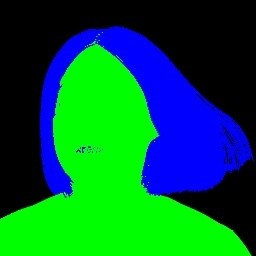} \\
     \tiny $\mathcal{M}(\vec{\theta}, \vec{\beta}, \mathcal{R})$ &
     \tiny $\mathcal{T}\big(\mathcal{M}(\vec{\theta}, \vec{\beta}, \mathcal{R})\big)$ &
     \tiny Hair Mask $\mathbf{M}_{\text{render}}$ \\
     \addlinespace[4pt]
     \includegraphics[width=0.32\linewidth]{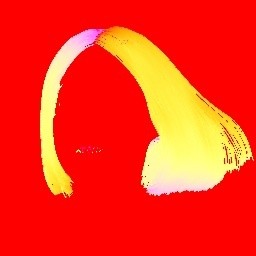} &
     \includegraphics[width=0.32\linewidth]{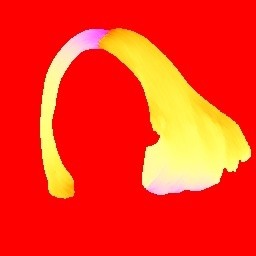} &
     \includegraphics[width=0.32\linewidth]{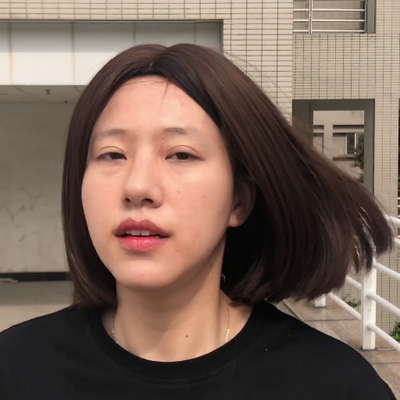} \\
     \tiny Strand Map $\mathbf{O}_{\text{render}}$ &
     \tiny GT Strand Map $\mathbf{O}_{\text{gt}}$&
     \tiny Input Image \\
    \end{tabular}
    \caption{Illustration of our differentiable rendering pipeline for single-view hair reconstruction.}
    \label{fig:single-view-pipe}
    \vspace{-2mm}
\end{wrapfigure}

With 2D observations as input, we can reconstruct 3D hair by optimizing \textsc{Perm} parameters to minimize the energy of generated hair when projected to the 2D images. To fit 3D hair to 2D supervisions, we design a differentiable rendering pipeline, as illustrated in Fig.~\ref{fig:single-view-pipe}.
In the pipeline, we pre-compute the transformation between our in-house head mesh and SMPL-X~\citep{pavlakos2019expressive}, denoted as the function $\mathcal{T}$, thus placing strand polylines generated from \textsc{Perm} onto SMPL-X.
We then attach a cylinder mesh onto each strand segment, as well as computing the 3D orientation of each vertex as the per-vertex feature.
These features are projected and rendered using Nvdiffrast~\citep{Laine2020diffrast} with the estimated camera parameters, thereby obtaining the rendered hair mask $\mathbf{M}_{\text{render}}$ and strand map $\mathbf{O}_{\text{render}}$.
By computing the pixel-wise mask loss and strand map loss similar to HairStep~\citep{zheng2023hair}, gradients are back-propagated to optimize $\vec{\theta}$ and $\vec{\beta}$ with decoupled weight decay regularization~\citep{loshchilov2017decoupled}. A penetration loss is applied as well to penalize hair intersecting the body geometry.

\begin{figure}[ht]
    \centering
    \addtolength{\tabcolsep}{-5.1pt}
    \begin{tabular}{ccccccc}
     \raisebox{28pt}{\rotatebox[origin=c]{90}{\small Input Images}} &
     \includegraphics[width=0.158\textwidth]{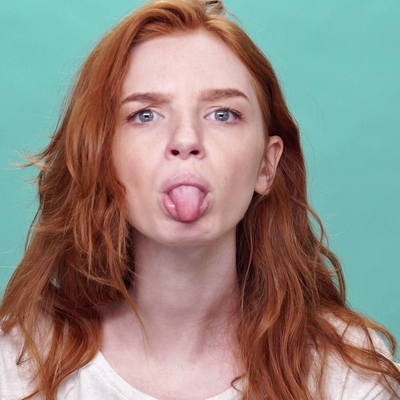} &
     \includegraphics[width=0.158\textwidth]{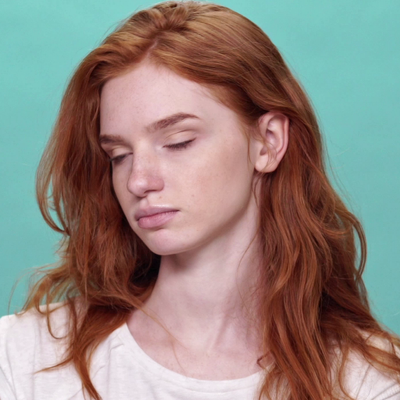} &
     \includegraphics[width=0.158\textwidth]{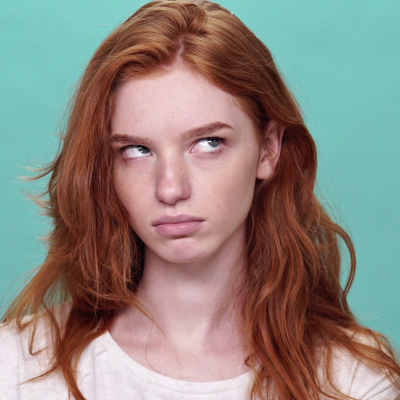} &
     \includegraphics[width=0.158\textwidth]{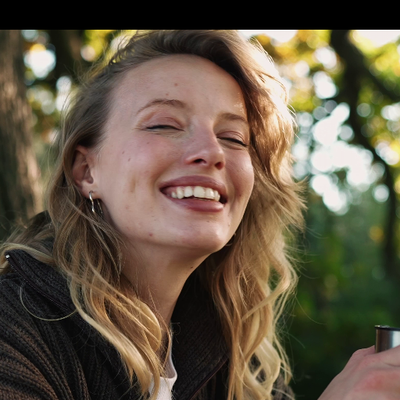} &
     \includegraphics[width=0.158\textwidth]{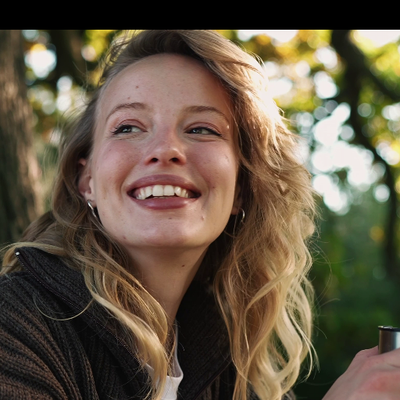} &
     \includegraphics[width=0.158\textwidth]{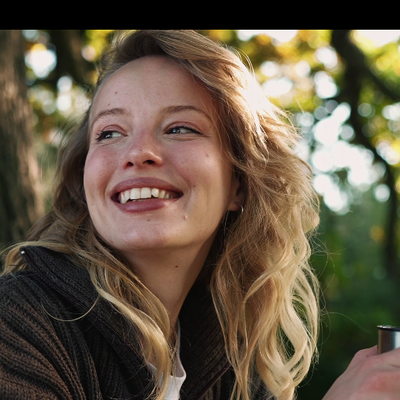} \\
     \addlinespace[-1pt]
     \raisebox{27pt}{\rotatebox[origin=c]{90}{\small Ours}} &
     \includegraphics[width=0.158\textwidth]{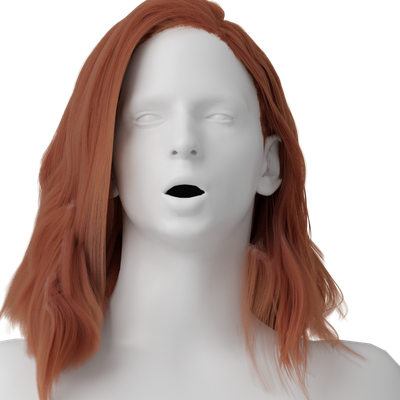} &
     \includegraphics[width=0.158\textwidth]{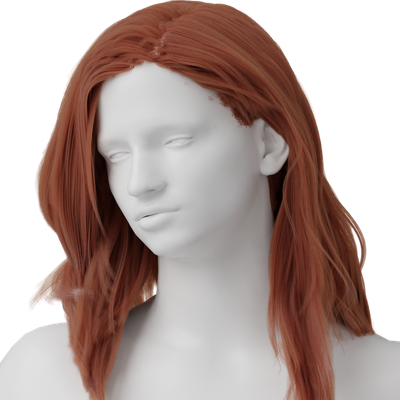} &
     \includegraphics[width=0.158\textwidth]{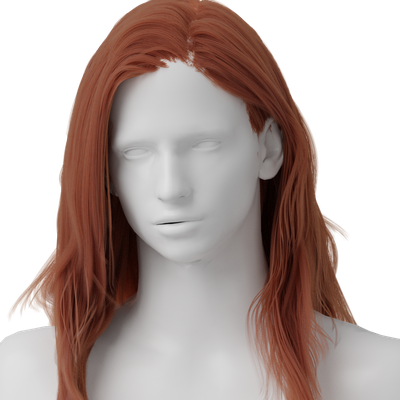} &
     \includegraphics[width=0.158\textwidth]{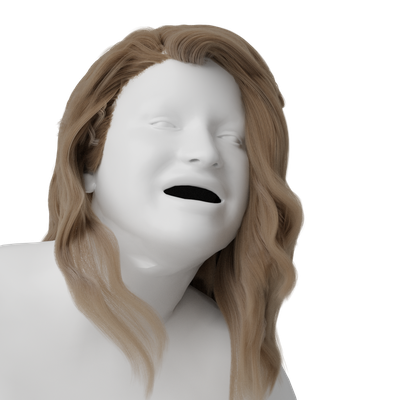} &
     \includegraphics[width=0.158\textwidth]{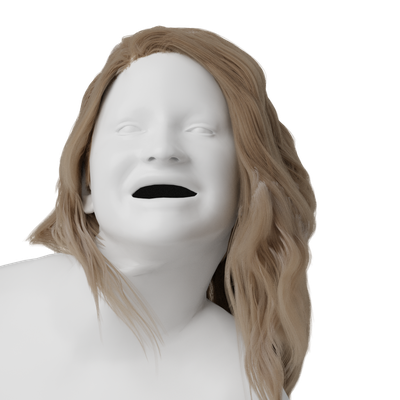} &
     \includegraphics[width=0.158\textwidth]{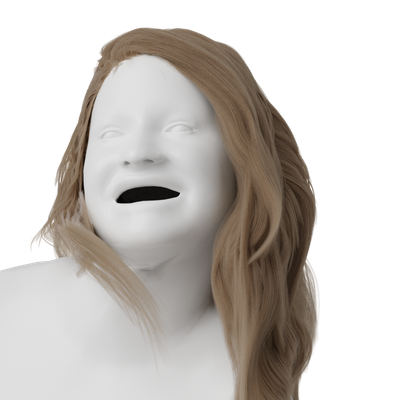} \\
     \addlinespace[-1pt]
     \raisebox{27pt}{\rotatebox[origin=c]{90}{\small HairStep}} &
     \includegraphics[width=0.158\textwidth]{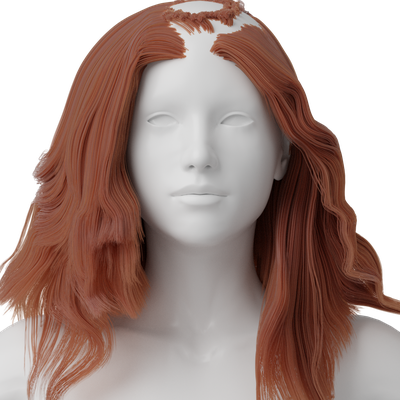} &
     \includegraphics[width=0.158\textwidth]{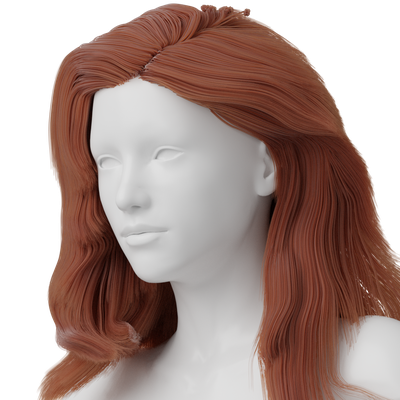} &
     \includegraphics[width=0.158\textwidth]{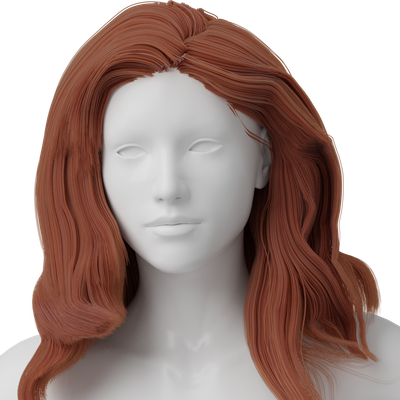} &
     \includegraphics[width=0.158\textwidth]{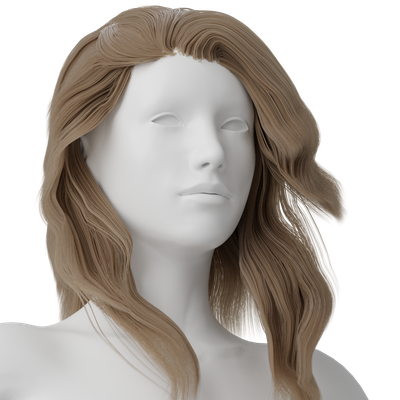} &
     \includegraphics[width=0.158\textwidth]{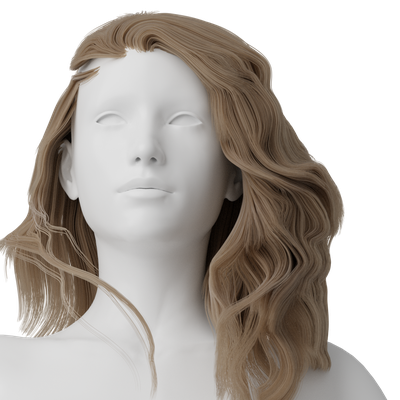} &
     \includegraphics[width=0.158\textwidth]{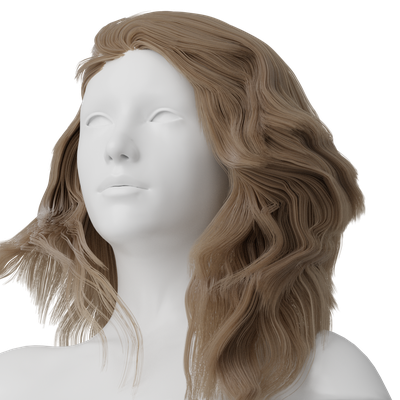} \\
    \end{tabular}
    \caption{Single-view hair reconstruction on images with curly hair and tilted head poses, with comparison to HairStep~\citep{zheng2023hair}. More examples are available in Fig.~\ref{fig:single-view-video}.}
    \label{fig:single-view-supp}
    \vspace{-2mm}
\end{figure}

In Fig.~\ref{fig:single-view-supp} and Fig.~\ref{fig:single-view-video}, we qualitatively compare our single-view reconstruction results with HairStep~\citep{zheng2023hair}, the state-of-the-art open-source method for single-view hair reconstruction. 
With \textsc{Perm} as a robust prior to ensure the generation quality, our method not only excels at reconstructing the global structure of the reference hairstyle but also better preserves local curl patterns compared to HairStep.
With the introduction of penetration loss, our algorithm effectively handles input images with large and tilted head poses, avoiding outputs with unintended bald areas, an artifact we occasionally observed in HairStep.
In Fig.~\ref{fig:haar} we show the single-view reconstruction results of HAAR~\citep{sklyarova2023haar}, a recent text-conditioned hair generation method. However, since textual descriptions cannot capture the intricate details of a hairstyle, its output can only produce a rough approximation.

\paragraph{\textsc{Perm} Parameter Editing}

As we introduce separate parameters to control the global hair shape and local strand details, we can edit the hairstyle with varying granularity by either swapping the parameters from different hairstyles or interpolating the corresponding parameters.
In the last row of Fig.~\ref{fig:single-view-video}, we edit our reconstructed hairstyles by swapping their $\vec{\beta}$ parameters from a wavy reference hairstyle, thereby altering their curl patterns while maintaining a similar haircut. More examples of hairstyle interpolation are provided in Appendix~\ref{sup:hair-interp}, with comparisons to \citep{weng2013hair} and \citep{zhou2018hairnet}.

\paragraph{Hair-conditioned Image Generation}
Latest text-to-image (T2I) models (e.g., Adobe Firefly~\citep{firefly}) can generate high-quality portrait images. However, the text embedding of simple prompts like \textit{``wavy and short hair''} is not precise enough to represent a specific hairstyle. To tackle this issue, we show the use of \textsc{Perm} in facilitating conditional image generation. Specifically, We could use \textsc{Perm} to sample and edit hairstyles in 3D, or reconstruct 3D hairstyles from images, and then feed the depth and edge information extracted from the hair geometry to the T2I models to generate the final images. As shown in Fig.~\ref{fig:firefly-eval}, pure text prompts often lead to images with different hairstyles, while combined with the input hair reference, the generated images show a much more consistent hairstyle. More examples are available in Fig.~\ref{fig:firefly}.

\section{Conclusion}
\label{sec:conclusion}

We present \textsc{Perm}, a parametric hair representation supporting independent control over the global hair structure and local curl patterns.
While our approach demonstrates strong performance when trained on 3D hair models, exploring methods to develop a 3D generative model for hair from 2D in-the-wild images remains a compelling direction for future research. 
Besides, since our $uv$-based hair representation is compatible with the existing diffusion pipeline, with more data available, it would be interesting to train a 3D diffusion model for full-head synthesis with controllable hair modeling.
Last but not least, aligning with trends in other 2D or 3D generative tasks, controllable 3D hair synthesis with multi-modal input signals presents another exciting avenue for future work in this domain.

\section*{Acknowledgments}
The authors would like to thank Yujian Zheng for providing the source code and model of HairStep, and for discussing the HairStep results in the paper with us. We also extend our sincere gratitude to Yao Feng for assisting us in setting up DELTA for face and shoulder reconstruction, and Hao Li for contributing a selection of the realistic head and hair assets used in this research.

\bibliography{main}

\begin{thebibliography}{60}
\providecommand{\natexlab}[1]{#1}
\providecommand{\url}[1]{\texttt{#1}}
\expandafter\ifx\csname urlstyle\endcsname\relax
  \providecommand{\doi}[1]{doi: #1}\else
  \providecommand{\doi}{doi: \begingroup \urlstyle{rm}\Url}\fi

\bibitem[Abdal et~al.(2019)Abdal, Qin, and Wonka]{Abdal_2019_ICCV}
Rameen Abdal, Yipeng Qin, and Peter Wonka.
\newblock Image2stylegan: How to embed images into the stylegan latent space?
\newblock In \emph{Proceedings of the IEEE/CVF International Conference on Computer Vision (ICCV)}, October 2019.

\bibitem[Adobe(2024)]{firefly}
Adobe.
\newblock Firefly, 2024.
\newblock https://www.adobe.com/products/firefly.html.

\bibitem[Autodesk(2024)]{maya}
Autodesk.
\newblock Maya, 2024.
\newblock https://www.autodesk.com/products/maya/.

\bibitem[Blanz \& Vetter(1999)Blanz and Vetter]{blanz19993dmm}
Volker Blanz and Thomas Vetter.
\newblock A morphable model for the synthesis of {3D} faces.
\newblock In \emph{Proceedings of the 26th annual conference on Computer graphics and interactive techniques}, pp.\  187--194, 1999.

\bibitem[Chai et~al.(2012)Chai, Wang, Weng, Yu, Guo, and Zhou]{chai2012single}
Menglei Chai, Lvdi Wang, Yanlin Weng, Yizhou Yu, Baining Guo, and Kun Zhou.
\newblock Single-view hair modeling for portrait manipulation.
\newblock \emph{ACM Transactions on Graphics (TOG)}, 31\penalty0 (4):\penalty0 1--8, 2012.

\bibitem[Chan et~al.(2022)Chan, Lin, Chan, Nagano, Pan, Mello, Gallo, Guibas, Tremblay, Khamis, Karras, and Wetzstein]{Chan2022eg3d}
Eric~R. Chan, Connor~Z. Lin, Matthew~A. Chan, Koki Nagano, Boxiao Pan, Shalini~De Mello, Orazio Gallo, Leonidas Guibas, Jonathan Tremblay, Sameh Khamis, Tero Karras, and Gordon Wetzstein.
\newblock Efficient geometry-aware {3D} generative adversarial networks.
\newblock In \emph{CVPR}, 2022.

\bibitem[Cignoni et~al.(2008)Cignoni, Callieri, Corsini, Dellepiane, Ganovelli, and Ranzuglia]{meshlab}
Paolo Cignoni, Marco Callieri, Massimiliano Corsini, Matteo Dellepiane, Fabio Ganovelli, and Guido Ranzuglia.
\newblock {MeshLab: an Open-Source Mesh Processing Tool}.
\newblock In Vittorio Scarano, Rosario~De Chiara, and Ugo Erra (eds.), \emph{Eurographics Italian Chapter Conference}. The Eurographics Association, 2008.
\newblock ISBN 978-3-905673-68-5.
\newblock \doi{10.2312/LocalChapterEvents/ItalChap/ItalianChapConf2008/129-136}.

\bibitem[De~La~Mettrie et~al.(2007)De~La~Mettrie, Saint-L{\'e}ger, Loussouarn, Garcel, Porter, and Langaney]{de2007shape}
Roland De~La~Mettrie, Didier Saint-L{\'e}ger, Genevi{\`e}ve Loussouarn, Anne-Lise Garcel, Crystal Porter, and Andr{\'e} Langaney.
\newblock Shape variability and classification of human hair: a worldwide approach.
\newblock \emph{Human Biology}, 79\penalty0 (3):\penalty0 265--281, 2007.

\bibitem[Goodfellow et~al.(2014)Goodfellow, Pouget-Abadie, Mirza, Xu, Warde-Farley, Ozair, Courville, and Bengio]{goodfellow2014generative}
Ian Goodfellow, Jean Pouget-Abadie, Mehdi Mirza, Bing Xu, David Warde-Farley, Sherjil Ozair, Aaron Courville, and Yoshua Bengio.
\newblock Generative adversarial nets.
\newblock \emph{Advances in neural information processing systems}, 27, 2014.

\bibitem[Herrera et~al.(2024)Herrera, Zhou, Sun, Shu, He, Pirk, and Michels]{alejandro2024ams}
Jorge Alejandro~Amador Herrera, Yi~Zhou, Xin Sun, Zhixin Shu, Chengan He, S\"{o}ren Pirk, and Dominik~L. Michels.
\newblock Augmented mass-spring model for real-time dense hair simulation.
\newblock \emph{arXiv preprint arXiv:2412.17144}, 2024.

\bibitem[Ho et~al.(2020)Ho, Jain, and Abbeel]{ho2020denoising}
Jonathan Ho, Ajay Jain, and Pieter Abbeel.
\newblock Denoising diffusion probabilistic models.
\newblock \emph{Advances in neural information processing systems}, 33:\penalty0 6840--6851, 2020.

\bibitem[Hoover et~al.(2023)Hoover, Alhajj, and Flores]{hoover2023physiology}
Ezra Hoover, Mandy Alhajj, and Jose~L Flores.
\newblock Physiology, hair.
\newblock In \emph{StatPearls [Internet]}. StatPearls Publishing, 2023.

\bibitem[Hu et~al.(2018)Hu, Shen, and Sun]{8578843}
Jie Hu, Li~Shen, and Gang Sun.
\newblock Squeeze-and-excitation networks.
\newblock In \emph{2018 IEEE/CVF Conference on Computer Vision and Pattern Recognition}, pp.\  7132--7141, 2018.
\newblock \doi{10.1109/CVPR.2018.00745}.

\bibitem[Hu et~al.(2015)Hu, Ma, Luo, and Li]{hu2015single}
Liwen Hu, Chongyang Ma, Linjie Luo, and Hao Li.
\newblock Single-view hair modeling using a hairstyle database.
\newblock \emph{ACM Transactions on Graphics (Proceedings SIGGRAPH 2015)}, 34\penalty0 (4), July 2015.

\bibitem[Karras et~al.(2019)Karras, Laine, and Aila]{karras2019style}
Tero Karras, Samuli Laine, and Timo Aila.
\newblock A style-based generator architecture for generative adversarial networks.
\newblock In \emph{Proceedings of the IEEE/CVF conference on computer vision and pattern recognition}, pp.\  4401--4410, 2019.

\bibitem[Karras et~al.(2020)Karras, Laine, Aittala, Hellsten, Lehtinen, and Aila]{Karras2019stylegan2}
Tero Karras, Samuli Laine, Miika Aittala, Janne Hellsten, Jaakko Lehtinen, and Timo Aila.
\newblock Analyzing and improving the image quality of {StyleGAN}.
\newblock In \emph{Proc. CVPR}, 2020.

\bibitem[Kerbl et~al.(2023)Kerbl, Kopanas, Leimk{\"u}hler, and Drettakis]{kerbl3Dgaussians}
Bernhard Kerbl, Georgios Kopanas, Thomas Leimk{\"u}hler, and George Drettakis.
\newblock 3d gaussian splatting for real-time radiance field rendering.
\newblock \emph{ACM Transactions on Graphics}, 42\penalty0 (4), July 2023.

\bibitem[Kingma \& Ba(2014)Kingma and Ba]{kingma2014adam}
Diederik~P Kingma and Jimmy Ba.
\newblock Adam: A method for stochastic optimization.
\newblock \emph{arXiv preprint arXiv:1412.6980}, 2014.

\bibitem[Kingma \& Welling(2013)Kingma and Welling]{kingma2013auto}
Diederik~P Kingma and Max Welling.
\newblock Auto-encoding variational bayes.
\newblock \emph{arXiv preprint arXiv:1312.6114}, 2013.

\bibitem[Kuang et~al.(2022)Kuang, Chen, Fu, Zhou, and Zheng]{kuang2022deepmvs}
Zhiyi Kuang, Yiyang Chen, Hongbo Fu, Kun Zhou, and Youyi Zheng.
\newblock Deepmvshair: Deep hair modeling from sparse views.
\newblock In \emph{SIGGRAPH Asia 2022 Conference Papers}, 2022.

\bibitem[Laine et~al.(2020)Laine, Hellsten, Karras, Seol, Lehtinen, and Aila]{Laine2020diffrast}
Samuli Laine, Janne Hellsten, Tero Karras, Yeongho Seol, Jaakko Lehtinen, and Timo Aila.
\newblock Modular primitives for high-performance differentiable rendering.
\newblock \emph{ACM Transactions on Graphics}, 39\penalty0 (6), 2020.

\bibitem[Li et~al.(2017)Li, Bolkart, Black, Li, and Romero]{FLAME:SiggraphAsia2017}
Tianye Li, Timo Bolkart, Michael.~J. Black, Hao Li, and Javier Romero.
\newblock Learning a model of facial shape and expression from {4D} scans.
\newblock \emph{ACM Transactions on Graphics, (Proc. SIGGRAPH Asia)}, 36\penalty0 (6):\penalty0 194:1--194:17, 2017.

\bibitem[Liu et~al.(2024)Liu, Li, Wu, and Lee]{liu2024visual}
Haotian Liu, Chunyuan Li, Qingyang Wu, and Yong~Jae Lee.
\newblock Visual instruction tuning.
\newblock \emph{Advances in neural information processing systems}, 36, 2024.

\bibitem[Loper et~al.(2015)Loper, Mahmood, Romero, Pons-Moll, and Black]{SMPL:2015}
Matthew Loper, Naureen Mahmood, Javier Romero, Gerard Pons-Moll, and Michael~J. Black.
\newblock {SMPL}: A skinned multi-person linear model.
\newblock \emph{ACM Trans. Graphics (Proc. SIGGRAPH Asia)}, 34\penalty0 (6):\penalty0 248:1--248:16, October 2015.

\bibitem[Loshchilov \& Hutter(2017)Loshchilov and Hutter]{loshchilov2017decoupled}
Ilya Loshchilov and Frank Hutter.
\newblock Decoupled weight decay regularization.
\newblock \emph{arXiv preprint arXiv:1711.05101}, 2017.

\bibitem[Loussouarn et~al.(2007)Loussouarn, Garcel, Lozano, Collaudin, Porter, Panhard, Saint-L{\'e}ger, and De~La~Mettrie]{loussouarn2007worldwide}
Genevi{\`e}ve Loussouarn, Anne-Lise Garcel, Isabelle Lozano, Catherine Collaudin, Crystal Porter, S{\'e}gol{\`e}ne Panhard, Didier Saint-L{\'e}ger, and Roland De~La~Mettrie.
\newblock Worldwide diversity of hair curliness: a new method of assessment.
\newblock \emph{International journal of dermatology}, 46:\penalty0 2--6, 2007.

\bibitem[Luo et~al.(2024)Luo, Ouyang, Zhao, Jiang, Zhang, Zhang, Yang, Xu, and Yu]{luo2024gaussianhair}
Haimin Luo, Min Ouyang, Zijun Zhao, Suyi Jiang, Longwen Zhang, Qixuan Zhang, Wei Yang, Lan Xu, and Jingyi Yu.
\newblock Gaussianhair: Hair modeling and rendering with light-aware gaussians.
\newblock \emph{arXiv preprint arXiv:2402.10483}, 2024.

\bibitem[Luo et~al.(2013)Luo, Li, and Rusinkiewicz]{luo2013structure}
Linjie Luo, Hao Li, and Szymon Rusinkiewicz.
\newblock Structure-aware hair capture.
\newblock \emph{ACM Transactions on Graphics (TOG)}, 32\penalty0 (4):\penalty0 1--12, 2013.

\bibitem[Mahmood et~al.(2019)Mahmood, Ghorbani, Troje, Pons-Moll, and Black]{AMASS:ICCV:2019}
Naureen Mahmood, Nima Ghorbani, Nikolaus~F. Troje, Gerard Pons-Moll, and Michael~J. Black.
\newblock {AMASS}: Archive of motion capture as surface shapes.
\newblock In \emph{International Conference on Computer Vision}, pp.\  5442--5451, October 2019.

\bibitem[Mehta et~al.(2021)Mehta, Gharbi, Barnes, Shechtman, Ramamoorthi, and Chandraker]{mehta2021modulated}
Ishit Mehta, Micha\"el Gharbi, Connelly Barnes, Eli Shechtman, Ravi Ramamoorthi, and Manmohan Chandraker.
\newblock Modulated periodic activations for generalizable local functional representations.
\newblock In \emph{Proceedings of the IEEE/CVF International Conference on Computer Vision (ICCV)}, pp.\  14214--14223, October 2021.

\bibitem[Mescheder et~al.(2018)Mescheder, Geiger, and Nowozin]{mescheder2018training}
Lars Mescheder, Andreas Geiger, and Sebastian Nowozin.
\newblock Which training methods for gans do actually converge?
\newblock In \emph{International conference on machine learning}, pp.\  3481--3490. PMLR, 2018.

\bibitem[Nam et~al.(2019)Nam, Wu, Kim, and Sheikh]{nam2019lpmvs}
Giljoo Nam, Chenglei Wu, Min~H. Kim, and Yaser Sheikh.
\newblock Strand-accurate multi-view hair capture.
\newblock In \emph{2019 IEEE/CVF Conference on Computer Vision and Pattern Recognition (CVPR)}, pp.\  155--164, 2019.
\newblock \doi{10.1109/CVPR.2019.00024}.

\bibitem[Pavlakos et~al.(2019)Pavlakos, Choutas, Ghorbani, Bolkart, Osman, Tzionas, and Black]{pavlakos2019expressive}
Georgios Pavlakos, Vasileios Choutas, Nima Ghorbani, Timo Bolkart, Ahmed~AA Osman, Dimitrios Tzionas, and Michael~J Black.
\newblock Expressive body capture: 3d hands, face, and body from a single image.
\newblock In \emph{Proceedings of the IEEE/CVF conference on computer vision and pattern recognition}, pp.\  10975--10985, 2019.

\bibitem[Richardson et~al.(2021)Richardson, Alaluf, Patashnik, Nitzan, Azar, Shapiro, and Cohen-Or]{richardson2021encoding}
Elad Richardson, Yuval Alaluf, Or~Patashnik, Yotam Nitzan, Yaniv Azar, Stav Shapiro, and Daniel Cohen-Or.
\newblock Encoding in style: a stylegan encoder for image-to-image translation.
\newblock In \emph{IEEE/CVF Conference on Computer Vision and Pattern Recognition (CVPR)}, June 2021.

\bibitem[Romero et~al.(2017)Romero, Tzionas, and Black]{MANO:SIGGRAPHASIA:2017}
Javier Romero, Dimitrios Tzionas, and Michael~J. Black.
\newblock Embodied hands: Modeling and capturing hands and bodies together.
\newblock \emph{ACM Transactions on Graphics, (Proc. SIGGRAPH Asia)}, 36\penalty0 (6), November 2017.

\bibitem[Ronneberger et~al.(2015)Ronneberger, Fischer, and Brox]{ronneberger2015u}
Olaf Ronneberger, Philipp Fischer, and Thomas Brox.
\newblock U-net: Convolutional networks for biomedical image segmentation.
\newblock In \emph{Medical Image Computing and Computer-Assisted Intervention--MICCAI 2015: 18th International Conference, Munich, Germany, October 5-9, 2015, Proceedings, Part III 18}, pp.\  234--241. Springer, 2015.

\bibitem[Rosu et~al.(2022)Rosu, Saito, Wang, Wu, Behnke, and Nam]{rosu2022neuralstrands}
Radu~Alexandru Rosu, Shunsuke Saito, Ziyan Wang, Chenglei Wu, Sven Behnke, and Giljoo Nam.
\newblock Neural strands: Learning hair geometry and appearance from multi-view images.
\newblock \emph{ECCV}, 2022.

\bibitem[Saito et~al.(2018)Saito, Hu, Ma, Ibayashi, Luo, and Li]{saito2018_hairvae}
Shunsuke Saito, Liwen Hu, Chongyang Ma, Hikaru Ibayashi, Linjie Luo, and Hao Li.
\newblock 3d hair synthesis using volumetric variational autoencoders.
\newblock \emph{ACM Trans. Graph.}, 37\penalty0 (6), dec 2018.
\newblock ISSN 0730-0301.

\bibitem[Shen et~al.(2020)Shen, Zhang, Fu, Zhou, and Zheng]{shen2020deepsketchhair}
Yuefan Shen, Changgeng Zhang, Hongbo Fu, Kun Zhou, and Youyi Zheng.
\newblock Deepsketchhair: Deep sketch-based 3d hair modeling.
\newblock \emph{IEEE transactions on visualization and computer graphics}, 27\penalty0 (7):\penalty0 3250--3263, 2020.

\bibitem[Shen et~al.(2023)Shen, Saito, Wang, Maury, Wu, Hodgins, Zheng, and Nam]{shen2023CT2Hair}
Yuefan Shen, Shunsuke Saito, Ziyan Wang, Olivier Maury, Chenglei Wu, Jessica Hodgins, Youyi Zheng, and Giljoo Nam.
\newblock Ct2hair: High-fidelity 3d hair modeling using computed tomography.
\newblock \emph{ACM Transactions on Graphics}, 42\penalty0 (4):\penalty0 1--13, 2023.

\bibitem[Sklyarova et~al.(2023{\natexlab{a}})Sklyarova, Chelishev, Dogaru, Medvedev, Lempitsky, and Zakharov]{sklyarova2023neural_haircut}
Vanessa Sklyarova, Jenya Chelishev, Andreea Dogaru, Igor Medvedev, Victor Lempitsky, and Egor Zakharov.
\newblock Neural haircut: Prior-guided strand-based hair reconstruction.
\newblock In \emph{Proceedings of IEEE International Conference on Computer Vision (ICCV)}, 2023{\natexlab{a}}.

\bibitem[Sklyarova et~al.(2023{\natexlab{b}})Sklyarova, Zakharov, Hilliges, Black, and Thies]{sklyarova2023haar}
Vanessa Sklyarova, Egor Zakharov, Otmar Hilliges, Michael~J Black, and Justus Thies.
\newblock Haar: Text-conditioned generative model of 3d strand-based human hairstyles.
\newblock \emph{ArXiv}, Dec 2023{\natexlab{b}}.

\bibitem[Takimoto et~al.(2024)Takimoto, Takehara, Sato, Zhu, and Zheng]{takimoto2024dr}
Yusuke Takimoto, Hikari Takehara, Hiroyuki Sato, Zihao Zhu, and Bo~Zheng.
\newblock Dr. hair: Reconstructing scalp-connected hair strands without pre-training via differentiable rendering of line segments.
\newblock \emph{arXiv preprint arXiv:2403.17496}, 2024.

\bibitem[Tewari et~al.(2020)Tewari, Fried, Thies, Sitzmann, Lombardi, Sunkavalli, Martin-Brualla, Simon, Saragih, Nie{\ss}ner, et~al.]{tewari2020state}
Ayush Tewari, Ohad Fried, Justus Thies, Vincent Sitzmann, Stephen Lombardi, Kalyan Sunkavalli, Ricardo Martin-Brualla, Tomas Simon, Jason Saragih, Matthias Nie{\ss}ner, et~al.
\newblock State of the art on neural rendering.
\newblock In \emph{Computer Graphics Forum}, volume~39, pp.\  701--727. Wiley Online Library, 2020.

\bibitem[Vaswani et~al.(2017)Vaswani, Shazeer, Parmar, Uszkoreit, Jones, Gomez, Kaiser, and Polosukhin]{vaswani2017attention}
Ashish Vaswani, Noam Shazeer, Niki Parmar, Jakob Uszkoreit, Llion Jones, Aidan~N Gomez, \L~ukasz Kaiser, and Illia Polosukhin.
\newblock Attention is all you need.
\newblock In \emph{Advances in Neural Information Processing Systems}, volume~30, 2017.

\bibitem[Wang et~al.(2009)Wang, Yu, Zhou, and Guo]{wang2009hair}
Lvdi Wang, Yizhou Yu, Kun Zhou, and Baining Guo.
\newblock Example-based hair geometry synthesis.
\newblock In \emph{ACM SIGGRAPH 2009 Papers}, SIGGRAPH '09, New York, NY, USA, 2009. Association for Computing Machinery.
\newblock ISBN 9781605587264.

\bibitem[Wang et~al.(2022)Wang, Nam, Stuyck, Lombardi, Zollh{\"o}fer, Hodgins, and Lassner]{wang2022hvh}
Ziyan Wang, Giljoo Nam, Tuur Stuyck, Stephen Lombardi, Michael Zollh{\"o}fer, Jessica Hodgins, and Christoph Lassner.
\newblock Hvh: Learning a hybrid neural volumetric representation for dynamic hair performance capture.
\newblock In \emph{Proceedings of the IEEE/CVF Conference on Computer Vision and Pattern Recognition}, pp.\  6143--6154, 2022.

\bibitem[Wang et~al.(2023{\natexlab{a}})Wang, Nam, Bozic, Cao, Saragih, Zollhoefer, and Hodgins]{wang2023local}
Ziyan Wang, Giljoo Nam, Aljaz Bozic, Chen Cao, Jason Saragih, Michael Zollhoefer, and Jessica Hodgins.
\newblock A local appearance model for volumetric capture of diverse hairstyle.
\newblock \emph{arXiv preprint arXiv:2312.08679}, 2023{\natexlab{a}}.

\bibitem[Wang et~al.(2023{\natexlab{b}})Wang, Nam, Stuyck, Lombardi, Cao, Saragih, Zollh{\"o}fer, Hodgins, and Lassner]{wang2023neuwigs}
Ziyan Wang, Giljoo Nam, Tuur Stuyck, Stephen Lombardi, Chen Cao, Jason Saragih, Michael Zollh{\"o}fer, Jessica Hodgins, and Christoph Lassner.
\newblock Neuwigs: A neural dynamic model for volumetric hair capture and animation.
\newblock In \emph{Proceedings of the IEEE/CVF Conference on Computer Vision and Pattern Recognition}, pp.\  8641--8651, 2023{\natexlab{b}}.

\bibitem[Weng et~al.(2013)Weng, Wang, Li, Chai, and Zhou]{weng2013hair}
Yanlin Weng, Lvdi Wang, Xiao Li, Menglei Chai, and Kun Zhou.
\newblock Hair interpolation for portrait morphing.
\newblock In \emph{Computer Graphics Forum}, volume~32, pp.\  79--84, 2013.

\bibitem[Wu et~al.(2022)Wu, Ye, Yang, Fu, Zhou, and Zheng]{wu2022neuralhdhair}
Keyu Wu, Yifan Ye, Lingchen Yang, Hongbo Fu, Kun Zhou, and Youyi Zheng.
\newblock Neuralhdhair: Automatic high-fidelity hair modeling from a single image using implicit neural representations.
\newblock In \emph{Proceedings of the IEEE/CVF Conference on Computer Vision and Pattern Recognition}, pp.\  1526--1535, 2022.

\bibitem[Wu et~al.(2024)Wu, Yang, Kuang, Feng, Han, Shen, Fu, Zhou, and Zheng]{wu2024monohair}
Keyu Wu, Lingchen Yang, Zhiyi Kuang, Yao Feng, Xutao Han, Yuefan Shen, Hongbo Fu, Kun Zhou, and Youyi Zheng.
\newblock Monohair: High-fidelity hair modeling from a monocular video.
\newblock \emph{arXiv preprint arXiv:2403.18356}, 2024.

\bibitem[Yang et~al.(2019)Yang, Shi, Zheng, and Zhou]{yang2019dynamic}
Lingchen Yang, Zefeng Shi, Youyi Zheng, and Kun Zhou.
\newblock Dynamic hair modeling from monocular videos using deep neural networks.
\newblock \emph{ACM Transactions on Graphics (TOG)}, 38\penalty0 (6):\penalty0 1--12, 2019.

\bibitem[Zakharov et~al.(2024)Zakharov, Sklyarova, Black, Nam, Thies, and Hilliges]{zakharov2024gh}
Egor Zakharov, Vanessa Sklyarova, Michael~J Black, Giljoo Nam, Justus Thies, and Otmar Hilliges.
\newblock Human hair reconstruction with strand-aligned 3d gaussians.
\newblock \emph{ArXiv}, Sep 2024.

\bibitem[Zhang \& Zheng(2019)Zhang and Zheng]{zhang2019hair}
Meng Zhang and Youyi Zheng.
\newblock Hair-gan: Recovering 3d hair structure from a single image using generative adversarial networks.
\newblock \emph{Visual Informatics}, 3\penalty0 (2):\penalty0 102–112, 2019.

\bibitem[Zhang et~al.(2018)Zhang, Isola, Efros, Shechtman, and Wang]{zhang2018perceptual}
Richard Zhang, Phillip Isola, Alexei~A Efros, Eli Shechtman, and Oliver Wang.
\newblock The unreasonable effectiveness of deep features as a perceptual metric.
\newblock In \emph{CVPR}, 2018.

\bibitem[Zheng et~al.(2023)Zheng, Jin, Li, Huang, Ma, Cui, and Han]{zheng2023hair}
Yujian Zheng, Zirong Jin, Moran Li, Haibin Huang, Chongyang Ma, Shuguang Cui, and Xiaoguang Han.
\newblock Hairstep: Transfer synthetic to real using strand and depth maps for single-view 3d hair modeling.
\newblock In \emph{Proceedings of the IEEE/CVF Conference on Computer Vision and Pattern Recognition}, 2023.

\bibitem[Zhou et~al.(2018)Zhou, Hu, Xing, Chen, Kung, Tong, and Li]{zhou2018hairnet}
Yi~Zhou, Liwen Hu, Jun Xing, Weikai Chen, Han-Wei Kung, Xin Tong, and Hao Li.
\newblock Hairnet: Single-view hair reconstruction using convolutional neural networks.
\newblock In \emph{Proceedings of the European Conference on Computer Vision (ECCV)}, pp.\  235--251, 2018.

\bibitem[Zhou et~al.(2023)Zhou, Chai, Pepe, Gross, and Beeler]{zhou2023groomgen}
Yuxiao Zhou, Menglei Chai, Alessandro Pepe, Markus Gross, and Thabo Beeler.
\newblock Groomgen: A high-quality generative hair model using hierarchical latent representations.
\newblock \emph{arXiv preprint arXiv:2311.02062}, 2023.

\bibitem[Zhou et~al.(2024)Zhou, Chai, Wang, Winberg, Wood, Sarkar, Gross, and Beeler]{zhou2024groomcap}
Yuxiao Zhou, Menglei Chai, Daoye Wang, Sebastian Winberg, Erroll Wood, Kripasindhu Sarkar, Markus Gross, and Thabo Beeler.
\newblock Groomcap: High-fidelity prior-free hair capture.
\newblock \emph{arXiv preprint arXiv:2409.00831}, 2024.

\end{thebibliography}
\bibliographystyle{iclr2025_conference}

\clearpage
\appendix
\section{Formulation Details}

\subsection{Hair Geometry Textures}
\label{supp:hair-texture}

Unlike triangle meshes that share the same topology between different bodies and faces, hair tends to have a different number of strands for different hairstyles, making traditional PCA-based blend shapes impossible on these data. Although we can force different hairstyles to have the same number of strands by introducing a resampling step, the computed blend shapes would still have a fixed number of strands, which are less flexible in real applications. For flexibility, we store each hair model as a 2D texture map of strand PCA coefficients. With the scalp parameterization proposed by~\citet{wang2009hair}, we are able to unwarp the 3D scalp surface to a 2D $uv$ plane. However, naively storing strand PCA coefficients on the projected 2D root positions will cause two problems: (1) different strands may be projected to the same texel due to discreterization, thus causing collision problems; (2) some texels may receive no strands, thus leaving missing values in the projected textures.

To address these issues, we fit our geometry textures with two steps: first, we find the nearest 2D hair root for each texel, and store the corresponding strand PCA coefficients at that texel. For those texels whose distance to its nearest hair root is above a threshold $\epsilon=0.01$, we store a special vector that will be decoded to strands with zero length, and mark those texels as a baldness map $\mathbf{M}$~\citep{zhou2023groomgen}. This step ensures no missing values in the texture, which we denote as the initialized geometry texture $\mathbf{T}_{\text{init}}$. Both $\mathbf{M}$ and $\mathbf{T}_{\text{init}}$ have the resolution $256 \times 256$, which is empirically set considering the trade-off between size and expressiveness. To ensure that geometry textures can properly recover the original 3D hairstyle, we optimize them directly in the second step:
\begin{equation}
    \mathbf{T}^\ast \coloneqq \argmin_{\mathbf{T}}\mathcal{L}_{\text{geo}}\Big(\mathcal{S}\big(\text{Sample}(\mathcal{R}; \mathbf{T})\big)\Big),
\end{equation}
where we use loss $\mathcal{L}_{\text{geo}}$ to measure the reconstruction difference on strand geometry.

\begin{wrapfigure}[19]{r}{0.65\textwidth}
    \centering
    \vspace{-20pt}
    \addtolength{\tabcolsep}{-3pt}
    \begin{tabular}{ccc}
        \includegraphics[width=0.2\textwidth]{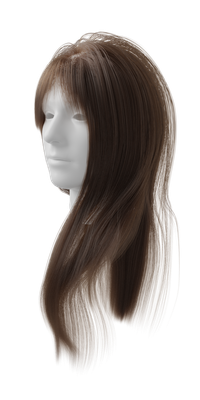} &
        \includegraphics[width=0.2\textwidth]{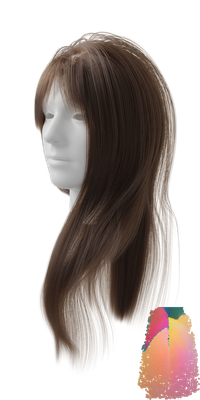} &
        \includegraphics[width=0.2\textwidth]{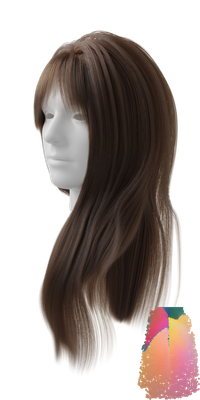} \\
        \small GT ($10k$ strands) & \small $10k$ samples & \small $100k$ samples \\
    \end{tabular}
    \caption{Illustration of fitted hair geometry textures and the corresponding 3D hairstyle. Here baldness maps are visualized as the alpha channel in the RGBA textures. 
    }
    \label{fig:hair-texture}
\end{wrapfigure}

We employ the Adam optimizer~\citep{kingma2014adam} with a learning rate of $0.001$. With $\mathbf{T}_{\text{init}}$ as initialization, our experiments show that the optimization process converges within $500$ iterations, and in Fig.~\ref{fig:hair-texture} we show an example of the original hairstyle, the fitted geometry texture, and the recovered 3D hairstyles by sampling and decoding from the texture with different numbers of roots. These fitted geometry textures finally form a unified representation across different hairstyles, which have the same size $ 256 \times 256 \times 64$ and allow for arbitrary sampling.

\subsection{Network Architecture}

\paragraph{StyleGAN2 Backbone}
Our StyleGAN2 backbone follows the official implementation of~\citep{Karras2019stylegan2}\footnote{\url{https://github.com/NVlabs/stylegan2-ada-pytorch}}, with a mapping network of $4$ hidden layers. We modify the output convolutions such that they produce a feature image of shape $32 \times 32 \times 10$. Subsequently, a small MLP decoder is employed to map the output features to $10$-dimensional strand PCA coefficients and a single scalar for guide mask. The MLP decoder consists of a single hidden layer of $64$ hidden units and uses the softplus activation function. Note that we do not utilize pre-trained StyleGAN2 checkpoints for our task, the entire module is trained from scratch.

\paragraph{U-Net Super Resolution}
Our U-Net module is implemented based on an unofficial online implementation\footnote{\url{https://github.com/milesial/Pytorch-UNet}}, which translates the bilinearly upsampled textures of shape $256 \times 256 \times 11$ to weight maps of shape $256 \times 256 \times 14$. The convolution layers progressively downsample the input to a shape of $16 \times 16 \times 512$, which is followed by $4$ bilinear upsampling and double convolution layers with skip connections to produce the weight map.

\paragraph{VAE}
In our VAE module, the encoder adopts a similar architecture to pSp~\citep{richardson2021encoding}, which contains $4$ IR-SE blocks~\citep{8578843} to extract a feature image of shape $4 \times 4 \times 512$. This feature image is then flattened and processed through a fully-connected layer to derive the mean and variance of $\vec{\beta}$ of $512$ dimensions. The decoder mirrors the StyleGAN2 generator but omits its mapping network. Its output size is modified to $256 \times 256 \times 54$, followed by the same MLP decoder that maps the output features to $54$-dimensional strand PCA coefficients and a single scalar for baldness map.

\section{Experiment Details}

\subsection{Datasets}
\label{supp:dataset}
We train \textsc{Perm} on USC-HairSalon~\citep{hu2015single}, which is a dataset comprising $343$ 3D hair models collected from online game communities. To increase diversity, we employ the data augmentation method proposed in HairNet~\citep{zhou2018hairnet}, where different hair models within the same style class are blended to produce novel hairstyles. The blended hairstyles are further augmented by horizontal flipping, resulting in a total of $21,054$ data samples used for training.

To assess the performance of our model, we compiled a dataset of 3D hair models from various publicly available resources, including CT2Hair~\citep{shen2023CT2Hair} ($10$ hairstyles), StructureAwareHair~\citep{luo2013structure} ($3$ hairstyles), and Cem Yuksel's website\footnote{\url{http://www.cemyuksel.com/research/hairmodels/}} (4 hairstyles). We preprocess these data to have the same number of points ($L=100$) on each strand, and register them onto the same head mesh. 

\subsection{Training Details}
We train our model and conduct all experiments on a desktop machine with an Intel\textsuperscript{\tiny\textregistered} Core\texttrademark\ i9-10850K CPU @ 3.60GHz, 64GB memory, and an NVIDIA RTX 3090 GPU. Our code is implemented with Python 3.9.18, PyTorch 1.11.0, and CUDA Toolkit 11.3.

In our model, each network module is trained separately using the Adam optimizer~\citep{kingma2014adam}. The StyleGAN2 backbone has a learning rate of $0.002$ for its generator and $0.001$ for its discriminator, leading to a stable training configuration in our case. The StyleGAN2 backbone is trained for $3,000$K images with a batch size of $4$, which takes around $1$ day on our machine. For both the U-Net and VAE, we set their learning rates to $0.002$ and train them for $2,000$K images with a batch size of $4$, each taking around $1$ day on our machine.

\subsection{Quantitative Metrics}
\label{supp:metrics}

To quantitatively measure the reconstruction capability of our model, we first report the mean \emph{position error} (pos. err.), which is essentially the average Euclidean distance between the corresponding points on the reconstructed strands and the ground truth. We further report the mean \emph{curvature error} (cur. err.) that measures the $L_1$ norm between the curvatures of reconstructed and ground truth strands, where the curvature is defined as the reciprocal of the circumradius of $3$ consecutive points $\mathbf{p}_{i-1}$, $\mathbf{p}_{i}$, and $\mathbf{p}_{i+1}$ on the strand, which can be computed as:
\begin{equation}
    \text{cur}(\mathbf{p}_i) = \frac{2\|(\mathbf{p}_{i-1} - \mathbf{p}_{i+1}) \times (\mathbf{p}_i - \mathbf{p}_{i+1})\|}{\|\mathbf{p}_{i-1} - \mathbf{p}_{i+1}\|\cdot\|\mathbf{p}_i - \mathbf{p}_{i+1}\|\cdot\|\mathbf{p}_{i-1} - \mathbf{p}_i\|}.
\end{equation}

\subsection{PCA-based Strand Representation}
\label{supp:strand-repr-exp-advanced}
\begin{figure}[ht]
\centering
\begin{minipage}[t]{.32\textwidth}
    \centering
    \includegraphics[width=\columnwidth]{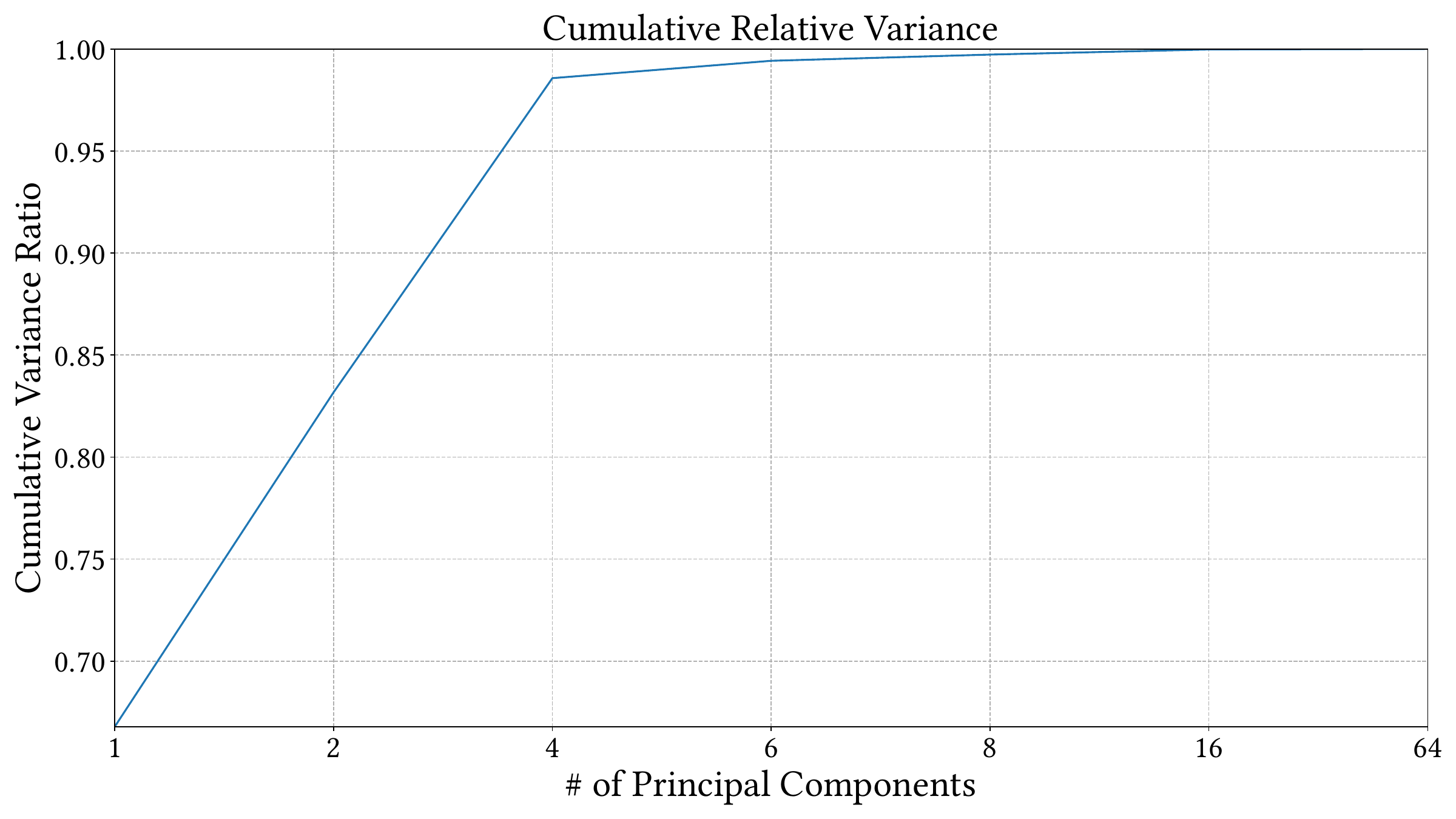}
    \caption{Cumulative relative variance. 
    }
    \label{fig:strand-pca}
\end{minipage}
\hfill
\begin{minipage}[t]{.32\textwidth}
    \centering
    \includegraphics[width=\columnwidth]{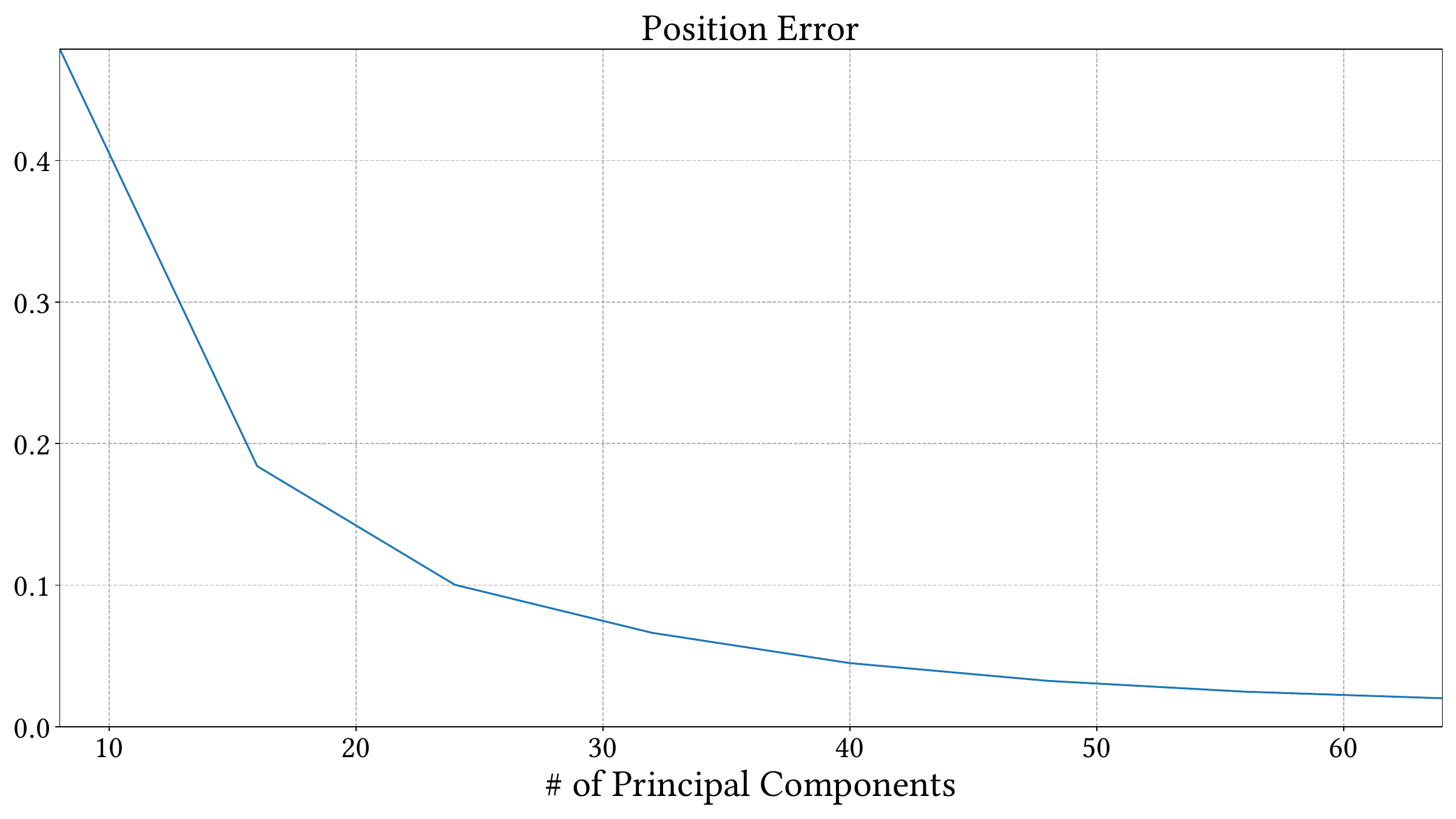}
    \caption{Position error.}
    \label{fig:strand-pca-coeff-pos}
\end{minipage}
\hfill
\begin{minipage}[t]{.32\textwidth}
    \centering
    \includegraphics[width=\columnwidth]{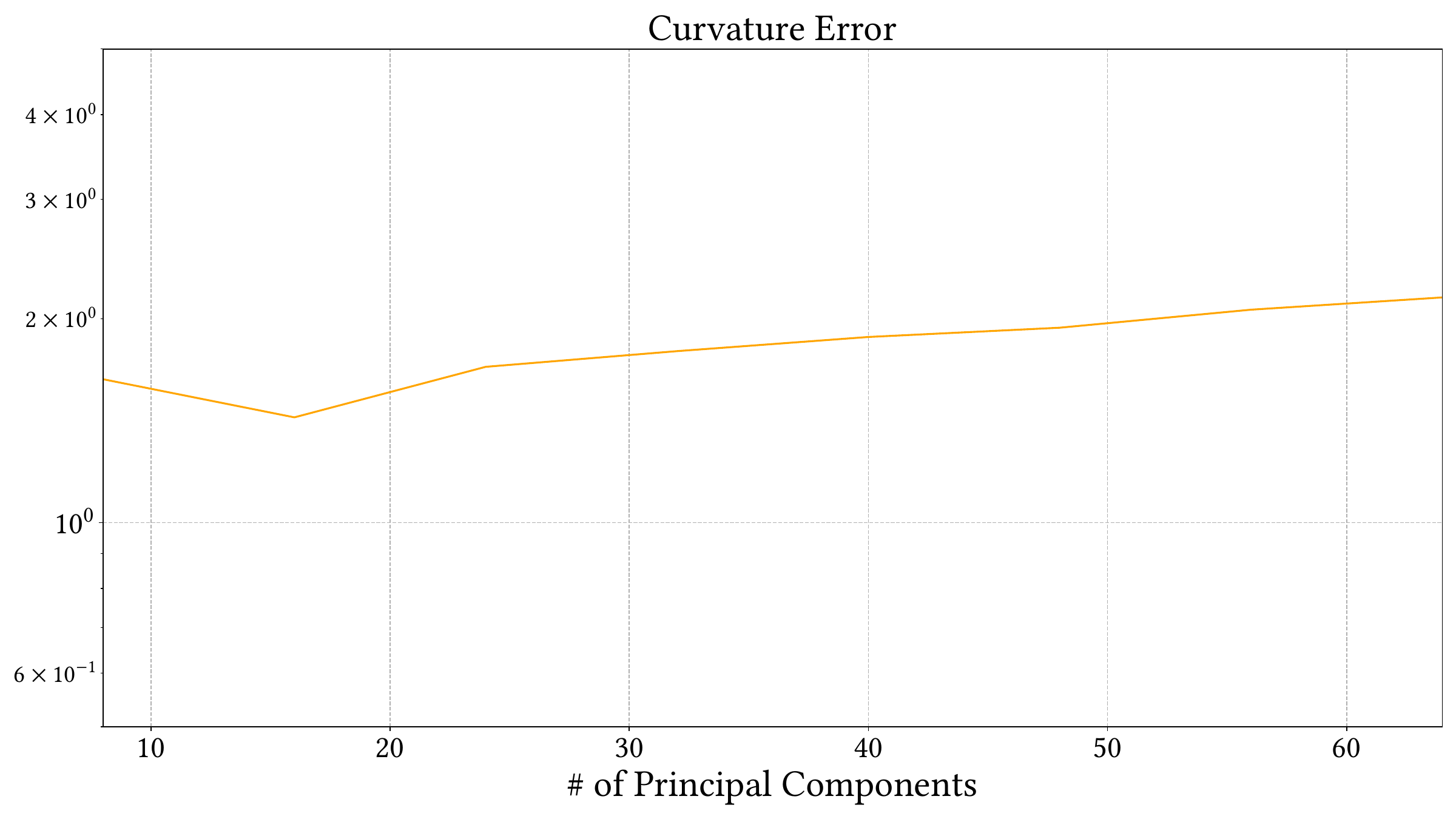}
    \caption{Curvature error.}
    \label{fig:strand-pca-coeff-cur}
\end{minipage}
\vspace{-2mm}
\end{figure}

In Fig.~\ref{fig:strand-pca}, we illustrate the explained cumulative relative variance against the number of principal components. Although $20$ PCA coefficients appear to capture nearly $100\%$ of the variance in the \emph{training set}, increasing the number of coefficients improves the generalizability of our representation to unseen data, as evidenced by the reconstruction errors shown in Fig.~\ref{fig:strand-pca-coeff-pos} and Fig.~\ref{fig:strand-pca-coeff-cur} on the \emph{testing set} with $17$ public hairstyles. 
We suspect that though high-frequency features are sparse and contribute little to the variance, they are important to generalize fitted PCs to unseen curlier data.
Considering this issue, we opt for $64$ principal components, a choice consistent with most previous work on strand representation~\citep{rosu2022neuralstrands, sklyarova2023neural_haircut, zhou2023groomgen}, while achieving significantly lower reconstruction errors.

\begin{figure}[ht]
    \centering
    \addtolength{\tabcolsep}{-5pt}
    \begin{tabular}{lccc}
        \includegraphics[height=0.318\textwidth]{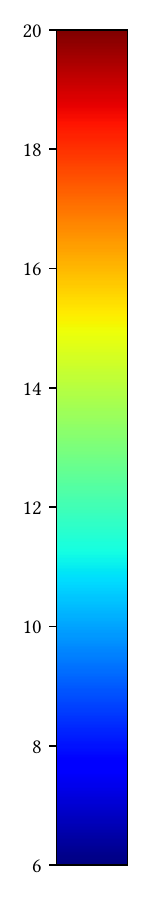}  &
        \includegraphics[width=0.308\textwidth]{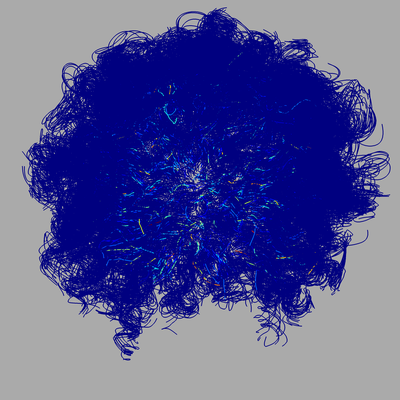} &
        \includegraphics[width=0.308\textwidth]{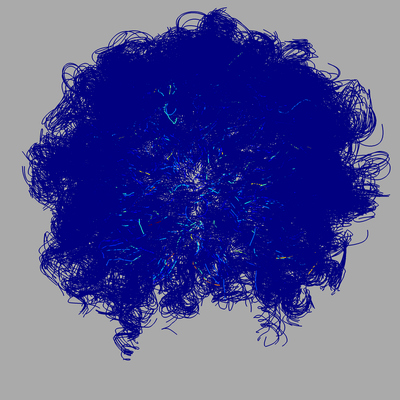} &
        \includegraphics[width=0.308\textwidth]{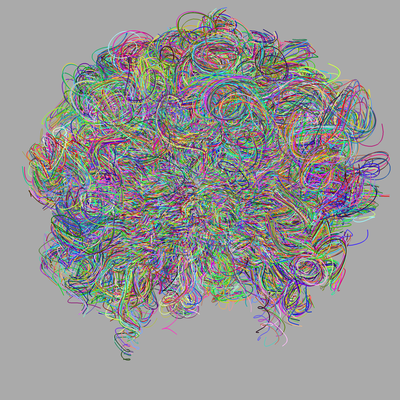} \\
         & \small PCA & \small Freq. PCA & \small Ground Truth \\
    \end{tabular}
    \caption{
    \new{Comparison of PCA and Freq. PCA-based strand representation. Reconstructed strands are color-coded by their curvature error.}
    }
    \label{fig:strand-repr-comp-cur}
    \vspace{-2mm}
\end{figure}

\new{To better elaborate the difference between PCA and Freq. PCA, we visualize the reconstructed hairstyles of these two representations in Fig.~\ref{fig:strand-repr-comp-cur}, where strands are color-coded by their curvature error. The average curvature errors for the entire hairstyle of these two representations are $2.671$ and $2.341$, respectively. These results demonstrate the clear advantage of our Freq. PCA representation in preserving strand curvature, both qualitatively and quantitatively.}

To further demonstrate the robustness of our strand representation, we trained all models on a dataset of $80$ manually groomed hairstyles, with curliness types varying from I to VI as defined in~\citep{loussouarn2007worldwide}, comprising a total of $4,368,679$ strands. The newly trained models are then tested on the same $17$ publicly available hairstyles, with reconstruction errors reported in Table~\ref{tab:strand-repr-advanced}. Even on this distinct dataset, our strand representation consistently achieves a significantly lower position error compared to other deep learning-based representations.
\begin{table}[ht]
    \centering
    \caption{Reconstruction errors reported on strand representations trained on a different strand dataset. Here \textbf{boldface} corresponds to the best result and \underline{underline} means the second best.}
    \label{tab:strand-repr-advanced}
    \addtolength{\tabcolsep}{-0.2em}
    \begin{tabularx}{\columnwidth}{*{5}{Xcccccc}}
    \toprule
                  & Freq. VAE & CNN VAE      & Transformer VAE & MLP VAE  & PCA          & Freq. PCA (Ours)   \\ \midrule
    \# params.    & $15.67$M  & $759.24$K    & $2.29$M         & $15.67$M & $19.31$K     & $19.50$K     \\ \midrule          
    pos. err.     & $1.450$   & $0.142$      & $0.294$         & $\underline{0.117}$  & $\bm{0.026}$     & $\bm{0.026}$ \\ \midrule
    cur. err.     & $\bm{1.409}$   & $4.125$      & $10.331$         & $\underline{2.120}$  & $2.490$      & $2.364$      \\ \bottomrule
  \end{tabularx}
  \vspace{-2mm}
\end{table}

\begin{figure}[ht]
\centering
\begin{minipage}[t]{.32\textwidth}
    \centering
    \includegraphics[width=\columnwidth]{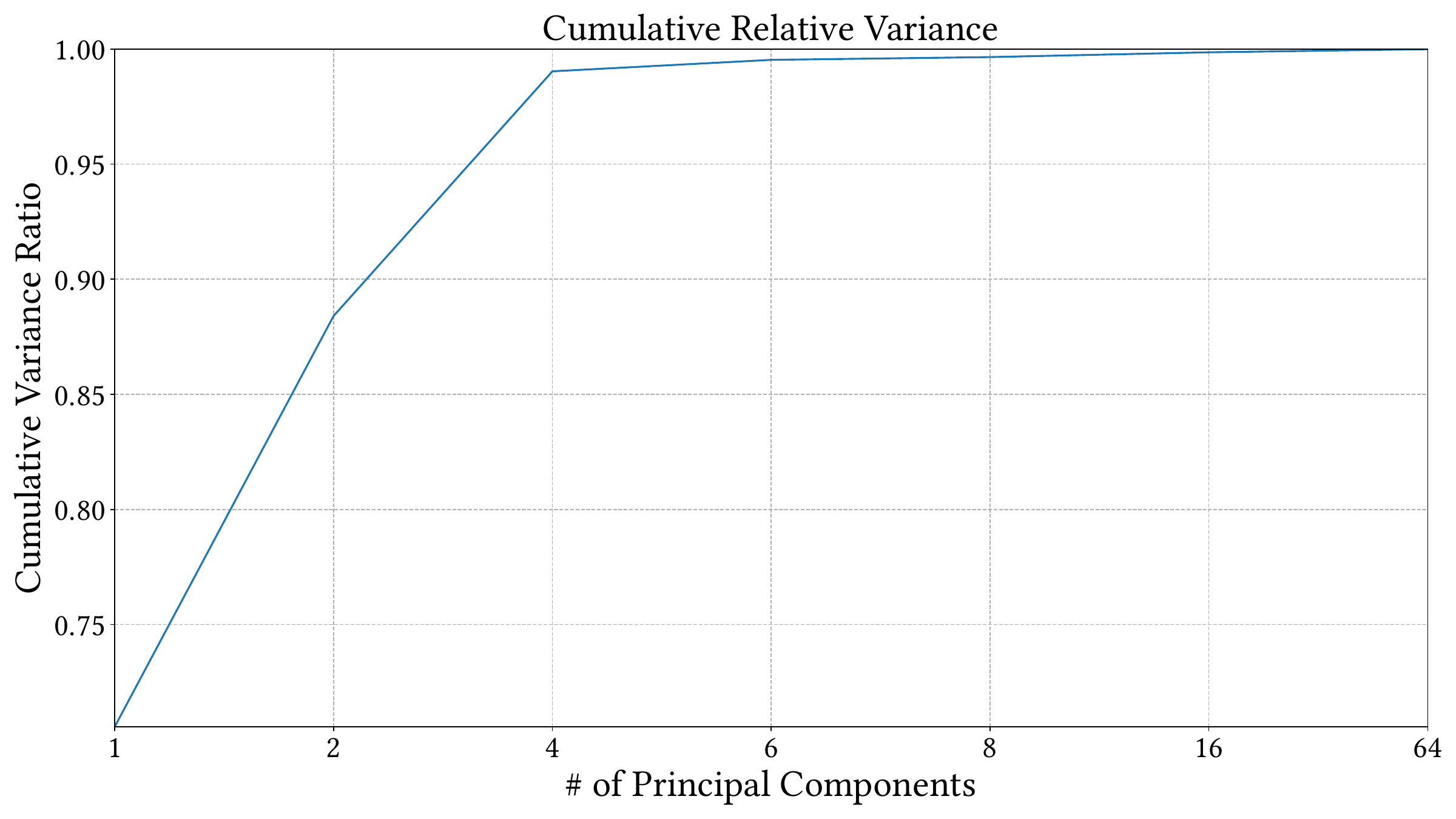}
    \caption{Cumulative relative variance. 
    }
    \label{fig:strand-pca-advanced}
\end{minipage}
\hfill
\begin{minipage}[t]{.32\textwidth}
    \centering
    \includegraphics[width=\columnwidth]{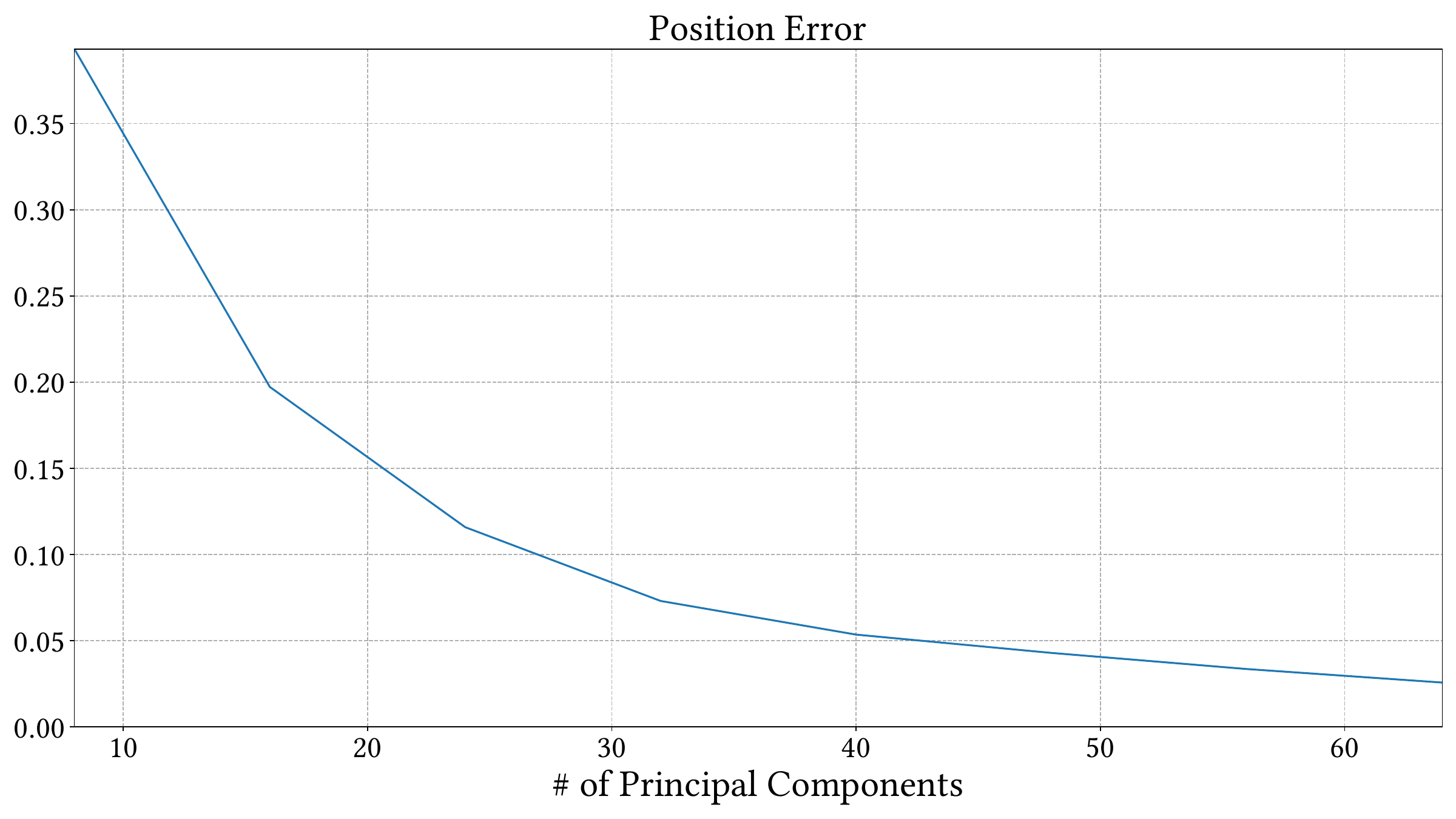}
    \caption{Position error.}
    \label{fig:strand-pca-coeff-advanced-pos}
\end{minipage}
\hfill
\begin{minipage}[t]{.32\textwidth}
    \centering
    \includegraphics[width=\columnwidth]{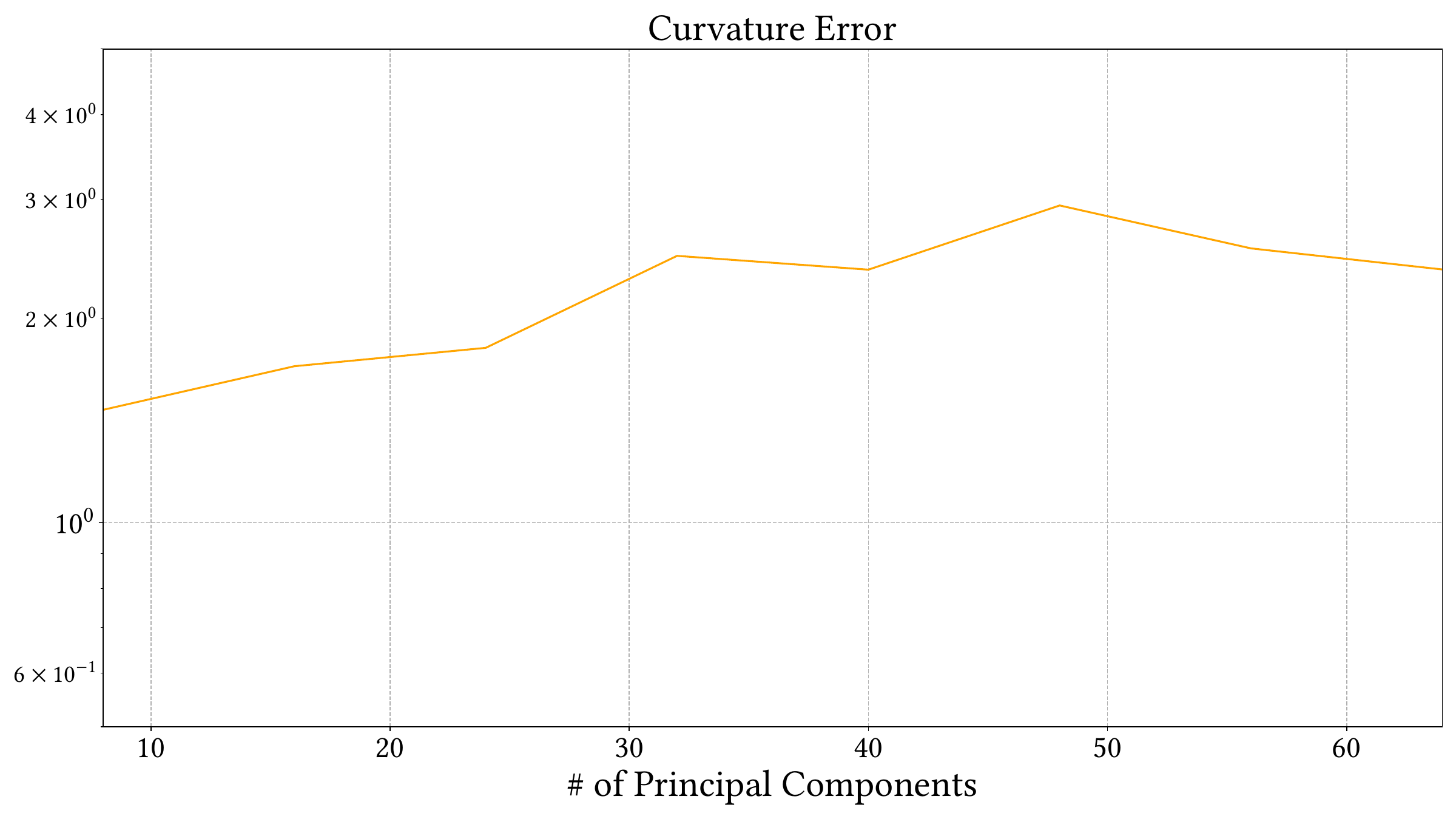}
    \caption{Curvature error.}
    \label{fig:strand-pca-coeff-advanced-cur}
\end{minipage}
\vspace{-2mm}
\end{figure}

Similarly, we plot the cumulative relative variance, reconstructed position error and curvature error as a function of the number of principal components in Fig.~\ref{fig:strand-pca-advanced}, Fig.~\ref{fig:strand-pca-coeff-advanced-pos}, and Fig.~\ref{fig:strand-pca-coeff-advanced-cur}. The overall trends in these figures closely align with the results we observed in our previous experiments.

\begin{figure*}[ht]
    \centering
    \addtolength{\tabcolsep}{-11pt}
    \begin{tabular}{cccccc}
        \includegraphics[width=0.19\textwidth]{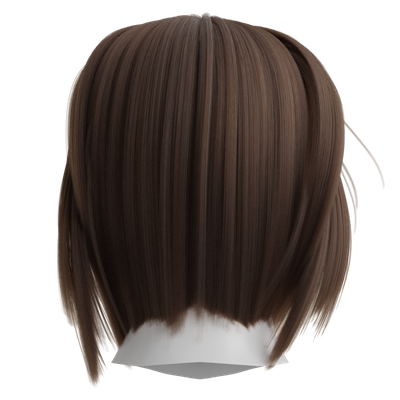} &
        \includegraphics[width=0.19\textwidth]{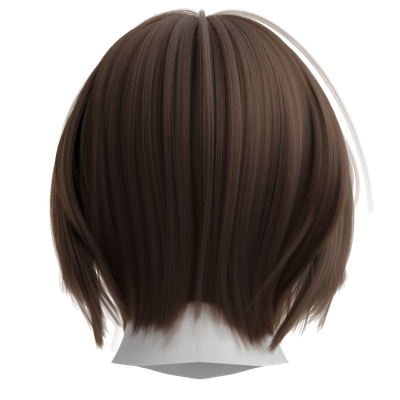} &
        \includegraphics[width=0.19\textwidth]{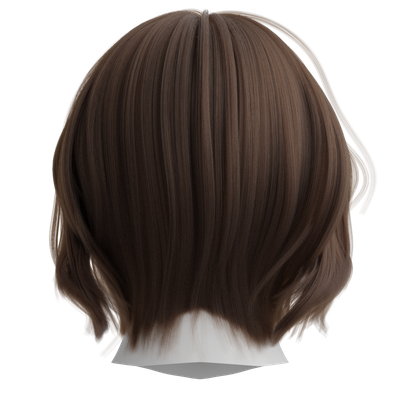} &
        \includegraphics[width=0.19\textwidth]{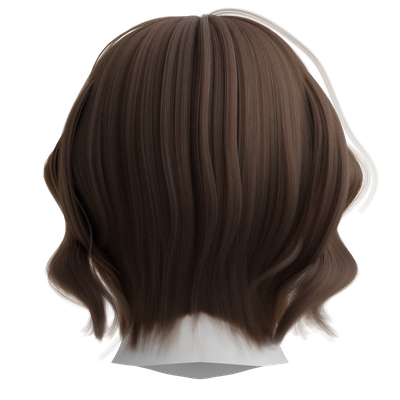} &
        \includegraphics[width=0.19\textwidth]{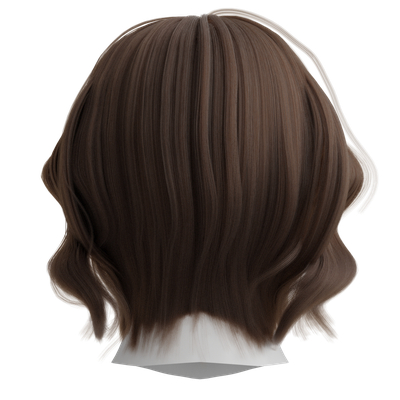} &
        \includegraphics[width=0.19\textwidth]{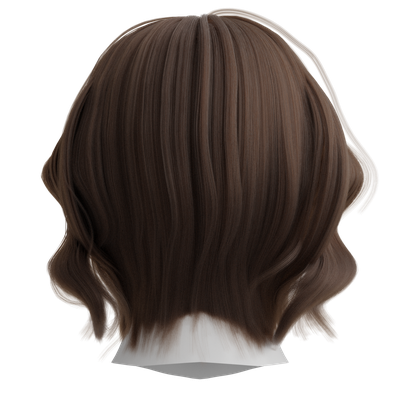} \\
        \small $5$ coeff. & \small $10$ coeff. & \small $15$ coeff. & \small $30$ coeff. & \small $64$ coeff. & \small Ground Truth
    \end{tabular}
    \caption{\new{Comparison of the same hairstyle reconstructed with different numbers of strand PCA coefficients.}}
    \label{fig:pca-analysis}
    \vspace{-2mm}
\end{figure*}

\new{
In Fig.~\ref{fig:pca-analysis} we compare the reconstruction of the same hairstyle using different numbers of strand PCA coefficients. $64$ coefficients per strand are sufficient to closely replicate the original hair, and its curl patterns are captured down to $15$ coefficients per strand. While smooth, $5$ coefficients per strand fail to represent the global hair structure, whereas $10$ coefficients per strand provide the most balanced output in terms of smoothness and global structure preservation.
Based on this observation, we choose to use $10$ coefficients to capture the global structure of each strand, leaving high-frequency details to the remaining 54 coefficients.
}

\subsection{3D Hair Parameterization}
\label{sup:perm-fitting}
To fit \textsc{Perm} parameters to target 3D hair models, we formulate it as an optimization problem, where the objective is defined as:
\begin{equation}
    \vec{\theta}^\ast, \vec{\beta}^\ast \coloneqq \argmin_{\vec{\theta}, \vec{\beta}} \|\mathcal{F}(\mathcal{G}(\vec{\theta}))\oplus\mathcal{D}(\vec{\beta}) - \mathbf{T}\|_1 + \mathcal{L}_{\text{geo}}.
\end{equation}
We employ the Adam optimizer~\citep{kingma2014adam} with an initial learning rate of $0.1$ and a cosine annealing schedule for the learning rate. For better convergence, we first optimize $\vec{\theta}$ only for $1,000$ iterations as a warm-up to match the global shape, and then jointly optimize $\vec{\theta}$ and $\vec{\beta}$ for $4,000$ iterations.

With \textsc{Perm}, we fit parameters to hundreds of publicly available 3D hair models sourced from the Internet. Similar to AMASS~\citep{AMASS:ICCV:2019}, we curated a dataset of 3D hair in a unified and parametric manner, which we released as well to facilitate future research.

\subsection{Baselines}

\paragraph{GroomGen~\citep{zhou2023groomgen}} As the authors of GroomGen have not publicly release their code, we implement it ourselves with Python 3.9.18 and Pytorch 1.11.0. The model is then trained on the same USC-HairSalon dataset as described in Appendix.~\ref{supp:dataset}. Our implementation is further verified with part of the official checkpoints we obtained from the authors.

\paragraph{HairStep~\citep{zheng2023hair}} We use the pre-trained HairStep model released by the authors\footnote{\url{https://github.com/GAP-LAB-CUHK-SZ/HairStep}}.

\paragraph{Strand VAEs} The architecture of different strand VAEs compared in Table~\ref{tab:strand-repr} is adapted from GroomGen~\citep{zhou2023groomgen}, where we only modify the decoder architectures to adopt CNN, Transformer~\citep{vaswani2017attention}, and ModSIREN~\citep{mehta2021modulated}.
\section{Ablation Study}

\subsection{Guide Texture Synthesis} 

\begin{wraptable}[11]{r}{0.38\textwidth}
    \centering
    \vspace{-12pt}
    \setlength{\tabcolsep}{4pt}
    \caption{Reconstruction errors reported on different architectures for guide textures.}
    \label{tab:model-repr}
    \begin{tabular}{@{}lcc@{}}
    \toprule
    \multirow{2}{*}{}             & \multicolumn{2}{c}{\textbf{Guide Textures}}\\ \cmidrule(l){2-3}
                                  & PCA     & StyleGAN2 (Ours) \\ \midrule
    pos. err.                     & $1.163$ & $\bm{0.611}$     \\ \midrule
    cur. err.                     & $2.939$ & $\bm{1.582}$     \\ \bottomrule
    \end{tabular}
\end{wraptable}

\begin{figure}[ht]
    \centering
    \addtolength{\tabcolsep}{-11pt}
    \begin{tabular}{cccccc}
        \includegraphics[width=0.19\textwidth, trim=10 0 10 0, clip]{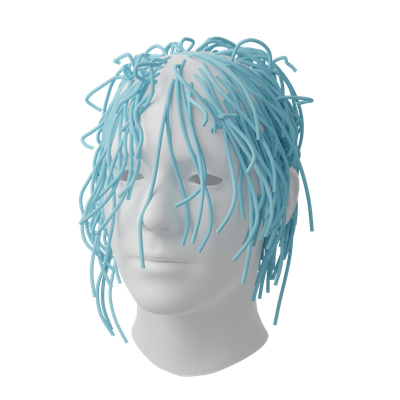} &
        \includegraphics[width=0.19\textwidth, trim=10 0 10 0, clip]{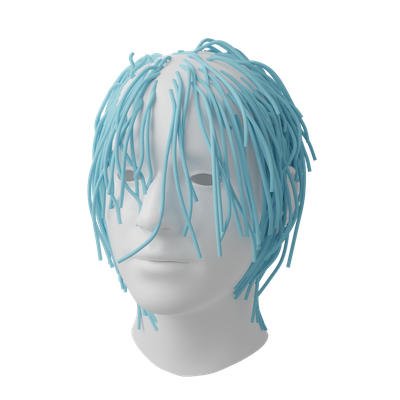} &
        \includegraphics[width=0.19\textwidth, trim=10 0 10 0, clip]{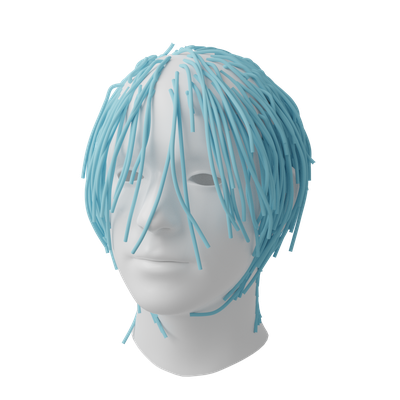} &
        \includegraphics[width=0.19\textwidth]{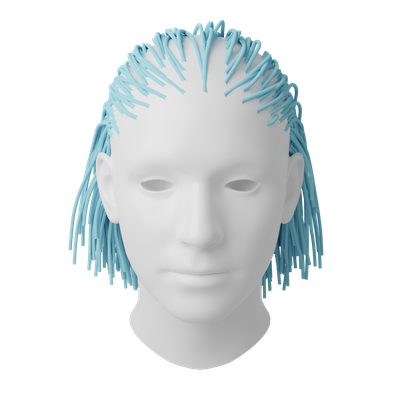} &
        \includegraphics[width=0.19\textwidth]{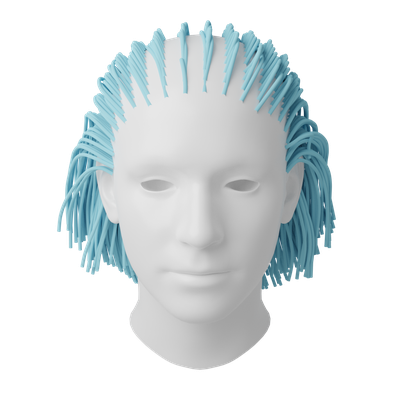} &
        \includegraphics[width=0.19\textwidth]{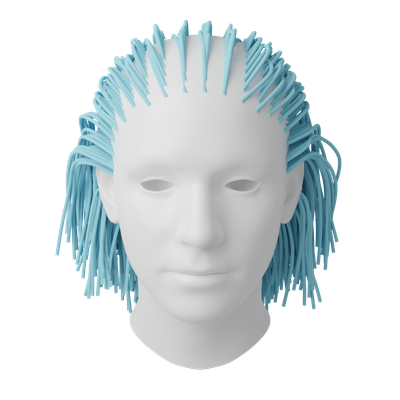} \\
        \small PCA & \small StyleGAN2 (Ours) & \small Ground Truth & \small PCA & \small StyleGAN2 (Ours) & \small Ground Truth \\
    \end{tabular}
    \caption{Comparison of PCA and StyleGAN2 in reconstructing given guide strands.}
    \label{fig:guide-strand-recon}
    \vspace{-2mm}
\end{figure}

To assess the performance of StyleGAN2 in guide texture synthesis, we first create a PCA-based representation for guide textures with a similar formulation as our strand representation, and set the subspace dimension to $512$ to align with \textsc{Perm}'s setup. 
We then embed guide textures from the testing set into these latent spaces, followed by a decoding step to reconstruct them. 
Strands are further decoded from the reconstructed textures to evaluate quantitative errors, which are reported in Table~\ref{tab:model-repr}. It is evident that results from StyleGAN2 exhibit both smaller position and curvature errors, indicating that StyleGAN2 learns a more expressive latent space than PCA.
Fig.~\ref{fig:guide-strand-recon} displays examples of reconstructed guide strands, illustrating that StyleGAN2 more faithfully preserves the structure of the ground truth.

Reasons for the failure of PCA are straightforward: USC-HairSalon contains a vast number of strands, specifically $3,150,559$, for us to learn the PCA subspace, with a compression rate of approximately $78.7\%$. However, for guide textures, we can only create $21,054$ samples after data augmentation, which is merely around $0.6\%$ of the total number of available strands. Moreover, the compression rate increases to $95\%$. 
These two factors lead to the poor performance of PCA in this case.

\subsection{Guide Texture Upsampling} 
\begin{figure*}[ht]
    \centering
    \addtolength{\tabcolsep}{-11pt}
    \begin{tabular}{cccccc}
        \includegraphics[width=0.19\textwidth]{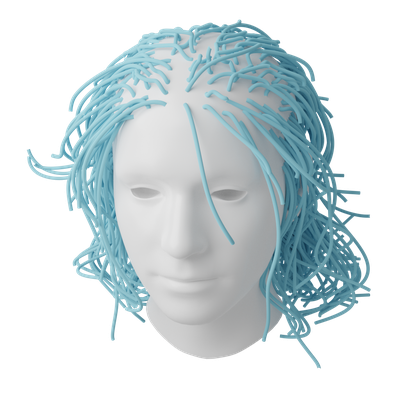} &
        \includegraphics[width=0.19\textwidth]{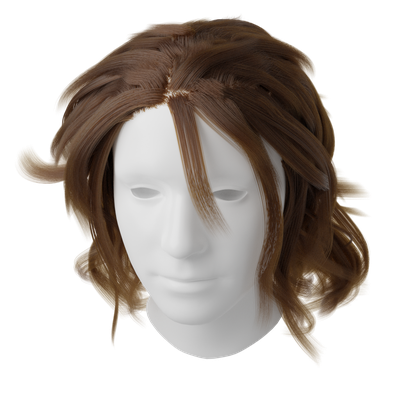} &
        \includegraphics[width=0.19\textwidth]{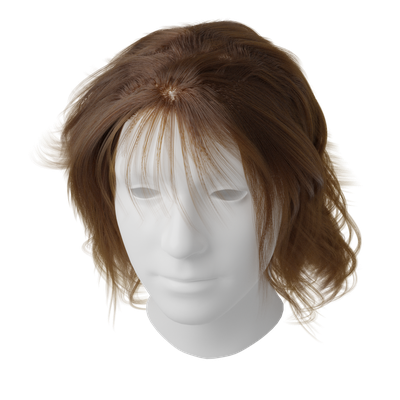} &
        \includegraphics[width=0.19\textwidth]{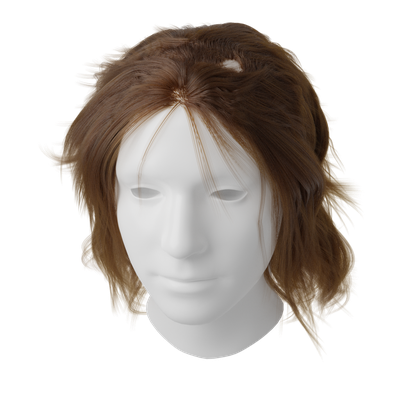} &
        \includegraphics[width=0.19\textwidth]{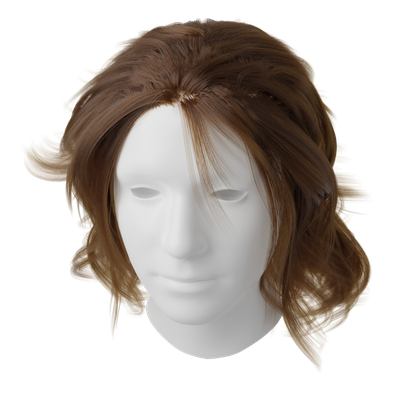} &
        \includegraphics[width=0.19\textwidth]{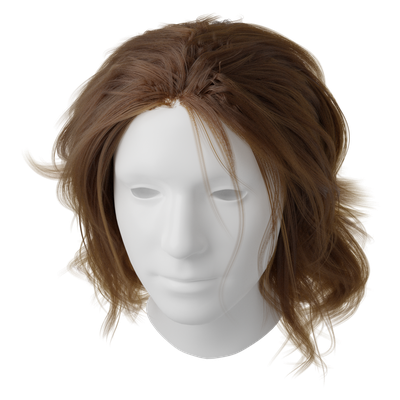} \\
        \small Guide Strands & \small Nearest & \small Bilinear & \small U-Net (coeff.) & \small Ours & \small Ground Truth
    \end{tabular}
    \caption{Comparison of strands upsampled with different interpolation methods.}
    \label{fig:guide-interp}
    \vspace{-2mm}
\end{figure*}

In Fig.~\ref{fig:guide-interp} we compare the strands decoded from textures upsampled with different interpolation methods. Nearest neighbor interpolation ($2$nd column) produces aliased strands as it involves simple repetition. Bilinear interpolation ($3$rd column) leads to smoother strands, but introduces unwanted flyaway fibers, particularly in the forehead area. Our method ($5$th column) achieves the most natural result, which resembles the shape of guide strands and forms reasonable hair partitions. In addition, we employ the same U-Net architecture but configure it to directly predict strand PCA coefficients rather than the blending weights. The output ($4$th column) contains more flyaway strands and a reduced fidelity in capturing the overall hair shape, particularly noticeable in the lower part of the hair.

\subsection{Residual Texture Synthesis} 
\begin{figure}[ht]
    \centering
    \addtolength{\tabcolsep}{-5pt}
    \begin{tabular}{cccc}
        \includegraphics[width=0.24\textwidth]{fig/img/pca_analysis/coeff-10.png} &
        \includegraphics[width=0.24\textwidth]{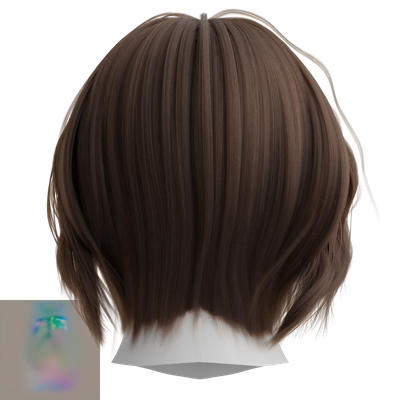} &
        \includegraphics[width=0.24\textwidth]{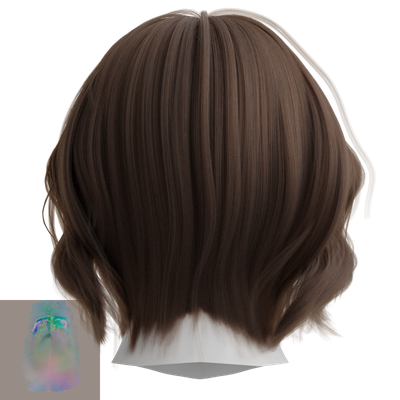} & 
        \includegraphics[width=0.24\textwidth]{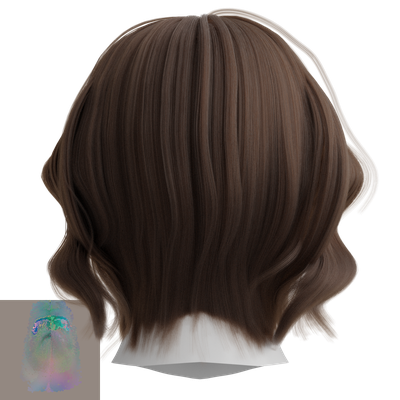} \\
        \small \new{No Res. Tex.} & \small StyleGAN2 & \small VAE (Ours) & \small Ground Truth \\
    \end{tabular}
    \caption{\new{Residual texture reconstruction with different network architectures. Insets show the visualization of the corresponding residual texture. Please zoom in for details.} 
    }
    \label{fig:restex}
    \vspace{-2mm}
\end{figure}

\new{In Fig.~\ref{fig:restex} we evaluate various network architectures for residual texture synthesis. When residual textures are omitted, the resulting hairstyle captures only the global structure of the ground truth, lacking all high-frequency details such as different curl patterns ($1$st column).} Though StyleGAN2 performed well in synthesizing guide strands, we found it collapsed to produce blurry residual textures when embedding them into the latent space, as shown in Fig.~\ref{fig:restex} \new{($2$nd column)}. We conjecture that this issue may be attributed to the large size of residual textures ($256 \times 256 \times 54$), as they contain more data than high-resolution RGB images ($1024 \times 1024 \times 3$). Moreover, our dataset is smaller than 
those image datasets used for StyleGAN training (e.g., FFHQ~\citep{karras2019style}, which contains $70$K high-quality portrait images), thus further suppressing the expressiveness of StyleGAN2. Considering these constraints, we devised our network architecture as a VAE, trading some sampling diversity for higher reconstruction fidelity, which reconstructs sharper residual textures \new{($3$rd column)}. 

\subsection{Analysis of GroomGen}
\label{supp:groomgen}

Since GroomGen~\citep{zhou2023groomgen} is not fully open-sourced, we first verify the correctness of our implementation by comparing it to the official checkpoints of the strand VAE and hairstyle VAE, which we obtained from the authors.

\begin{wrapfigure}[13]{r}{0.62\textwidth}
    \vspace{-10pt}
    \centering
    \addtolength{\tabcolsep}{-5pt}
    \begin{tabular}{lccc}
        \includegraphics[height=0.22\textwidth]{fig/img/strand_repr_comp/strand_err.pdf}  &
        \includegraphics[width=0.18\textwidth, trim=40 0 40 0, clip]{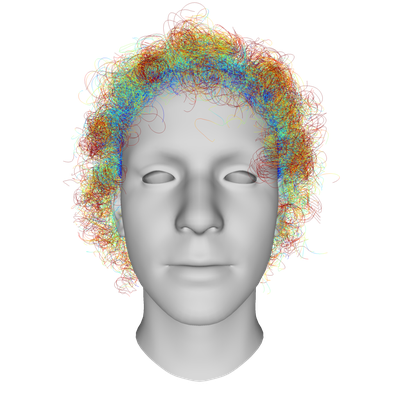} &
        \includegraphics[width=0.18\textwidth, trim=40 0 40 0, clip]{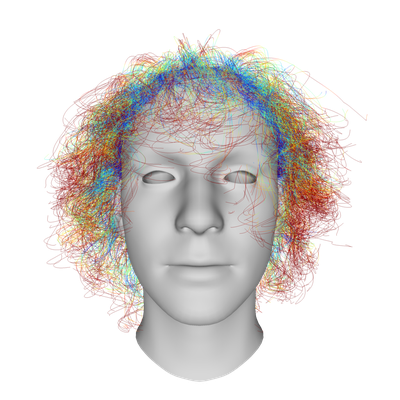} &
        \includegraphics[width=0.18\textwidth, trim=40 0 40 0, clip]{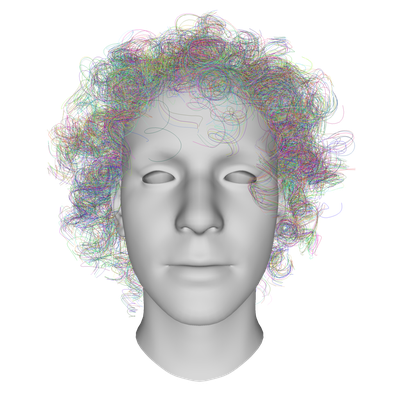} \\
         & \small Official Strand VAE & \small Our Implementation & \small Ground Truth \\
    \end{tabular}
    \caption{
    Comparison of our strand VAE implementation with the official checkpoint of GroomGen~\citep{zhou2023groomgen}.
    }
    \label{fig:groomgen-strand-vae}
    \vspace{-2mm}
\end{wrapfigure}

For the strand VAE, we conducted the same strand reconstruction experiments as described in Sec.~\ref{sec:strand-repr-exp} and visualize the reconstructed results in Fig.~\ref{fig:groomgen-strand-vae}, where both our implementation and the official version struggle to preserve the overall structure of the given hairstyle. From a quantitative perspective, our implementation even outperformed the official model, achieving lower position and curvature errors on the test data. Specifically, the official checkpoint yielded a position error of $1.521$ and a curvature error of $1.337$, whereas our implementation achieved $1.211$ and $0.910$, respectively.

\begin{figure}[ht]
    \centering
    \addtolength{\tabcolsep}{-11pt}
    \begin{tabular}{cccccc}
        \includegraphics[width=0.19\textwidth]{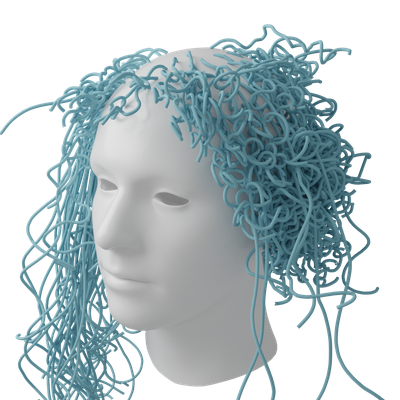} &
        \includegraphics[width=0.19\textwidth]{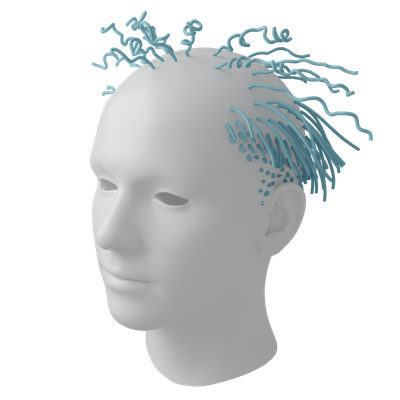} &
        \includegraphics[width=0.19\textwidth]{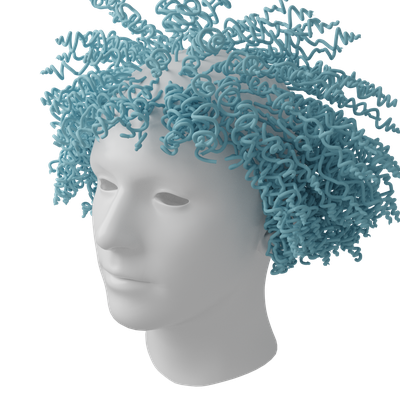} &
        \includegraphics[width=0.19\textwidth]{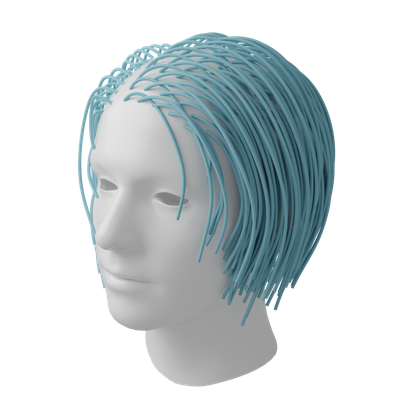} &
        \includegraphics[width=0.19\textwidth]{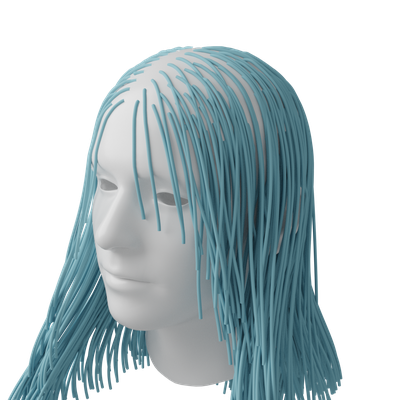} &
        \includegraphics[width=0.19\textwidth]{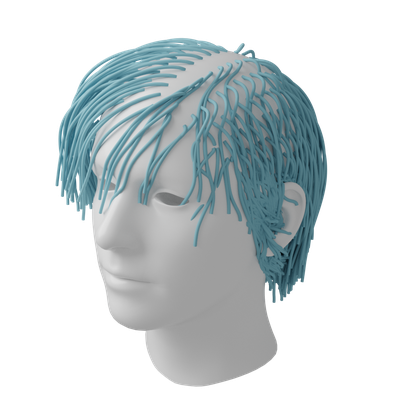} \\
        \multicolumn{3}{c}{\small Official Hairstyle VAE} & \multicolumn{3}{c}{\small Our Implementation} \\
    \end{tabular}
    \caption{Comparison of our hairstyle VAE implementation with the official checkpoint of GroomGen~\citep{zhou2023groomgen}. Our implementation and the official version are sampled using the same Gaussian noise.}
    \label{fig:groomgen-hairstyle-vae}
    \vspace{-2mm}
\end{figure}

For the hairstyle VAE, we were unable to embed guide strands from our head mesh into GroomGen's latent space due to the lack of $uv$ parameterization of the head mesh they used. Instead, we randomly sampled the latent spaces of both our implementation and the official version, using the same Gaussian noise, and visualized the decoded guide strands in Fig.~\ref{fig:groomgen-hairstyle-vae}. As expected, our implementation generates natural-looking guide strands, though they are sometimes overly smooth. In contrast, the official hairstyle VAE produces curlier guide strands, but sometimes, the results appear less natural. We hypothesize that these differences arise from the difference of the training data.

For the neural upsampler, as the authors cannot share their checkpoint, we can only use our own implementation as reference. However, as a GAN, we found this module very unstable to train, easily collapsing to weird outputs (see Fig.~\ref{fig:groomgen-comp} and Fig.~\ref{fig:perm-sampling}).

\begin{figure}[ht]
    \centering
    \addtolength{\tabcolsep}{-6pt}
    \begin{tabular}{ccc}
        \includegraphics[width=0.33\textwidth]{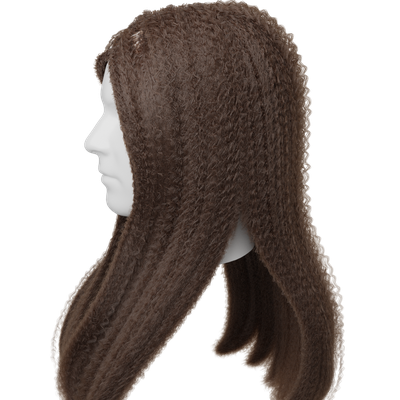} & 
        \includegraphics[width=0.33\textwidth]{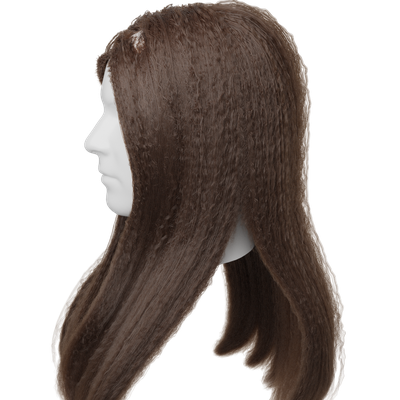} & 
        \includegraphics[width=0.33\textwidth]{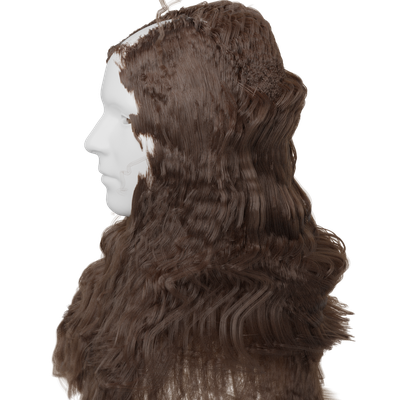} \\
        \small Original Kinky Hair & \small Downsampled Hair & \small Reconstructed Hair \\
        \small $200$ points / strand & \small $100$ points / strand & \small $100$ points / strand \\
    \end{tabular}
    \caption{Verification of our GroomGen implementation on a manually made kinky hairstyle.}
    \label{fig:groomgen-verify}
    \vspace{-2mm}
\end{figure}
To assess whether our implementation of GroomGen can reproduce long kinky hairstyles, we manually created a test hairstyle (Fig.~\ref{fig:groomgen-verify} $1$st column) which originally contains $200$ points per strand to accurately capture the fine curls. We downsampled the strands to $100$ points to fit our design, which led to a noticeable degradation in curl fidelity (Fig.~\ref{fig:groomgen-verify} $2$nd column). We then searched for the optimal latent code to reconstruct the downsampled hair, yielding the results shown in the last column of Fig.~\ref{fig:groomgen-verify}. Despite these efforts, our implementation struggled to faithfully reconstruct the hairstyle with accurate fine details.
\section{Additional Results}

\subsection{Random Hairstyle Synthesis}
\label{sup:sampling}

\begin{figure}[ht]
    \centering
    \addtolength{\tabcolsep}{-10pt}
    \begin{tabular}{ccccc}
        \raisebox{55pt}{\rotatebox[origin=c]{90}{\small PCA}} &
        \includegraphics[width=0.26\textwidth, trim=40 0 40 0, clip]{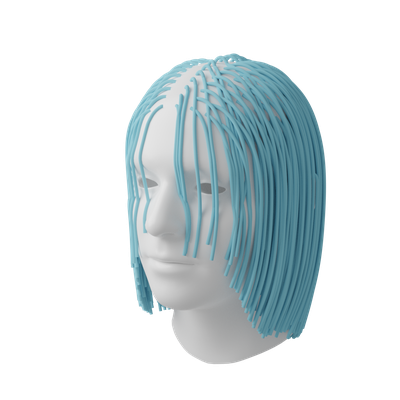} &
        \includegraphics[width=0.26\textwidth, trim=40 0 40 0, clip]{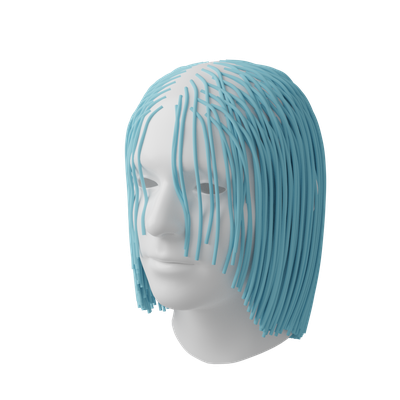} &
        \includegraphics[width=0.26\textwidth, trim=40 0 40 0, clip]{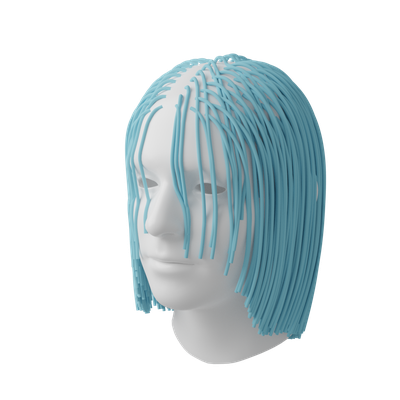} & 
        \includegraphics[width=0.26\textwidth, trim=40 0 40 0, clip]{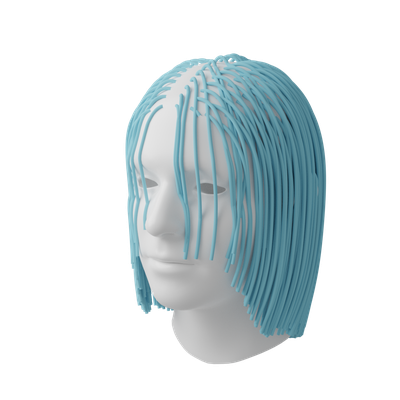} \\
        \raisebox{55pt}{\rotatebox[origin=c]{90}{\small StyleGAN2}} &
        \includegraphics[width=0.26\textwidth, trim=40 0 40 0, clip]{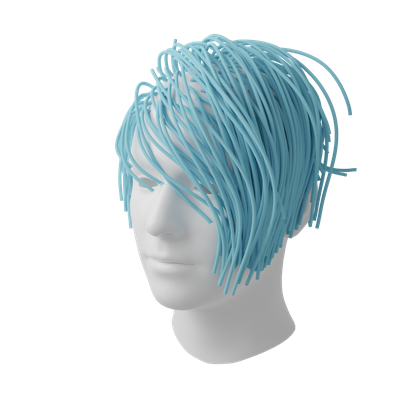} &
        \includegraphics[width=0.26\textwidth, trim=40 0 40 0, clip]{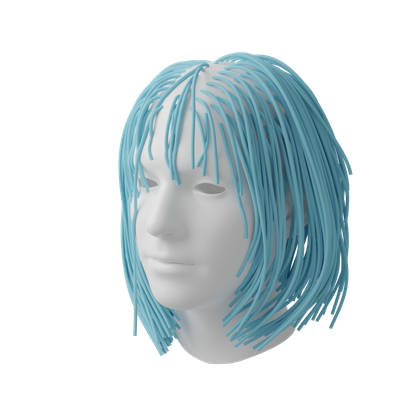} &
        \includegraphics[width=0.26\textwidth, trim=40 0 40 0, clip]{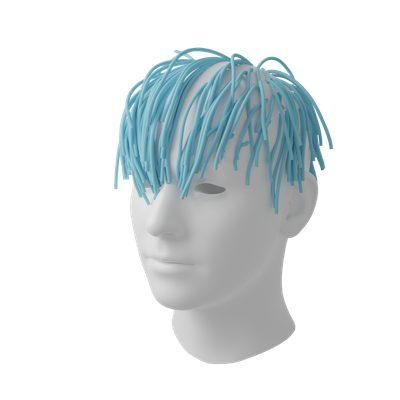} & 
        \includegraphics[width=0.26\textwidth, trim=40 0 40 0, clip]{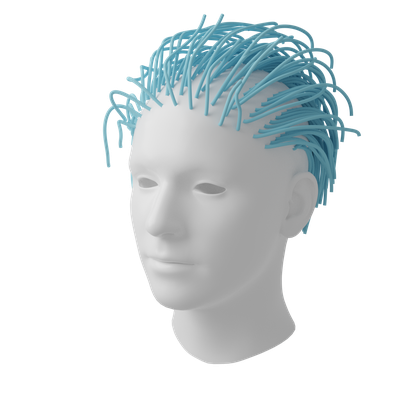} \\
    \end{tabular}
    \caption{Guide strands synthesized from different Gaussian noise (Top: PCA; Bottom: StyleGAN2). Note that PCA samples are collapsed to similar results due to the distribution difference between Gaussian and PCA subspace.}
    \label{fig:guide-strand-sample}
    \vspace{-2mm}
\end{figure}

In Fig.~\ref{fig:guide-strand-sample}, we showcase several random guide strands generated by sampling the parameter space of $\vec{\theta}$ with Gaussian noise, highlighting that the results from StyleGAN2 exhibit greater diversity compared to those generated by the PCA alternative discussed in our main paper. Given that PCA lacks constraints on the distribution of its subspace, obtaining reasonable guide strands by sampling its subspace with Gaussian noise is challenging, and our results indicate that most of them are collapsed into similar outputs.

\begin{figure}[ht]
    \centering
    \addtolength{\tabcolsep}{-1pt}
    \begin{tabular}{cc}
     \raisebox{48pt}{\rotatebox[origin=c]{90}{\tiny GroomGen~\citep{zhou2023groomgen}}} &
     \includegraphics[width=0.936\textwidth]{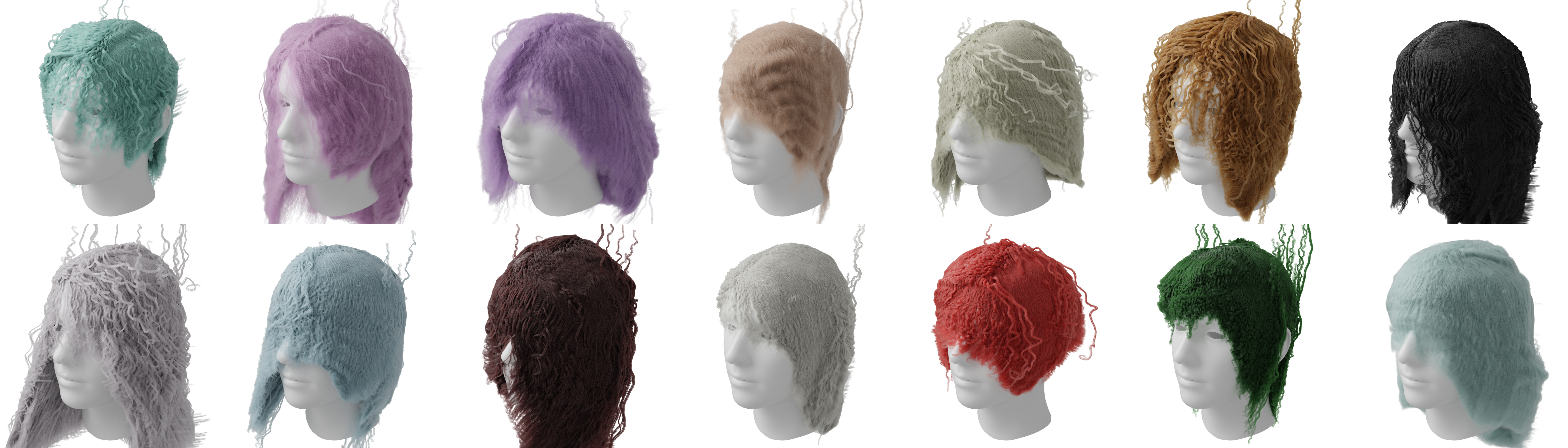} \\
     \raisebox{50pt}{\rotatebox[origin=c]{90}{\tiny Ours}} &
     \includegraphics[width=0.936\textwidth]{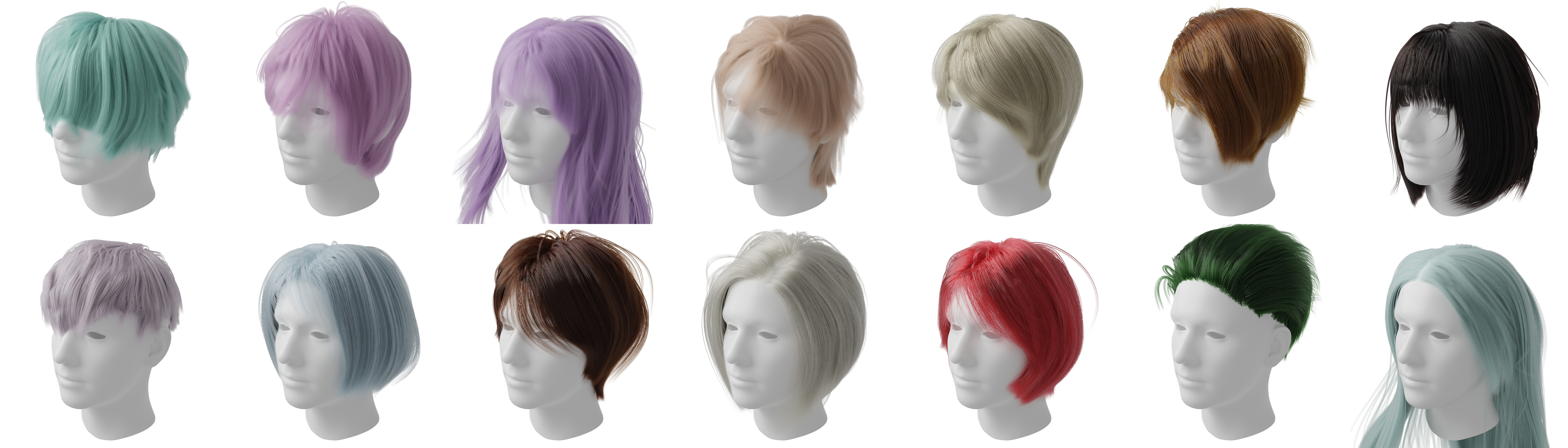} \\
    \end{tabular}
    \caption{\new{Full hair models synthesized from different Gaussian noise, with comparison to our implementation of GroomGen~\citep{zhou2023groomgen} trained on the same dataset. Our model and GroomGen are sampled using the same Gaussian noise. Hair colors are manually assigned for aesthetic purposes.}}
    \label{fig:perm-sampling}
    \vspace{-2mm}
\end{figure}

In Fig.~\ref{fig:perm-sampling}, we illustrate several random full hair models generated by sampling the parameter spaces of $\vec{\theta}$ and $\vec{\beta}$ with Gaussian noise, and compare these results to GroomGen~\citep{zhou2023groomgen}.
Note that we sample our parameter space and GroomGen's latent space with the same Gaussian noise for a fair comparison. From our observation, GroomGen tends to generate hairstyles with implausible curls and flyaway strands, which do not appear in our results. 

\subsection{Hairstyle Interpolation}
\label{sup:hair-interp}

\begin{figure*}[ht]
    \centering
    \addtolength{\tabcolsep}{-2pt}
    \begin{tabular}{cc}
        \raisebox{30pt}{\rotatebox[origin=c]{90}{\tiny Interpolate $\vec{\theta}$ Only}} & \includegraphics[width=0.95\textwidth]{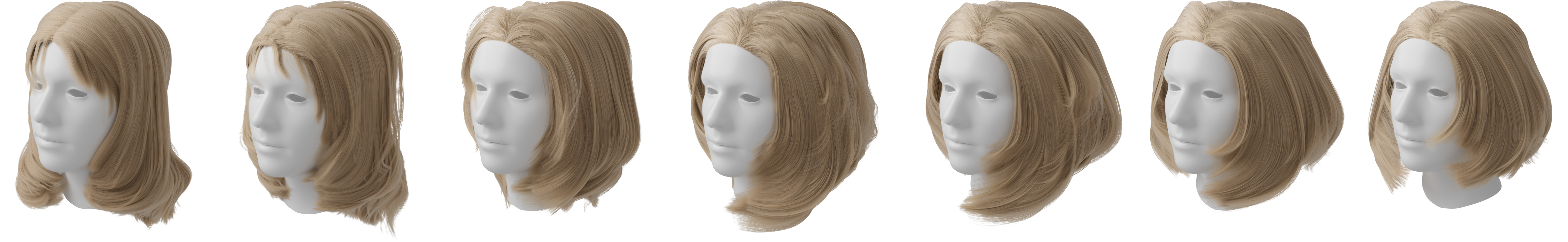}  \\
            \addlinespace[-6pt]
            \raisebox{25pt}{\rotatebox[origin=c]{90}{\tiny Interpolate $\vec{\beta}$ Only}} & \includegraphics[width=0.95\textwidth]{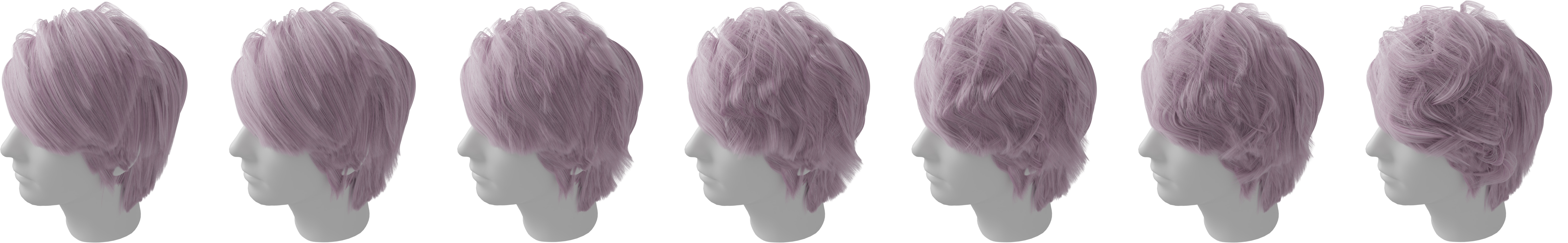}  \\
            \raisebox{40pt}{\rotatebox[origin=c]{90}{\tiny Joint Interpolation}} & \includegraphics[width=0.95\textwidth]{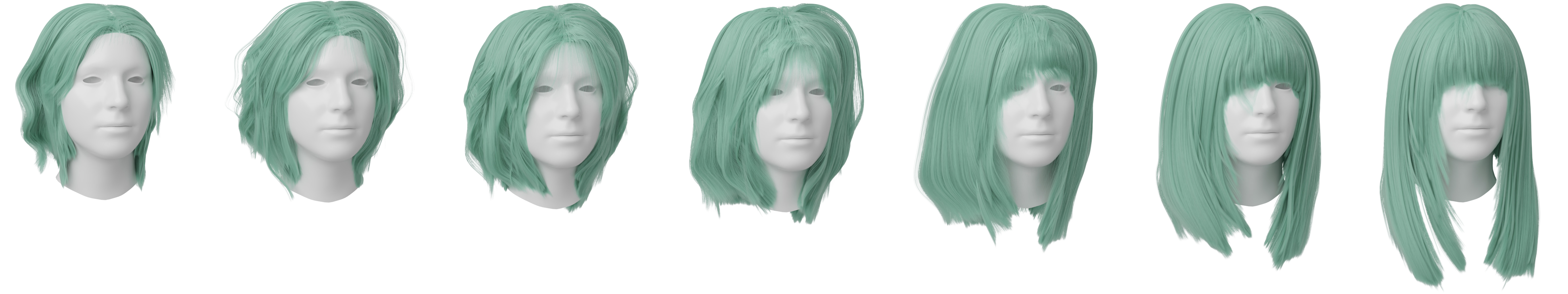}  \\
    \end{tabular}
    \vspace{-10pt}
    \caption{Hairstyle interpolation with different granularity. Top: interpolate $\vec{\theta}$ only and keep $\vec{\beta}$ fixed; Middle: interpolate $\vec{\beta}$ only and keep $\vec{\theta}$ fixed; Bottom: jointly interpolate $\vec{\theta}$ and $\vec{\beta}$.}
    \label{fig:hair-morphing}
     \vspace{-2mm}
\end{figure*}

In Fig.~\ref{fig:hair-morphing} we show hairstyle interpolation with different granularity. In the top row, we linearly interpolate the guide strand parameter $\vec{\theta}$, resulting in hairstyles with different global structures. Since the hair styling parameter $\vec{\beta}$ is kept fixed, local curl patterns, such as curls at the tips of strands, remain consistent.
In the middle row, we fix the parameter $\vec{\theta}$ and linearly interpolate the parameter $\vec{\beta}$, thereby generating novel hairstyles ranging from straight to curly while maintaining a nearly identical haircut.
In the bottom row, we jointly interpolate $\vec{\theta}$ and $\vec{\beta}$, demonstrating the transition from a short wavy hairstyle to a long straight hairstyle.

\begin{figure}[ht]
    \centering
    \addtolength{\tabcolsep}{-4.3pt}
    \begin{tabular}{cccccccc}
     \raisebox{33pt}{\rotatebox[origin=c]{90}{\tiny \citep{weng2013hair}}} &
     \includegraphics[width=0.13\textwidth]{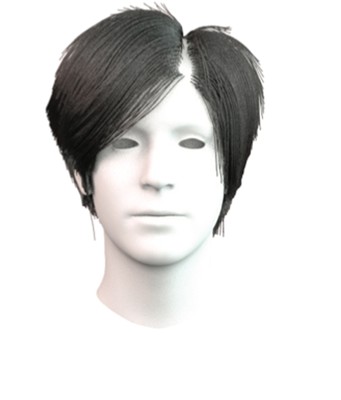} &
     \includegraphics[width=0.13\textwidth]{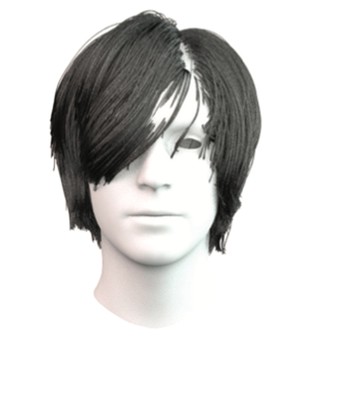} &
     \includegraphics[width=0.13\textwidth]{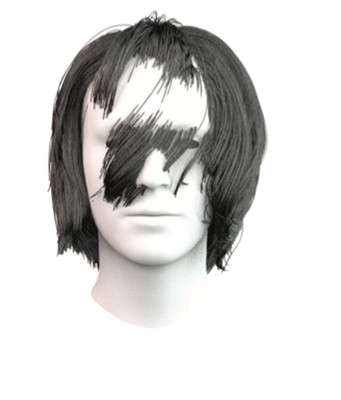} &
     \includegraphics[width=0.13\textwidth]{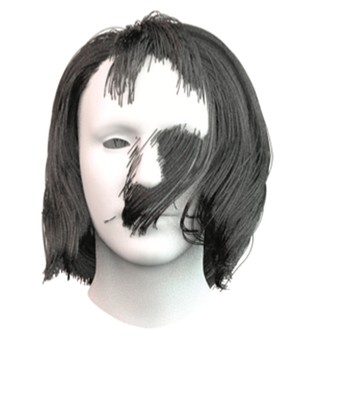} &
     \includegraphics[width=0.13\textwidth]{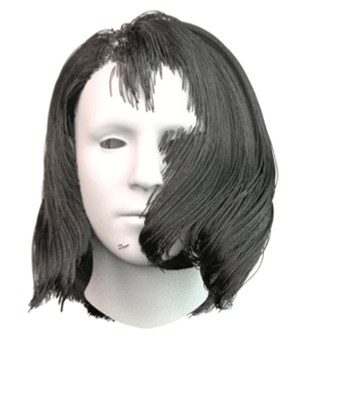} &
     \includegraphics[width=0.13\textwidth]{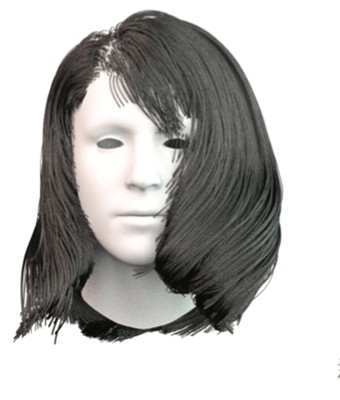} &
     \includegraphics[width=0.13\textwidth]{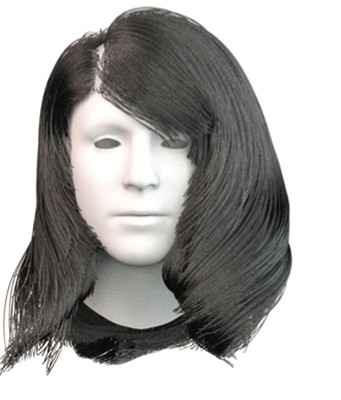} \\
     \addlinespace[-6pt]
     \raisebox{30pt}{\rotatebox[origin=c]{90}{\tiny \citep{zhou2018hairnet}}} &
     \includegraphics[width=0.13\textwidth]{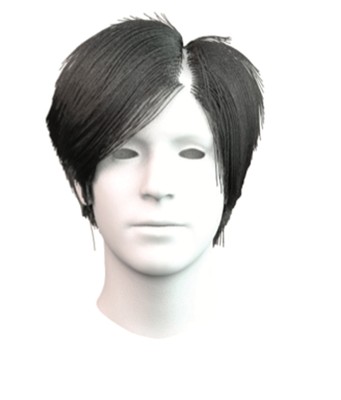} &
     \includegraphics[width=0.13\textwidth]{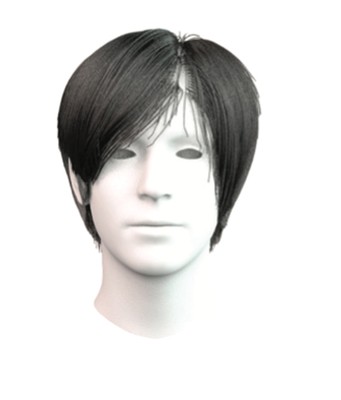} &
     \includegraphics[width=0.13\textwidth]{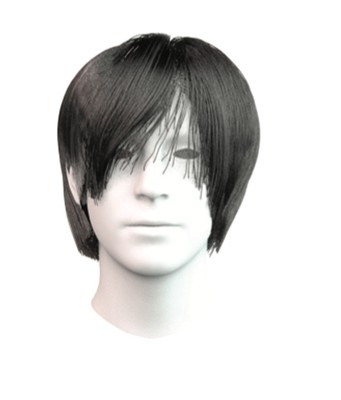} &
     \includegraphics[width=0.13\textwidth]{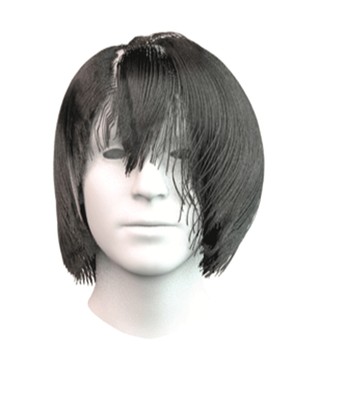} &
     \includegraphics[width=0.13\textwidth]{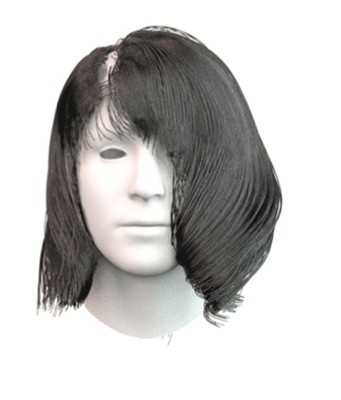} &
     \includegraphics[width=0.13\textwidth]{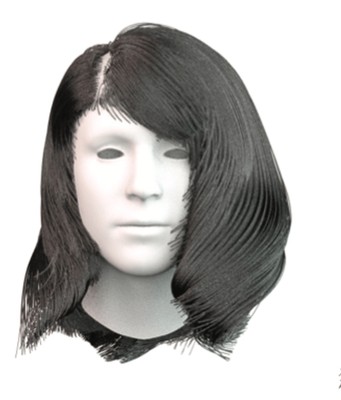} &
     \includegraphics[width=0.13\textwidth]{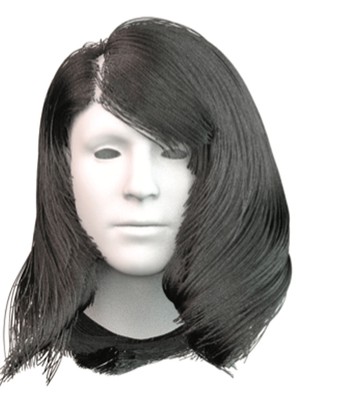} \\
     \addlinespace[-6pt]
     \raisebox{30pt}{\rotatebox[origin=c]{90}{\tiny Ours}} &
     \includegraphics[width=0.13\textwidth]{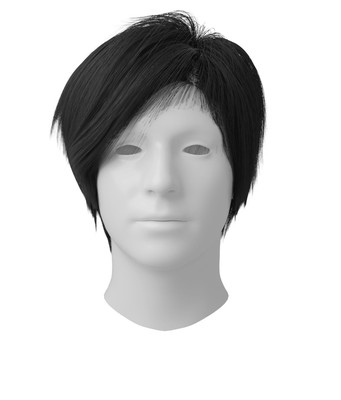} &
     \includegraphics[width=0.13\textwidth]{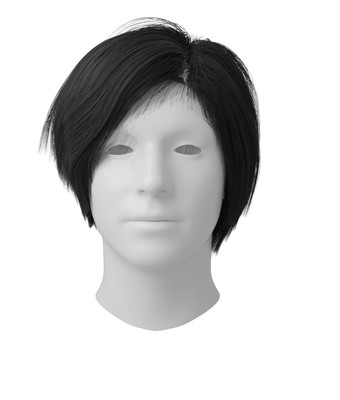} &
     \includegraphics[width=0.13\textwidth]{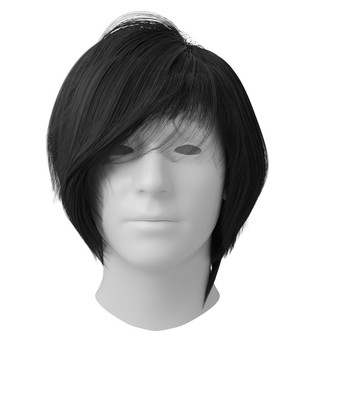} &
     \includegraphics[width=0.13\textwidth]{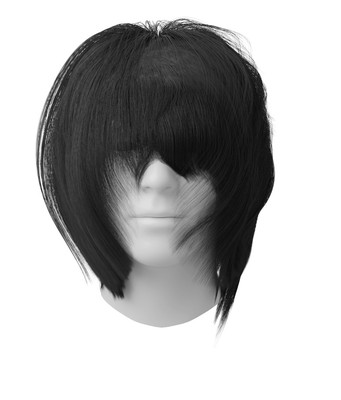} &
     \includegraphics[width=0.13\textwidth]{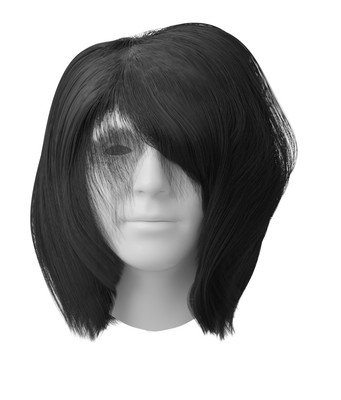} &
     \includegraphics[width=0.13\textwidth]{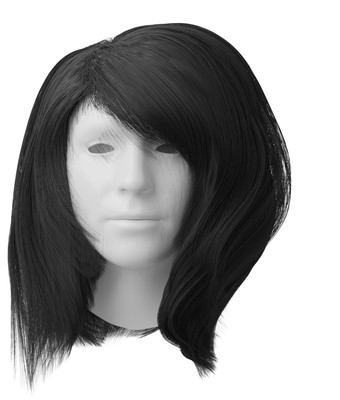} &
     \includegraphics[width=0.13\textwidth]{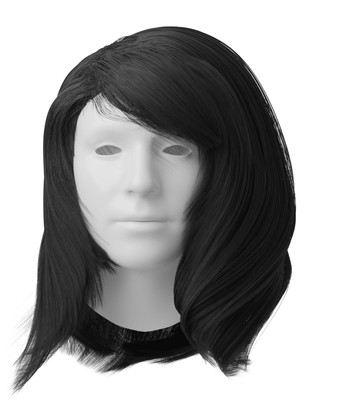} \\
      & \small Hairstyle $A$ & \multicolumn{5}{c}{\small Interpolation Results} & \small Hairstyle $B$
    \end{tabular}
    \caption{Interpolation comparison with \citep{weng2013hair} and \citep{zhou2018hairnet}.}
    \label{fig:hair-interp}
    \vspace{-2mm}
\end{figure}

Additionally, we compare our method with~\citep{weng2013hair} and~\citep{zhou2018hairnet} on hairstyle interpolation. Since neither~\citep{weng2013hair} nor~\citep{zhou2018hairnet} are publicly available, we embed the $2$ hairstyles selected by~\citep{zhou2018hairnet} into our parameter space and jointly interpolate the projected $\vec{\theta}$ and $\vec{\beta}$ to obtain our results. Qualitative comparisons are provided in Fig.~\ref{fig:hair-interp}, demonstrating that our method achieves performance comparable to~\citep{zhou2018hairnet}, with both outperforming the results from~\citep{weng2013hair}.
Note that in the middle column of interpolation, our method generates strands naturally covering the forehead, rather than severely intersecting with the head mesh like~\citep{weng2013hair}. Our starting and ending hairstyles are a bit different from others, which is the result of our parameterization process, as some details cannot be fully reconstructed.

\subsection{Single-view Hair Reconstruction}
\label{sup:single-view}

\begin{figure}[ht]
    \centering
    \addtolength{\tabcolsep}{-5.1pt}
    \begin{tabular}{ccccccc}
     \raisebox{28pt}{\rotatebox[origin=c]{90}{\tiny Input Images}} &
     \includegraphics[width=0.158\textwidth]{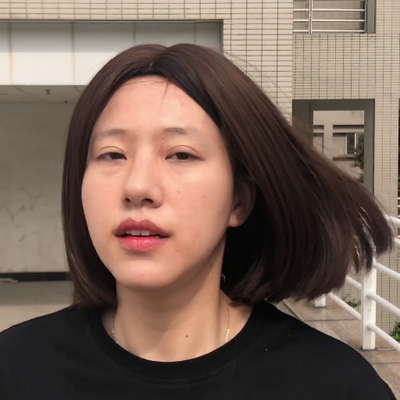} &
     \includegraphics[width=0.158\textwidth]{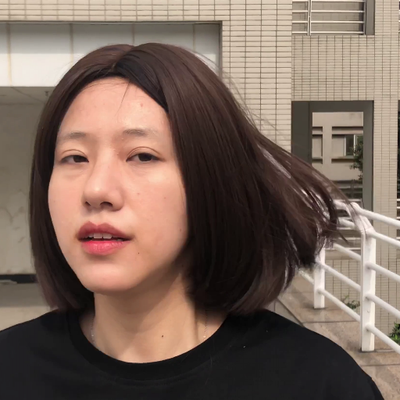} &
     \includegraphics[width=0.158\textwidth]{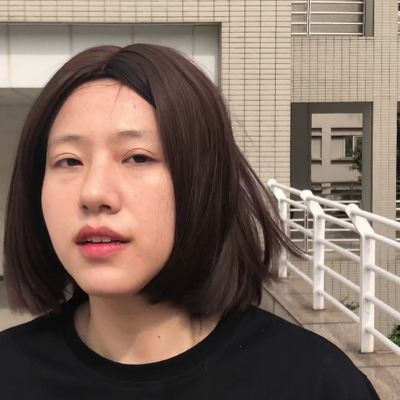} &
     \includegraphics[width=0.158\textwidth]{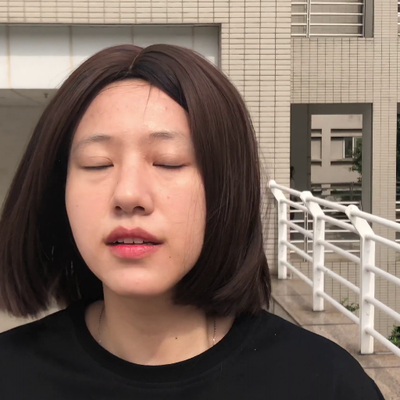} &
     \includegraphics[width=0.158\textwidth]{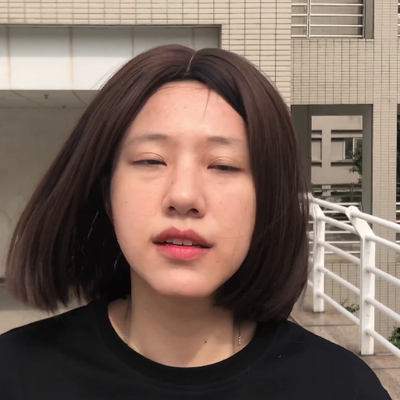} &
     \frame{\adjincludegraphics[height=0.157\textwidth, trim={0 {.2\height} {.7\width} {.5\height}}, clip]{fig/img/single_view/rgb/bbj-000035.png}} \\
     \addlinespace[-1pt]
     \raisebox{28pt}{\rotatebox[origin=c]{90}{\tiny \citep{yang2019dynamic}}} &
     \includegraphics[width=0.158\textwidth]{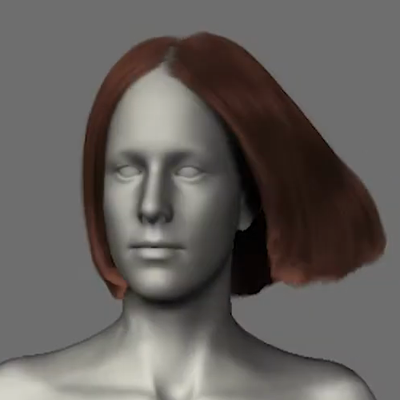} &
     \includegraphics[width=0.158\textwidth]{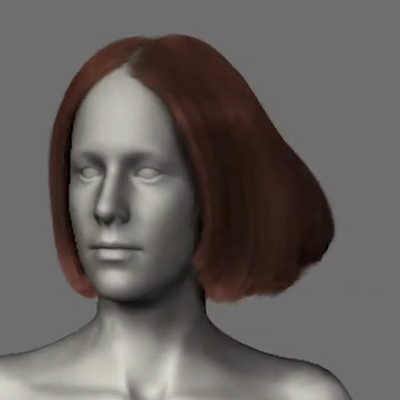} &
     \includegraphics[width=0.158\textwidth]{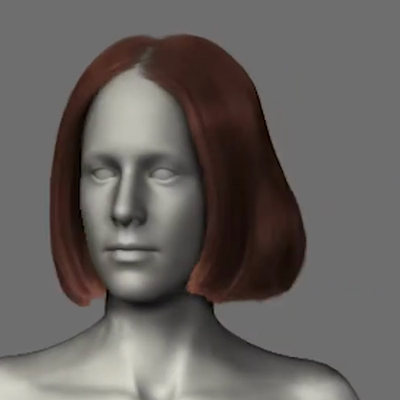} &
     \includegraphics[width=0.158\textwidth]{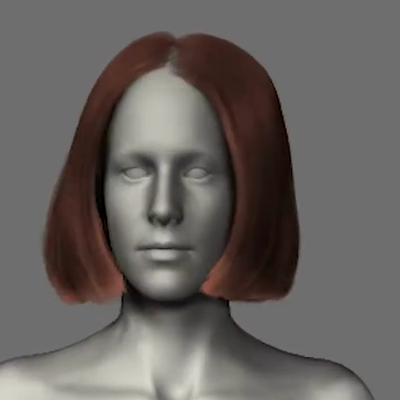} &
     \includegraphics[width=0.158\textwidth]{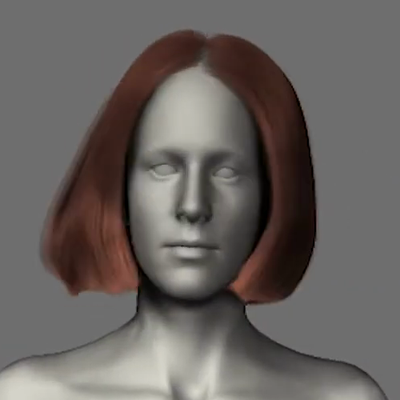} &
     \frame{\adjincludegraphics[height=0.157\textwidth, trim={0 {.2\height} {.7\width} {.5\height}}, clip]{fig/img/single_view/zju/157.png}} \\
     \addlinespace[-1pt]
     \raisebox{27pt}{\rotatebox[origin=c]{90}{\tiny HairStep}} &
     \includegraphics[width=0.158\textwidth]{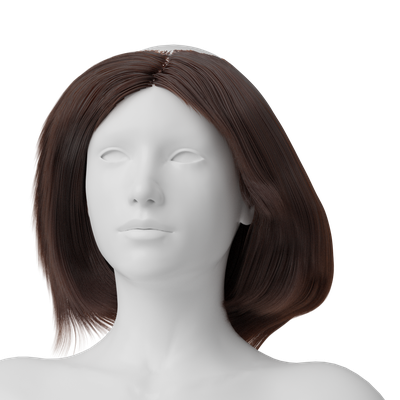} &
     \includegraphics[width=0.158\textwidth]{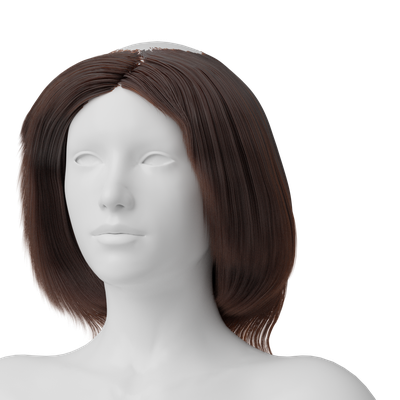} &
     \includegraphics[width=0.158\textwidth]{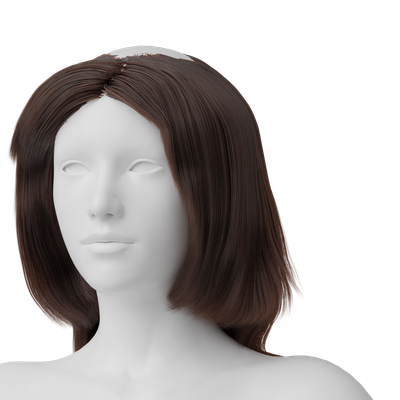} &
     \includegraphics[width=0.158\textwidth]{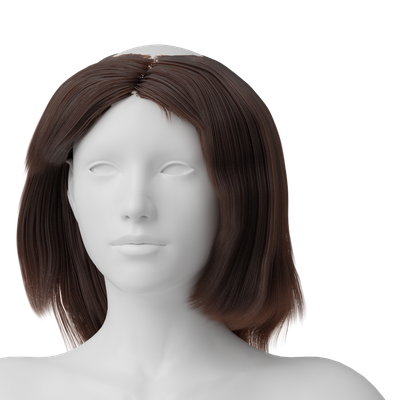} &
     \includegraphics[width=0.158\textwidth]{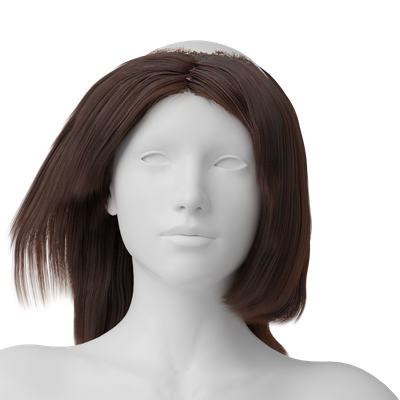} &
     \frame{\adjincludegraphics[height=0.157\textwidth, trim={0 {.2\height} {.7\width} {.5\height}}, clip]{fig/img/single_view/hairstep/bbj-000035.png}} \\
     \addlinespace[-1pt]
     \raisebox{27pt}{\rotatebox[origin=c]{90}{\tiny Ours}} &
     \includegraphics[width=0.158\textwidth]{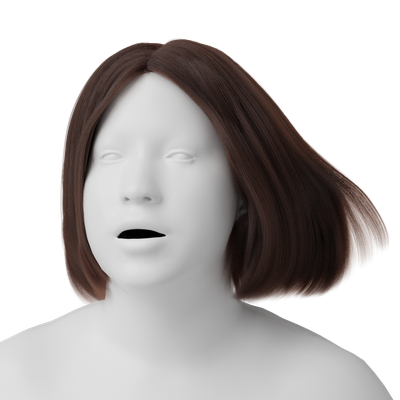} &
     \includegraphics[width=0.158\textwidth]{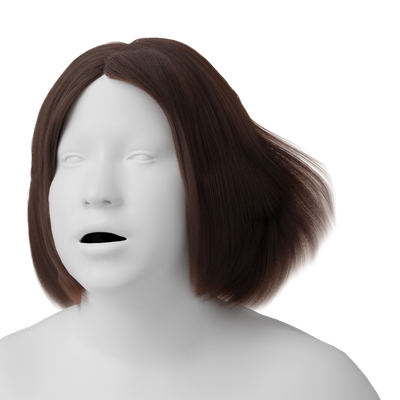} &
     \includegraphics[width=0.158\textwidth]{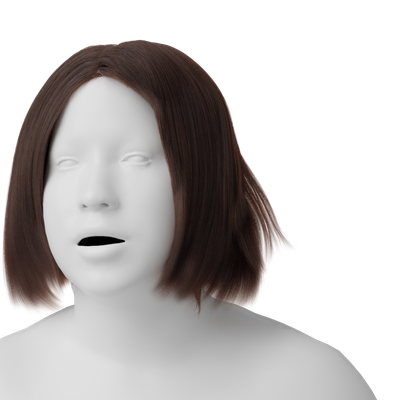} &
     \includegraphics[width=0.158\textwidth]{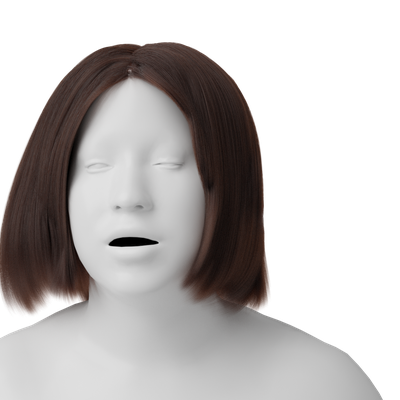} &
     \includegraphics[width=0.158\textwidth]{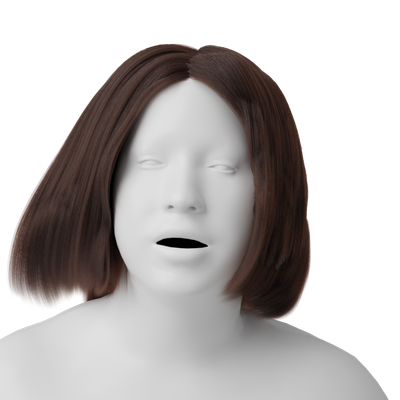} &
     \frame{\adjincludegraphics[height=0.157\textwidth, trim={0 {.2\height} {.7\width} {.5\height}}, clip]{fig/img/single_view/ours/bbj-000035.png}} \\
     & \multicolumn{5}{c}{\small Single-view Reconstruction} & \small Zoom-in Details \\
     \addlinespace[-1pt]
     \raisebox{27pt}{\rotatebox[origin=c]{90}{\tiny Ours}} &
     \includegraphics[width=0.158\textwidth]{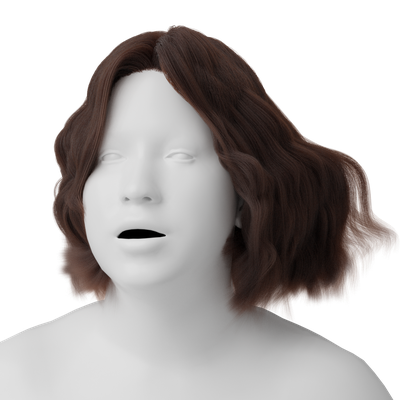} &
     \includegraphics[width=0.158\textwidth]{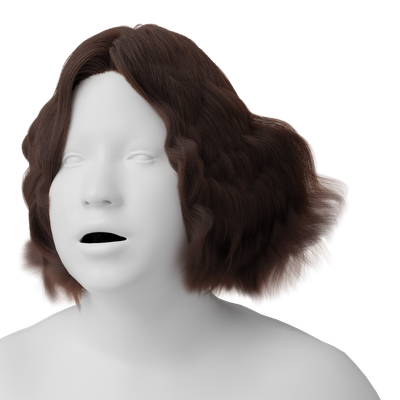} &
     \includegraphics[width=0.158\textwidth]{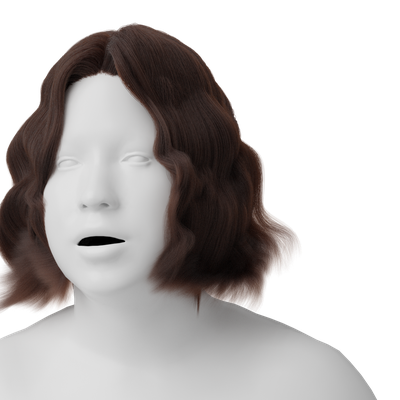} &
     \includegraphics[width=0.158\textwidth]{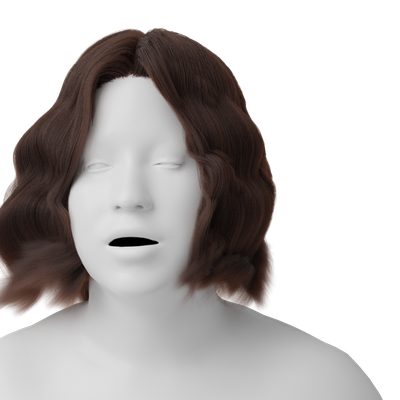} &
     \includegraphics[width=0.158\textwidth]{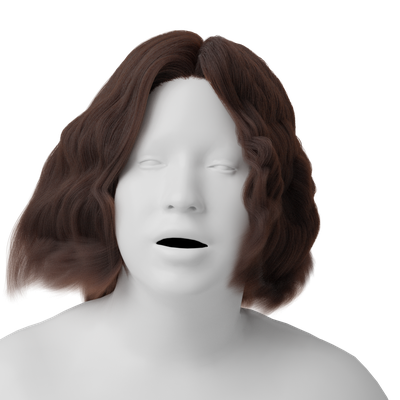} &
     \frame{\includegraphics[width=0.157\textwidth]{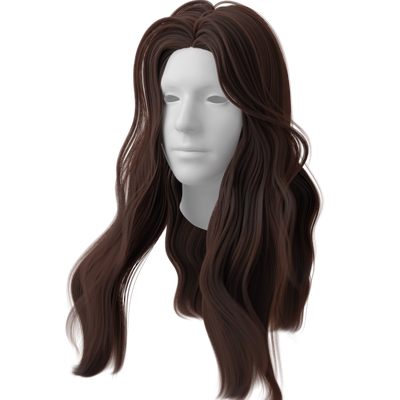}} \\
      & \multicolumn{5}{c}{\small Curl Pattern Editing} & \small Reference \\
    \end{tabular}
    \caption{Single-view hair reconstruction and editing on image sequences, with comparison to~\citep{yang2019dynamic} and HairStep~\citep{zheng2023hair}.}
    \label{fig:single-view-video}
    \vspace{-2mm}
\end{figure}

To test reconstruction for hair under dynamics, we run our algorithm on image sequences sampled from a video in~\citep{yang2019dynamic}, and compare our results to both~\citep{yang2019dynamic} and HairStep~\citep{zheng2023hair} in Fig.~\ref{fig:single-view-video}. Since \citet{yang2019dynamic} have not released the training data or pre-trained model of their method, we can only use their provided videos for comparison.

Among these methods, our method achieves the best reconstruction quality regarding both the global hair structure and local curl patterns, while also avoiding artifacts such as bald areas observed in HairStep.
Despite being trained solely on static data, our model demonstrates the ability to generalize and capture the dynamic effects of hair in the images. We conjecture that it is because our disentangled guide strand parameters are flexible enough to capture large hair deformation variations.
Additionally, since our reconstructed hairstyles are represented as parameters, we can substitute their $\vec{\beta}$ parameters with that of a wavy hairstyle (shown as reference in the last row), thereby transferring the wavy patterns to the reconstructed results while preserving their overall structure.

\begin{wrapfigure}[14]{r}{0.5\textwidth}
    \centering
    \addtolength{\tabcolsep}{-5pt}
    \begin{tabular}{cc}
        \includegraphics[width=0.24\textwidth]{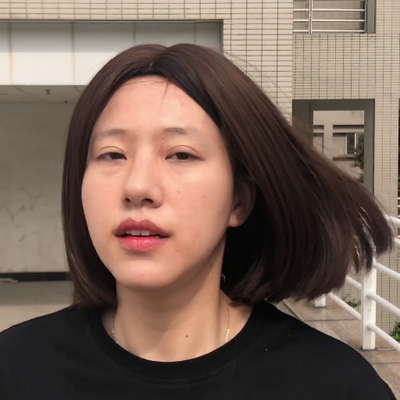} &
        \includegraphics[width=0.24\textwidth]{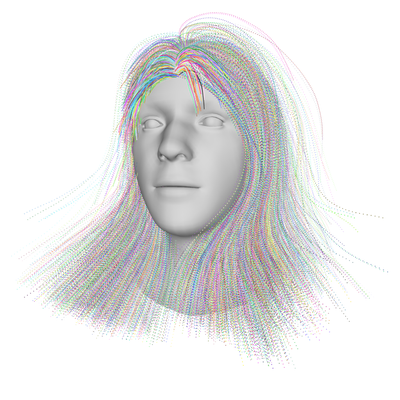} \\
        \tiny Input Image & \tiny HAAR~\citep{sklyarova2023haar} \\
    \end{tabular}
    \caption{Single-view hair reconstruction of HAAR~\citep{sklyarova2023haar}.}
    \label{fig:haar}
    \vspace{-2mm}
\end{wrapfigure}

Another recent work, HAAR~\citep{sklyarova2023haar}, also demonstrates the capability to generate 3D hairstyles from input images. As a text-conditioned diffusion model, their pipeline involves first extracting hairstyle descriptions from the input image using LLaVA~\citep{liu2024visual}, then generates a strand-based hairstyle based on these descriptions. However, since textual descriptions cannot capture all the intricate details of a hairstyle, HAAR often fails to accurately reproduce the hairstyle from the input image, resulting in only a rough resemblance (see Fig.~\ref{fig:haar}).

\subsection{Hair-conditioned Image Generation}

\begin{figure}[ht]
    \centering
    \addtolength{\tabcolsep}{-4.7pt}
    \begin{tabular}{cccccc}
     & & \multicolumn{4}{c}{\small \textit{``wavy and short hair, white sweater''}} \\
     \includegraphics[width=0.16\textwidth]{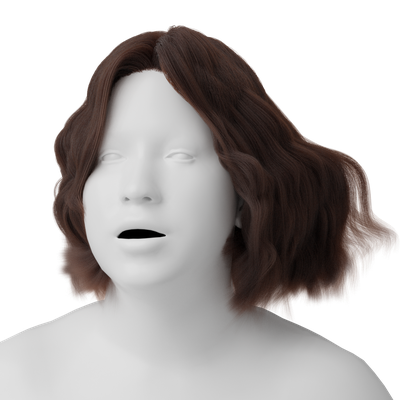} &
     \includegraphics[width=0.16\textwidth]{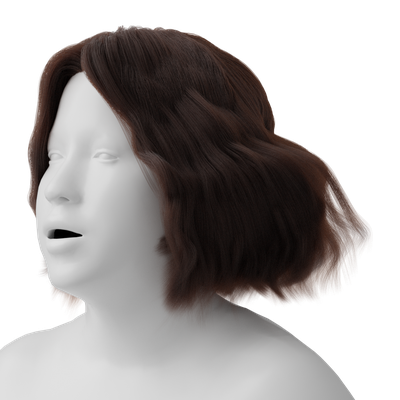} &
     \includegraphics[width=0.16\textwidth]{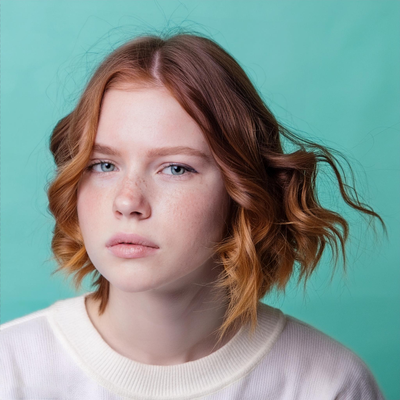} &
     \includegraphics[width=0.16\textwidth]{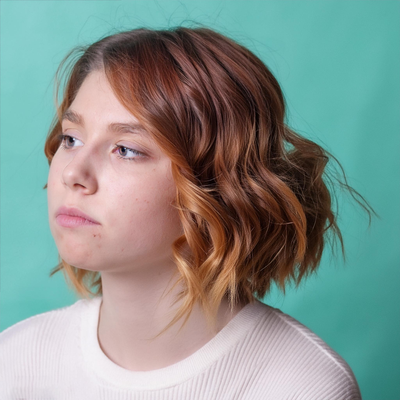} &
     \includegraphics[width=0.16\textwidth]{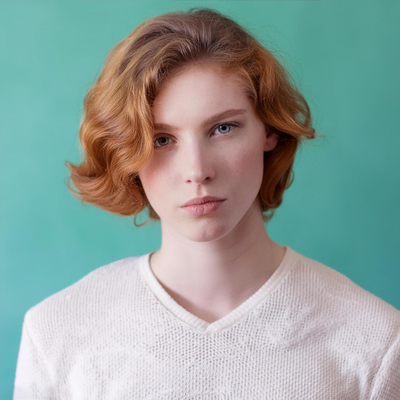} &
     \includegraphics[width=0.16\textwidth]{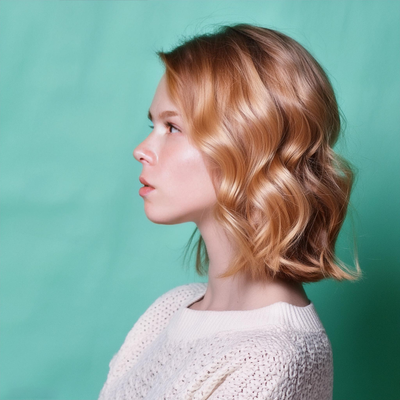} \\
     \multicolumn{2}{c}{\small Input Hair Conditions} & \multicolumn{2}{c}{\small w/ Conditions} & \multicolumn{2}{c}{\small w/o Conditions}
    \end{tabular}
    \caption{Comparison of image generation with and without hair conditions. These images are generated with the same text prompt \textit{``wavy and short hair, white sweater''}. Additional text prompts like \textit{``front face''} or \textit{``side face''} are appended to assist the head pose in the generated images.}
    \label{fig:firefly-eval}
    \vspace{-2mm}
\end{figure}

In Fig.~\ref{fig:firefly-eval} we present a comparison of image generation results with and without our input hair conditions. Note that we use the pre-visualization from MeshLab~\citep{meshlab} as the structural image condition, which we found to perform better than the final renderings. Without our hair conditions, images generated with rough text prompts like \textit{``wavy and short hair''} cannot guarantee the production of the desired hairstyle, and their hairstyles often vary with pose, resulting in a loss of multi-view consistency. Leveraging the rich information encoded in the 3D hair geometry, our hair conditions effectively address these issues, producing high-quality portrait images with more consistent hairstyles.

\begin{figure}[ht]
    \centering
    \addtolength{\tabcolsep}{-4pt}
    \begin{tabular}{cccc}
     \includegraphics[width=0.24\textwidth]{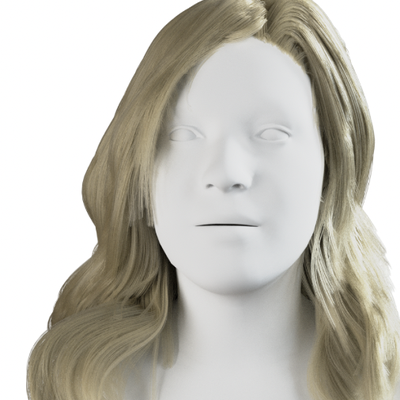} &
     \includegraphics[width=0.24\textwidth]{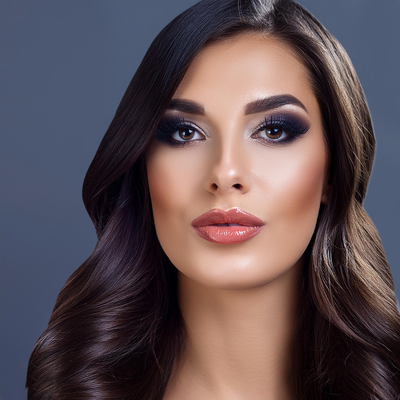} &
     \includegraphics[width=0.24\textwidth]{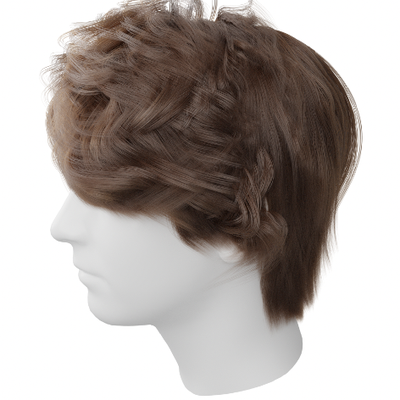} &
     \includegraphics[width=0.24\textwidth]{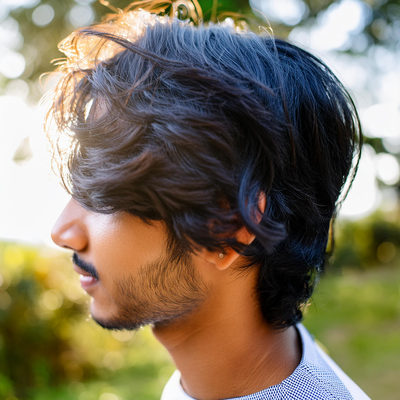} \\
     \small Input Hairstyle $1$ & \tiny \textit{``digital twin of a human''} & 
     \small Input Hairstyle $2$ & \tiny \textit{``side face, man, curly short hair''} \\
     \addlinespace[6pt]
     \includegraphics[width=0.24\textwidth]{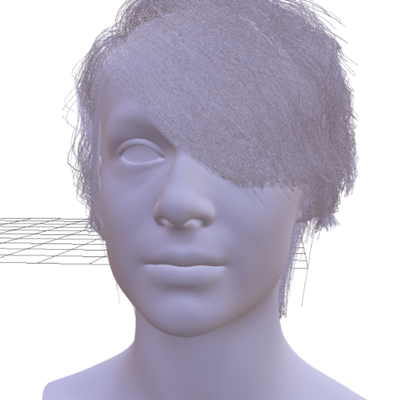} &
     \includegraphics[width=0.24\textwidth]{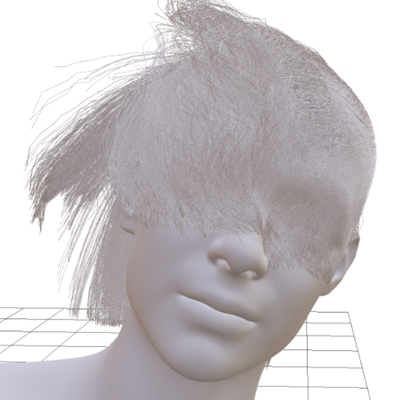} &
     \includegraphics[width=0.24\textwidth]{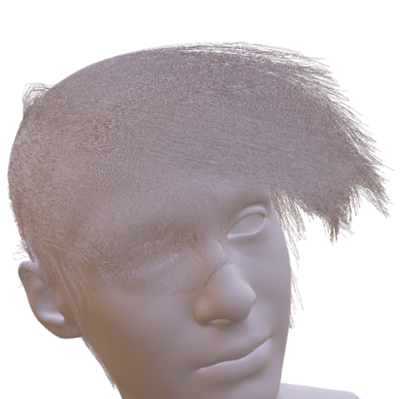} &
     \includegraphics[width=0.24\textwidth]{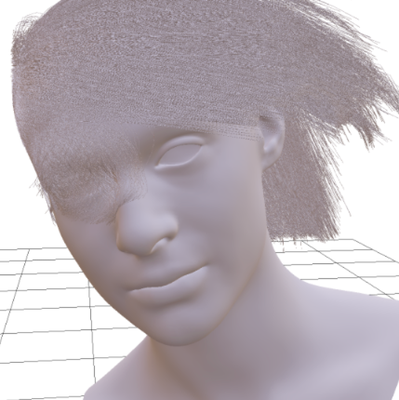} \\
     \multicolumn{4}{c}{\small Input Hairstyles Generated by \textsc{Perm} and Simulated by~\citep{alejandro2024ams}} \\
     \addlinespace[4pt]
     \includegraphics[width=0.24\textwidth]{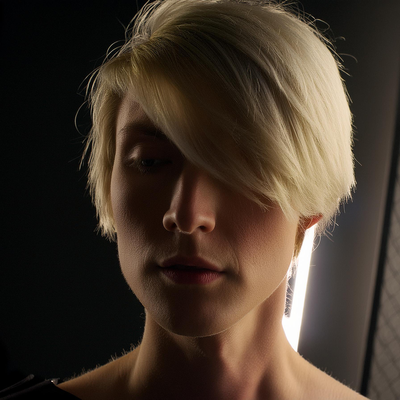} &
     \includegraphics[width=0.24\textwidth]{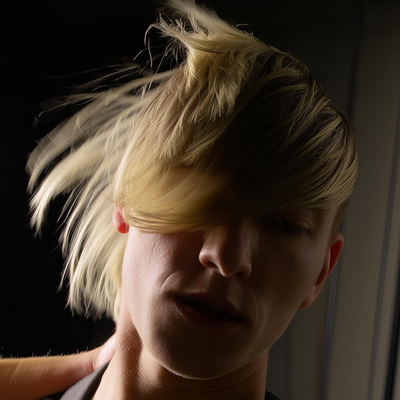} &
     \includegraphics[width=0.24\textwidth]{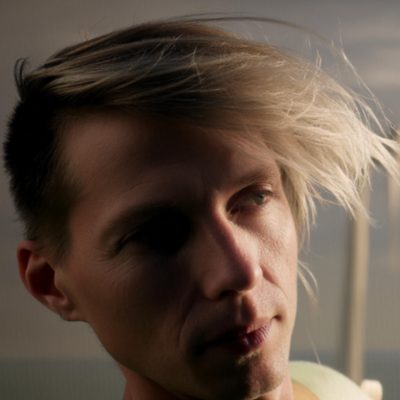} &
     \includegraphics[width=0.24\textwidth]{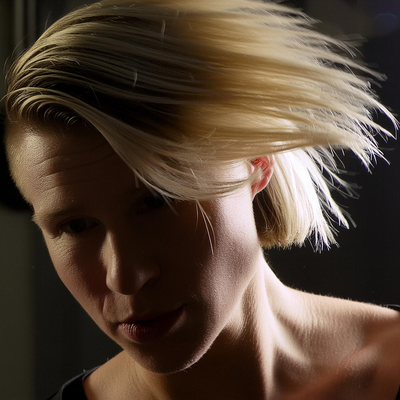} \\
     \multicolumn{4}{c}{\small \textit{``a man with short hair is shaking his head''}}
    \end{tabular}
    \caption{Hair-conditioned image generation using Adobe Firefly~\citep{firefly}.}
    \label{fig:firefly}
    \vspace{-2mm}
\end{figure}

In Fig.~\ref{fig:firefly} we further showcase some generated images conditioned on various input hairstyles. The texture colors in the input hairstyle renderings are only for aesthetic purposes and have no effect on the skin or hair colors in the generated images.
Although structural conditioned image generation is not our focus and needs more future investigation on improving the structural alignment, this application reveals the potential of deploying current T2I models as a ``neural renderer''~\citep{tewari2020state} within the traditional CG pipeline.

\new{\section{Limitations}}

\begin{figure}[ht]
    \centering
    \addtolength{\tabcolsep}{-6pt}
    \begin{tabular}{cccc}
        \includegraphics[width=0.25\textwidth]{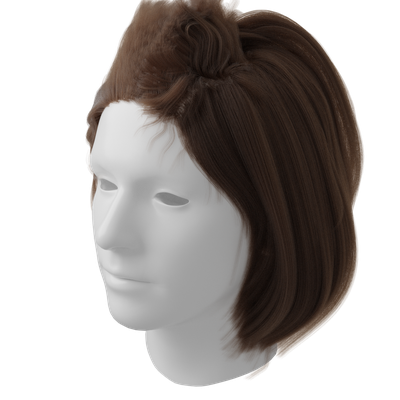} &
        \includegraphics[width=0.25\textwidth]{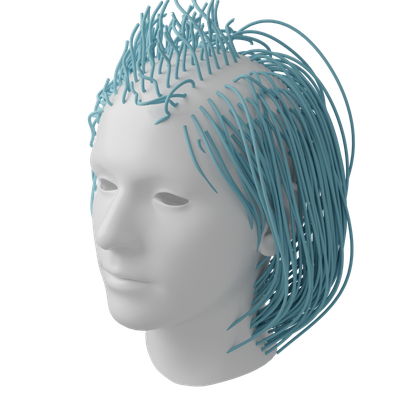} &
        \includegraphics[width=0.25\textwidth]{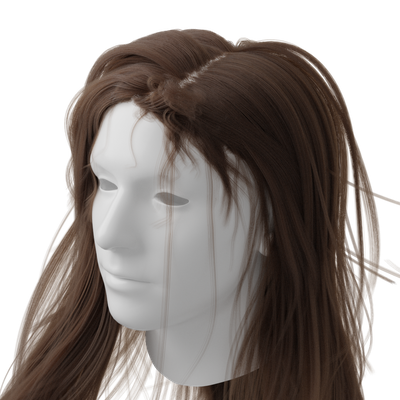} & 
        \includegraphics[width=0.25\textwidth]{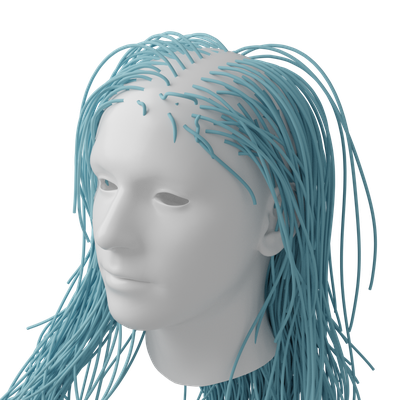} \\
    \end{tabular}
    \caption{\new{Unnatural hairstyles generated from random sampling.}}
    \label{fig:perm-sampling-failure}
    \vspace{-2mm}
\end{figure}

Our work still suffers from several limitations. First, our training data is limited, as high-quality 3D hair assets are notoriously challenging to acquire. Consequently, certain intricate hairstyles, such as afro styles and various braided patterns, are underrepresented, which limits the generative capabilities of our model and leads to unnatural outputs (see Fig.~\ref{fig:perm-sampling-failure}). Addressing this issue requires the development of a systematic framework for efficiently capturing a diverse range of real-world data -- a critical challenge that remains unresolved in the hair modeling area. However, \textsc{Perm} could serve as a pre-trained prior for efficient data capture, since it has the potential to fill in invisible parts of the hair, such as interiors and occluded regions, thereby accommodating more sparse image inputs. The captured hair data could also be used to fine-tune \textsc{Perm}, further bridging the domain gap between synthetic and real-world hairstyles.

\begin{figure}[ht]
    \centering
    \addtolength{\tabcolsep}{-5pt}
    \begin{tabular}{cccc}
        \includegraphics[width=0.245\textwidth]{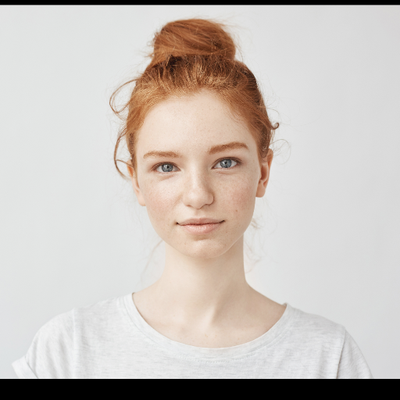} &
        \includegraphics[width=0.245\textwidth]{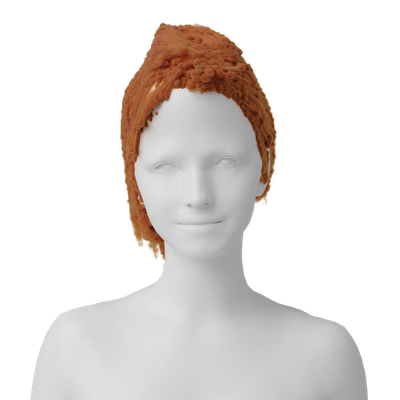} &
        \includegraphics[width=0.245\textwidth]{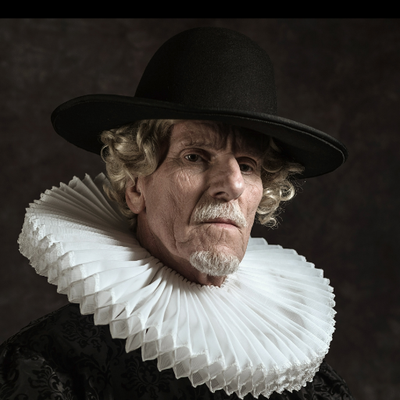} & 
        \includegraphics[width=0.245\textwidth]{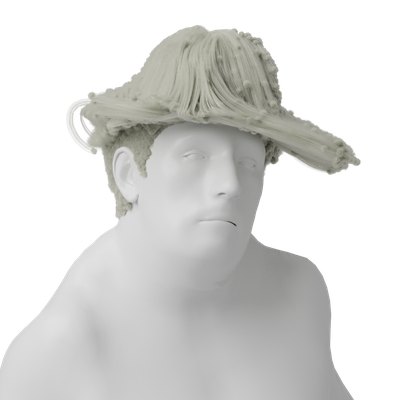} \\
    \end{tabular}
    \caption{\new{Suboptimal single-view reconstruction results in extreme cases, where the hair is tied in certain hairstyles such as buns (left) or heavily occluded by accessories such as hats (right).}}
    \label{fig:single-view-failure}
    \vspace{-2mm}
\end{figure}

Second, our current single-view hair reconstruction pipeline adopts a relatively straightforward approach, relying on an optimization process that primarily focuses on per-pixel strand information encoded in the 2D rendered images. These supervisions may fail to capture fine strand-level details, yielding suboptimal outputs in extreme cases (see Fig.~\ref{fig:single-view-failure}). To enhance reconstruction quality, existing methods~\citep{chai2012single, nam2019lpmvs} could be incorporated to trace continuous strand segments from single-view or multi-view input images, providing a more reliable geometric clue for our optimization.
Additionally, more advanced optimization techniques could be explored. For example, integrating patch-based losses, such as LPIPS~\citep{zhang2018perceptual}, may help encourage the preservation of strand details at the patch level, thereby improving the fidelity of the reconstructed hairstyles.

\end{document}